\documentclass[a4paper]{article}

\usepackage[utf8]{inputenc}
\usepackage[T1]{fontenc}
\usepackage{cite}
\usepackage{amsmath}
\usepackage{mathtools}
\usepackage{amssymb}
\usepackage{gensymb}
\usepackage{tabularray}

\usepackage[toc,page]{appendix}
\usepackage{caption}
\usepackage{subcaption}
\usepackage{float}

\usepackage{tikz}
\usetikzlibrary{calc,patterns,
                 decorations.pathmorphing,
                 decorations.markings}
\usetikzlibrary{positioning}
\usetikzlibrary{positioning, shapes, fit} 
\usetikzlibrary{bayesnet}

\usepackage[hidelinks]{hyperref} 
\usepackage[a4paper,
            total={170mm, 230mm}]{geometry}

\title{Active transfer learning for structural health monitoring}

\date{\vspace{-5ex}}

\begin{document}
\author{
  \small
  J. Poole$^a$, N. Dervilis$^a$, K. Worden$^a$, P. Gardner$^a$, 
  V. Giglioni$^b$, R. S. Mills$^a$, and A. J. Hughes$^a$ \\
  \small
  $^a$Dynamics Research Group, 
  School of Mechanical, Aerospace and Civil Engineering, \\ 
  \small
  University of Sheffield, Mappin Street, Sheffield S1 3JD, UK \\
  \small
  $^b$Department of Civil and Environmental Engineering, University of Perugia, \\ 
  \small
   Via G. Duranti 93, Perugia 06125, Italy
}

\maketitle

\begin{abstract}
\textbf{Abstract.} Data for training structural health monitoring (SHM) systems are often expensive
and/or impractical to obtain, particularly for labelled data. Population-based SHM (PBSHM) aims to address this limitation by leveraging data from multiple structures. However, data from different structures will follow distinct distributions, potentially leading to large generalisation errors for models learnt via conventional machine learning methods. To address this issue, transfer learning -- in the form of domain adaptation (DA) -- can be used to align the data distributions. Most previous approaches have only considered \emph{unsupervised} DA, where no labelled target data are available; they do not consider how to incorporate these technologies in an online framework -- updating as labels are obtained throughout the monitoring campaign. This paper proposes a Bayesian framework for DA in PBSHM, that can improve unsupervised DA mappings using a limited quantity of labelled target data. In addition, this model is integrated into an active sampling strategy to guide inspections to select the most informative observations to label -- leading to further reductions in the required labelled data to learn a target classifier. The effectiveness of this methodology is evaluated on a population of experimental bridges. Specifically, this population includes data corresponding to several damage states, as well as, a comprehensive set of environmental conditions. It is found that combining transfer learning and active learning can improve data efficiency when learning classification models in label-scarce scenarios. This result has implications for data-informed operation and maintenance of structures, suggesting a reduction in inspections over the operational lifetime of a structure -- and therefore a reduction in operational costs -- can be achieved.
\\


\end{abstract}

\section{Introduction}

Data for training structural health monitoring (SHM) systems are often expensive and/or difficult to obtain, particularly for labelled data relating to rare health-states, such as damage data. While unsupervised approaches have been shown to be capable of detecting damage~\cite{dervilis2014damage, FarrarC.R.CharlesR.2013Shm:}, data-based SHM systems typically cannot provide contextual information -- often encoded as labels -- without labelled data. Population-based SHM (PBSHM) is an emerging field that aims to address the issue of data scarcity by considering data from across a population of structures \cite{Gardner2021, Gosliga2021, Bull2021}.  However, datasets derived from different structures will not follow the same underlying generative distribution, meaning the assumption that training and testing data were drawn from the same distribution is invalid; thus, supervised machine-learning techniques will likely have a large generalisation error \cite{Murphy2014}.\\

This issue motivates the application of transfer learning (TL), a field of machine learning that aims to improve the performance in a target domain (structure) by leveraging related information from a more data-rich source domain (structure) \cite{yang2020transfer}. Unsupervised-transfer learning approaches, such as unsupervised domain adaptation (DA), have emerged as a promising method and have been demonstrated in a variety of SHM applications \cite{Gardner2020, gardner2022population, Gardner2021a, bull2021transfer, Michau2019, wang2020triplet, Li2020}. These methods do not require labelled data in the target, meaning that they have the potential to facilitate damage classification when label information is missing in the target, by allowing a predictive function learnt using source data to generalise to the target. Nevertheless, these approaches are not infallible and they may result in performance degradation - so-called negative transfer \cite{yang2020transfer} – if the datasets are not sufficiently related. They are also particularly sensitive to issues related to class imbalance \cite{Cao2018a} -- where certain classes are under-represented in the training data -- which is already a pertinent issue in SHM datasets, as some health-states are naturally more common.\\

\begin{figure}[t!]
\centering
\includegraphics[width=0.95\textwidth]{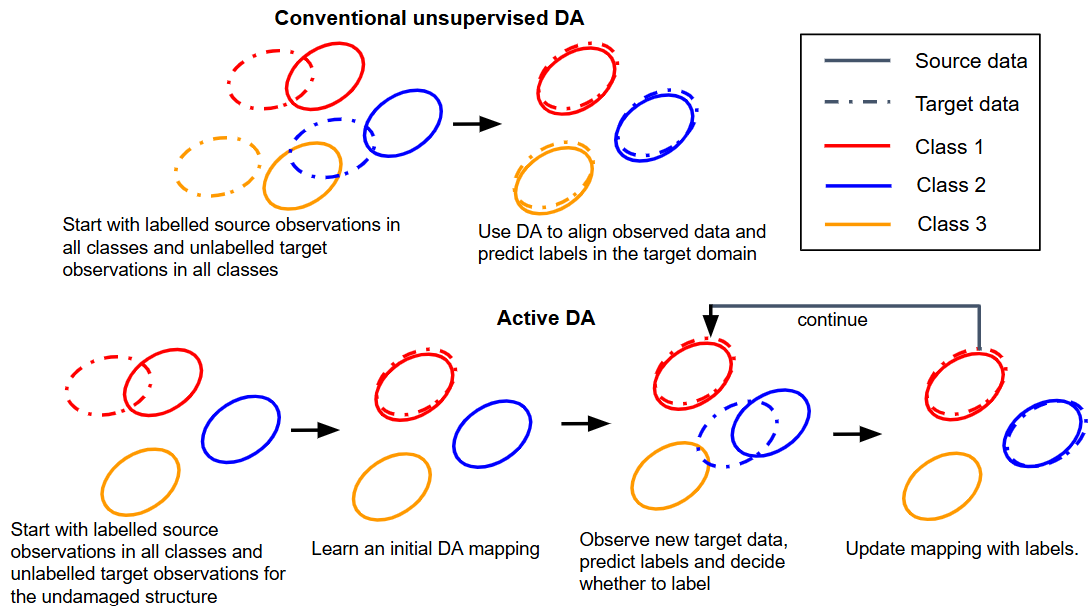}
\caption{A demonstration of the assumptions made by conventional unsupervised DA (top) and the active DA approach proposed in this paper (bottom).}
\label{fig:demon}
\end{figure}

Previous applications of DA in PBSHM have mostly focussed on scenarios where the target domain consists solely of unlabelled data potentially representing all health-states of interest \cite{ Gardner2021, gardner2022population,Gardner2021a, bull2021transfer, Michau2019}. However, in practice, health-state data are sequentially observed and potentially labelled via inspections; the typical DA setting is compared with an illustrative online SHM scenario in Figure \ref{fig:demon}.  To leverage these sequential observations effectively DA algorithms are required to be robust in the following two scenarios. Firstly, at the onset of the monitoring campaign, the target structure may only have data related to normal operation, while the source dataset(s) could encompass a wider range of the health-states of interest; this situation where datasets are imbalanced is shown at the start of the active DA illustration in Figure \ref{fig:demon}, where only data for one class is avialable in the target dataset, in comparison to three classes in the source dataset. This necessitates robustness in DA algorithms to highly-imbalanced datasets, where data for many classes will be unavailable in the target. Second, these methods should be capable of adapting as contextual information is acquired during the monitoring campaign; this may improve generalisation and reduce the likelihood of negative transfer as the monitoring campaign progresses \cite{Ben-David2010}, when maintenance decisions become more critical.\\

Given the cost of acquiring labels, it would be beneficial to schedule inspections to coincide with the most informative data. Active learning has been demonstrated to significantly reduce the label requirement in SHM by using a predictive model to infer which unlabelled data would provide the largest improvement if they were labelled \cite{feng2017deep, chakraborty2015adaptive, bull2019probabilistic, hughes2022risk, hughes2022robust}.  In the transfer-learning literature, guided sampling strategies have been proposed to leverage source data to improve the initial model \cite{rai2010domain, saha2011active, xie2022active, ma2021active}, and have been demonstrated to mitigate the class-imbalance issue in DA \cite{ma2021active}; these methods will be referred to as \emph{active transfer learning} methods. However, to the authors knowledge active transfer learning has not been investigated in SHM. \\

This paper proposes the first online transfer-learning strategy for PBSHM by incorporating a novel Bayesian DA method into an active-learning framework. This online framework is able to update mappings estimated via unsupervised DA using limited labels; the general proposed approach is illustrated in Figure~\ref{fig:demon}. This approach is validated using an experimental dataset consisting of three laboratory-scale bridges with varied support locations; these structures were subjected to testing over a range of damage-states and environmental conditions using an environmental chamber.\\

This paper is structured as follows. Section 2 outlines the necessary background about transfer learning and active learning. Section 3 introduces the proposed DA methodology and Section 4 presents the experimental datasets and demonstrates the transfer of a damage classifier using the proposed method. Finally, conclusions are presented in Section 5 and potential future work is highlighted.

\section{Towards an online framework for transfer learning in PBSHM}

Transfer learning can be used to learn predictive models with entirely unlabelled data or only a limited amount of labelled data in the target domain \cite{Zhuang2021}, by leveraging data from a related source domain. However, the applicable TL methods are dependent on the quantity and type of data required in the target dataset; methods that can leverage labelled target data generally have a lower likelihood of leading to high generalisation error \cite{Ben-David2010} - motivating the active transfer learning framework presented in this paper. This section provides a brief overview of transfer learning, and its associated issues for SHM applications, as well as active learning as a method to guide the labelling process of online streams of data.

\subsection{Transfer learning: background and problem statement}

In transfer learning, a \emph{domain} \(\Omega = \{\mathcal{X}, p(\mathbf{x})\} \) is characterised by a $d$-dimensional feature space \( \mathcal{X} \in \mathbb{R}^d \) and a marginal probability distribution \( p(\mathbf{x}) \) over that space, where \( \mathbf{x} \in \mathcal{X} \)  \cite{Zhuang2021}. Associated with each domain is a \emph{task} \( \mathcal{T} = \{\mathcal{Y}, f(\cdot)\} \), defined by a label space \( \mathcal{Y} \) specifying the set of possible label values and a predictive function \( f(\cdot) \), predicting labels  \( y \in \mathcal{Y} \). In a probabilistic framework, the predictive function may also be interpreted as modelling the conditional distribution \( p(y|\mathbf{x}) \). The source and target domain distributions will be denoted by $p_s(\cdot)$ and $p_s(\cdot)$, respectively.\\

 In \emph{unsupervised TL}, a source dataset $\mathcal{D}_s=\{\mathbf{x}_{s,i},y_{s,i}\}_{i=1}^{n_s}$, with $n_s$ source observations, each with labels $y_{s,i}$ and an unlabelled target dataset $\mathcal{D}_{t,u}=\{\mathbf{x}_{t,j}\}_{j=1}^{n_{t,u}}$ with $n_{t,u}$ unlabelled target instances $\mathbf{x}_{t,j}$, are used to learn a predictive function that generalises to the target domain \footnote{This definition of TL follows \cite{pan2020transfer}, unsupervised TL is also often referred to as transductive TL as in \cite{Pan2010}.}.  \\
 
 A prominent approach for unsupervised TL involves \textit{mapping-based} domain adaptation (DA) \cite{Zhuang2021}. For a predictive model to generalise effectively between source and target domains, their conditional distributions must be similar, i.e. \( p_s(y|\mathbf{x}) \approx p_t(y|\mathbf{x}) \), implying that the predictive relationship between inputs and outputs is domain-invariant. Mapping-based DA addresses the scenario where both the marginal distributions \( p_s(\mathbf{x}) \neq p_t(\mathbf{x}) \) and the conditional distributions \( p_s(y|\mathbf{x}) \neq p_t(y|\mathbf{x}) \) differ. Thus, the goal is to learn a mapping \( \phi \) that projects both domains into a shared feature space such that the source and target conditional distributions are invariant, i.e. $ p_s(y|\phi(\mathbf{x})) = p_t(y|\phi(\mathbf{x}))$. Estimating the conditional distributions directly is challenging without labelled target data. Therefore, unsupervised DA algorithms typically make the implicit assumption that  $\phi$ can be learnt by minimising a measure of divergence between the unlabelled data, such as the marginal-distribution divergence \cite{JialinPan2011, Long2013, Ganin2017}. Whether this assumption holds for a given transfer task depends on domain similarity, typically assessed using domain knowledge \cite{pan2020transfer}.\\
 
Unsupervised TL could significantly increase the value of SHM data, as labelled source data could be used to learn predictive models in related structures without directly damaging or performing costly inspections of the target structure. However, unsupervised transfer learning relies on the assumption that a mapping $\phi$ that satisfies $p_s(y|\phi(\mathbf{x})) = p_t(y|\phi(\mathbf{x}))$ can be learnt without labelled target data. In cases where this assumption is not valid, transfer learning can result in \emph{negative transfer}, where transfer learning results in worse performance compared to using only the target data. Negative transfer can be specifically quantified by assessing the risk in the target domain, which is given by,
\begin{equation}
R(f) = \mathbb{E}_{(\mathbf{x}_t,y_t) \sim \mathcal{D}_{t,l}}[\ell(f(\mathbf{x}_t), y_t)]
\end{equation}
where $R(f)$ represents the risk of a predictive function $f(\cdot)$, output by an algorithm taking the source and target dataset $f = A(\mathcal{D}_s, \mathcal{D}_t)$\footnote{In this paper, the algorithm $A(\cdot)$ refers to both the transfer learning algorithm and an active learning method.}, and this predictive function applied to a labelled target dataset $\mathcal{D}_{t,l}$, to obtain the expectation $\mathbb{E}[\cdot]$ of a loss function $\ell(\cdot,\cdot)$. Negative transfer can be defined as the event the risk of the predictive function, learnt via transfer learning, $R(A(\mathcal{D}_s, \mathcal{D}_t))$, is greater than the risk of a predictive function learnt using only the target data, $R(A(\emptyset, \mathcal{D}_t))$, as follows \cite{wang2019characterizing},
\begin{equation}
    R(A(\mathcal{D}_s, \mathcal{D}_t)) > R(A(\emptyset, \mathcal{D}_t))
\end{equation}
where $\emptyset$ represents an empty set. Given the potential cost of misinforming decisions in SHM, it is pertinent that transfer learning strategies are robust for SHM applications - particularly near the end of a structures operating life, where the likelihood of incipient damage and unsafe operating conditions is higher. To avoid negative transfer, generally the transfer problem must be well-posed -- the source and target domains must be similar -- and the TL method should be able to effectively leverage this shared information.    \\

A potential method to reduce the likelihood of negative transfer is to incorporate target labels into the learning process, which generally leads to a lower generalisation error by providing data to directly minimise the empirical target risk as discussed in \cite{Ben-David2010}. This paper addresses the problem of identifying a TL method that can leverage data in an online setting to reduce the likelihood of negative transfer in comparison to fully unsupervised TL; related domains are identified using engineering expertise and modal analysis \cite{poole2023towards}. More principled approaches for quantifying domain similarity and the likelihood of negative transfer remain active areas of research \cite{Gosliga2021}. For a deeper discussion on the issue of negative transfer the interested reader may refer to \cite{Ben-David2010, wang2019characterizing}.  \\

Unsupervised TL is particularly challenging in SHM, as in many cases only data from the undamaged and perhaps a few damage-states, will be available in the target domain, as illustrated in Figure \ref{fig:demon}. In such scenarios, domain adaptation must be robust to variations in the label space of the target domain. Specifically, the target label space \( \mathcal{Y}_t \) may be a subset of the source label space $\mathcal{Y}_s$, i.e., \( \mathcal{Y}_t \subseteq \mathcal{Y}_s \). Alternatively, novel health states may be present in the target domain that are absent from the source. In this case, a mapping must be learned via the shared subset of classes, such that \( \mathcal{Y}_t \cap \mathcal{Y}_s \neq \emptyset \). When there is a mismatch between the source and target label spaces, conventional unsupervised DA is prone to negative transfer as a measure on the distance between the marginal data distributions using the available data does not reflect the discrepancy been the underlying marginal distributions; thus, specialised methods are generally required \cite{Cao2018a}. Another advantage of leveraging labelled data is that methods that leverage target labels are generally more robust to situations where the target dataset includes only a small subset of all possible health-states, as class distributions or boundaries can be directly aligned \cite{hoffman2013efficient}.\\

While in SHM it is often unfeasible to obtain comprehensive labelled datasets because of budget constraints and/or safety/accessibility issues, it may still be feasible to obtain labels for a few health-states throughout the operation of the target structure via periodic, or guided, inspections. In such cases, supervised TL could be used to increase the available information to learn shared regularities between domains \cite{wang2019characterizing}. In supervised TL, both the source dataset, $\mathcal{D}_{s}$, and the target dataset, $\mathcal{D}_{t,l}=\{\mathbf{x}_{t,j}, y_{t,j}\}_{j=1}^{n_{t,l}}$, contain labelled data, with $n_{t,l}$ representing the number of labelled target instances. \\

To facilitate a practical online framework for PBSHM, this paper aims to combine the advantages of both unsupervised and supervised TL methods, including labelled target data as it becomes available, to continually improve predictive performance. To achieve this objective, a TL model is proposed with two main objectives:

\begin{enumerate}
    \item From the start of the monitoring campaign, the method should allow for classification of health-states that have only been observed in the source domain; this method should be applicable using limited target data, potentially only related to the undamaged target structure. It should ideally also reflect uncertainty on the mapping, such that the uncertainty estimates provided by predictive models trained using source data are calibrated to the target domain.
    \item As labels are acquired throughout the operation of a structure, a TL method should be able to update to leverage this additional information. Furthermore, it would be beneficial if this information could be used to improve the prediction of classes which have only been observed (and labelled) in the source domain.  
\end{enumerate}

\subsection{Selecting informative labels: probabilistic active learning}

As labels are acquired throughout the target structure's monitoring campaign, it may be possible to improve generalisation and mitigate the likelihood of negative transfer of a transfer learner. However, budget restrictions will limit the number of observations in the target domain that can be labelled. Thus, it would be beneficial for inspections to coincide with the most informative samples to label. One approach to guide inspections is to use an initial model to classify (online) streams of data, and use the predictions to inform which samples should be labelled; generally, this is the main objective of stream-based active learning \cite{settles2009active}. \\

Active learning generally aims to develop approaches for two main settings: stream-based and pool-based \cite{settles2009active}. In stream-based active learning, data are acquired sequentially, and the active learner must determine whether to label, or \emph{query}, the current observation; generally, if the observation is not labelled in this instance, it cannot be labelled retrospectively. Alternatively, pool-based methods aim to label the more informative data from a previously obtained unlabelled dataset. In SHM, it is typically not possible to obtain labels of previously obtained data; only the current condition can be investigated. Thus, stream-based methods are the focus of this paper.\\

The specification of the sampling strategy is crucial, as it determines which data are most likely to be selected for labelling. One of the most widely used approaches is uncertainty sampling \cite{settles2009active}. For example, maximum entropy sampling (MES) selects data with the highest entropy, prioritising queries for observations where the current model yields the most uncertain or ``confused'' label probabilities \cite{settles2009active}. Commonly, uncertainty is measured using the Shannon entropy \cite{mackay2003information} of the posterior-predictive-distribution,

\begin{equation}
    H(\hat{y}_i) = - \sum_{c=1}^{C} p(\hat{y}_i = c | \mathbf{x}_i, \mathcal{D}_{l}) \text{log} \ p(\hat{y}_i = c | \mathbf{x}_i, \mathcal{D}_{l})
\end{equation}

\noindent where $C$ is the total number of classes and $\mathcal{D}_{l}$ represents the labelled training dataset. Entropy-based sampling typically results in labelling samples that lie close to the boundaries of the classifier, which should be the most informative data for defining classification boundaries between previously observed classes \cite{vapnik2013nature}. A weakness of this approach is that for most classifiers, observations at the extremities of the model will not be queried, meaning it may not query data corresponding to novel classes. When using generative models, another approach to uncertainty-based sampling is to sample observations with low-likelihood values \cite{settles2009active}. These queries would appear more ``novel'' to the model, rather than confused; thus, this query strategy is well-suited for novelty detection. To combine the benefits of either approach, these measures can be combined into a joint strategy to obtain more varied labelled datasets and reduce sampling bias \cite{bull2019probabilistic}. Other approaches aim to label samples which are expected to improve the model as quickly as possible. These methods often select samples which would lead to the largest reduction in entropy of the posterior distribution of a Bayesian model \cite{settles2009active}.\\

Labelling data based on a criterion has been shown capable of reducing overall labelling efforts \cite{settles2009active}; however, training datasets will not be representative of the underlying distributions - a phenomenon known as \emph{sampling bias}. For example, data may be over-represented near boundaries using MES.  This issue may lead to worse performance than random sampling, particularly as larger datasets are obtained \cite{hughes2022robust, dasgupta2008hierarchical}. In the worse cases, poor initial models can cause suboptimal model convergence, where data relating the the optimal model will never be sampled under the selection criterion \cite{bull2018active}. This issue is generally dependent on the labelling criterion used; development of criterion that mitigate sampling bias is a major research focus in active learning \cite{settles2009active, bull2022sampling}.\\

\begin{figure}[h!]
    \centering
        \includegraphics[width=0.8\textwidth]{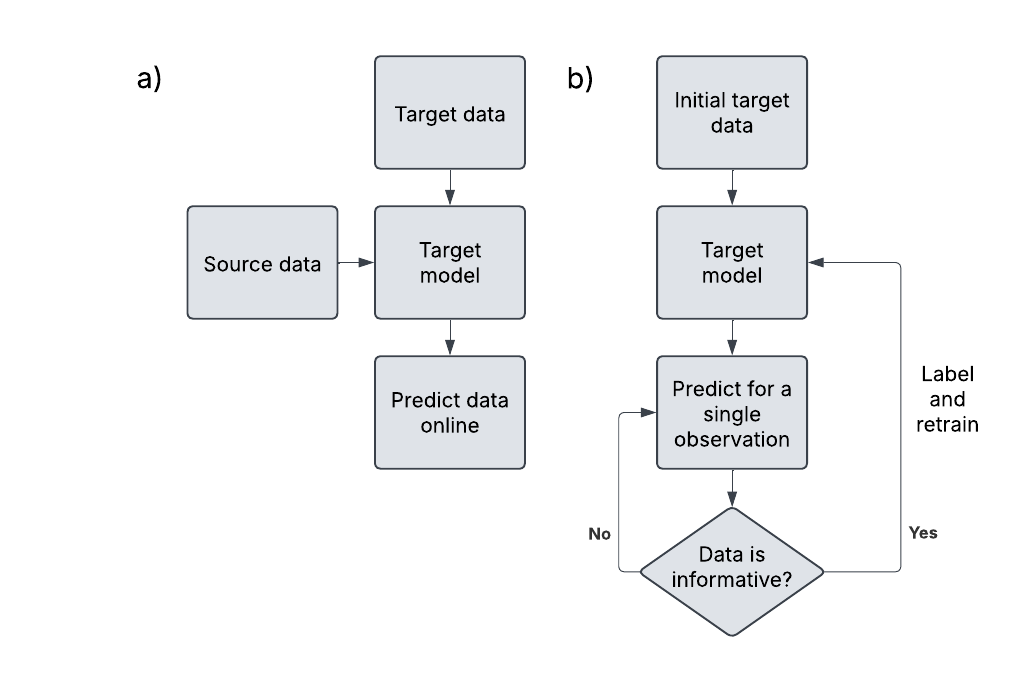}
    \caption{Flow chart showing the process of learning a predictive model with transfer learning (a) and stream-based active learning (b).}
    \label{fig:flow2}
\end{figure}

\subsection{Active transfer learning}

The process of learning a predictive model using transfer learning or active learning for use with online SHM data is illustrated in Figure \ref{fig:flow2}(a) and Figure \ref{fig:flow2}(b), respectively. In comparison, active transfer learning aims to initialise the target modelling by also using source data. By considering TL in an active-learning framework, the drawbacks of considering either approach independently can be alleviated.\\

As previously discussed, from a TL perspective incorporating labels can reduce the likelihood of negative transfer and improve generalisation where unsupervised TL alone achieve insufficient classification performance \cite{wang2019characterizing}. Using more informative labelled datasets selected via active sampling could achieve these improvements with fewer labels \cite{xie2022active}; thus, facilitating the application of supervised TL with smaller labelling budgets. \\

From the perspective of active learning, leveraging source data has several advantages. First, using transfer to initialise the active learner may result in a stronger initial model, meaning that it can select more meaningful samples from the start of the process \cite{rai2010domain}. However, it should be noted that there is also a risk that important target samples will not be labelled if initial transfer is poor; this will be discussed more in the following section. In addition, uncertainty-based methods will likely require fewer samples from classes with abundant source data before they can be classified with low uncertainty -- leading to fewer samples overall \cite{xie2022active}. Furthermore, while conventional active learning allows for classifiers to be learnt without a fully labelled dataset \emph{a priori}, observations can only be labelled as classes that have been previously observed. However, using an appropriate TL strategy, classification of classes that have only been observed in the source domain could also be attempted.\\

Active transfer learning presents a promising approach for training predictive models in sparse data settings. However, using a more informative transfer-based model also has the potential to compound the effect of sampling bias. For example, if a source classifier has well-defined boundaries for a wide range of classes, using MES the likelihood of obtaining data in these regions would be low, potentially meaning incorrect boundaries defined using source data are not corrected via sampling target data. Thus, the issue of negative transfer is still an important issue and requires the transfer task to be well posed \cite{pan2020transfer}. To mitigate this problem, this paper uses a probabilistic mapping to increase uncertainty in target predictions, particularly at the start of the active sampling process.  \\

\subsection{Related work}

This paper presents both a novel DA method and introduces the concept of active transfer learning to PBSHM. Thus, this section provides a summary of how the proposed methodology related to the previous TL and active learning literature. \\

Many of the previous applications of TL to SHM have focused on unsupervised DA to transfer source labels in the absence of any target labels. For example, it has been applied in the context of PBSHM to learn damage classifiers for multi-storey structures \cite{Gardner2020}, bridges \cite{gardner2022domain, giglioni2024domain}, and aircraft wings \cite{Gardner2020b, gardner2022population}. Xu \emph{et al.} \cite{Xu2020} used multi-source DA to perform damage quantification. Domain adaptation has also been used to improve damage detection in tailplanes  \cite{bull2021transfer} and bridges \cite{figueiredo2023transfer}. There have also been a number of applications of deep-DA architectures proposed to perform fault diagnosis in machines under changing loading conditions and rotation speeds \cite{wang2020triplet, Li2019, Michau2019, Li2020, jiao2020residual}. However, these methods assume there are unlabelled observations for each of the damage-states of interest in the target domain, as depicted in the top flow chart in Figure \ref{fig:demon}; whereas, in practice, it is unlikely there would be observations of multiple damage scenarios in the target domain. Moreover, if damage in the target structure is detected, a few labels could be collected; however, these previous applications do not incorporate any labels.  \\

A few examples exist where target labels have been used for TL in SHM. Previous applications of supervised TL mostly focus on fine-tuning of neural networks; for example, to perform crack detection in images \cite{cao2018preprocessing, Gao2018, dorafshan2018comparison, zhu2020vision} and unprocessed frequency response data \cite{cao2018preprocessing, teng2023structural}. Meta-learning using neural networks  has also been used to learn a model that can be updated with a novel damage class with few labels \cite{xu2021attribute}. However, these applications all require labelled data from all classes of interest in the target domain. In practice, damage will be observed and (potentially) labelled sequentially throughout the monitoring campaign. Thus, to transfer a classifier trained using data from multiple damage-states in the source domain, transfer can only be performed using solely unlabelled target data, or a limited set of target labels which only represent a subset of all classes in the source, i.e $\mathcal{Y}_t \subseteq \mathcal{Y}_s$. As far as the authors are aware, Gardner \emph{et al.} \cite{gardner2022application} is the only example of supervised domain adaptation for PBSHM. In \cite{gardner2022application}, kernelised Bayesian transfer learning (KBTL), was applied to learn a shared classifier, and a shared feature space, across multiple structures with different feature dimensions. It was shown this approach could classify damage-states where there were no labels in that specific domain; however, the case studies assume most classes included labels. \\ 

Beyond proposing an active transfer learning framework for PBSHM, this paper also introduces a novel Bayesian DA model that uses a low-variance interpretable-probabilistic mapping, regularised via informed priors, while fully exploiting source data via a shared-flexible predictive model. While, performing DA via a joint classifier has been investigated in a few previous studies \cite{gonen2014kernelized, duan2012learning, hoffman2013efficient}, the proposed classifier has a unique set of properties that make it suitable for active transfer learning with sparse target datasets. In \cite{duan2012learning}, a shared feature space was found via a joint binary support vector machine (SVM). This approach differs from the proposed method since it cannot learn a shared space common to multiple classes, so it cannot be used to predict classes that have previously not been observed in the target domain. Hoffman \emph{et al.} proposed the first approach to learn a shared feature space common to multiple binary SVM classifiers, with the objective of predicting classes in the target domain which have not been previously labelled \cite{hoffman2013efficient}. However, this method requires both the mapping and classifier to be learnt in the same feature space (typically a nonlinear kernel basis), and is not a probabilistic model. The most similar approach is KBTL \cite{gonen2014kernelized}, which finds a projection into a latent space shared between multiple domains in a Bayesian framework. The main differences to the proposed approach is KBTL learns a nonlinear DA mapping (via a kernel mapping), and does not maintain the interpretability of the original feature space. This mapping is powerful in scenarios where feature dimensions differ between domains, or where dimensionality reduction is required. However, the flexibility of the mapping may not be suitable when target data are sparse and it does not result in an interpretable feature space, which may make defining an informative prior mapping challenging.\\

A few previous studies have demonstrated that active learning can reduce the number of labels required to train conventional machine-learning models for SHM. For example, generative mixture models have been used with a mixture of entropy- and likelihood-based \cite{bull2019probabilistic, bull2022sampling} and risk-based sampling strategies \cite{hughes2022risk}. In addition, to further reduce label requirements and mitigate the effects of sampling bias the combination of semi-supervised and active learning has been investigated \cite{bull2022sampling, hughes2022robust}, as well as the use of efficient discriminative classifiers (the relevance vector machine) \cite{hughes2022robust}. Uncertainty sampling has also been used with neural networks to classify images of defects \cite{chakraborty2015adaptive
} and in \cite{feng2017deep} a Bayesian convolutional neural network was used for tool monitoring. Finally, \cite{martinez2019towards} proposed a probabilistic framework for active sampling for a damage-progression model. However, active learning has only recently been considered in the context of PBSHM for multi-task learning (using hierarchical modelling) for regression \cite{clarkson2024active}, and has not been investigated in the context of classification or TL.\\

In reliability analysis, data are often queried based on the U-function \cite{echard2011ak, dai2025adaptive}. The U-function aims to select samples that are both uncertain and correspond to system states operating close to failure. In this sense, it is similar to risk-based active learning \cite{hughes2022risk}, as both approaches aim to prioritise data associated with decision-critical health states - i.e., if the system is identified to be close to failure, it should be shut down and inspected. While MES or likelihood-based sampling can be effective at improving model performance, they may lead to the labelling of data that have minimal effect on the decision-making process. An interesting future direction would be to extend the framework proposed in this work to incorporate such risk-based sampling strategies.\\

\section{Classifier-based Bayesian domain adaptation}

This section presents a novel Bayesian model for DA, the DA-relevance vector machine (DA-RVM). The DA-RVM is proposed to facilitate classification in both domains via a flexible classifier, learnt largely using source data, while defining a probabilistic mapping that can be defined using minimal parameters that can be regularised via informative priors. By considering uncertainty on the mapping, the model reduces confidence of classification of target data in comparison to directly applying a source classifier to the target data via a deterministic mapping, which is a may prevent missing informative target data in the active sampling process. The section following section outlines the proposed model, as well as methods used for defining prior mapping parameters and the active sampling scheme.\\

\subsection{Model assumptions}

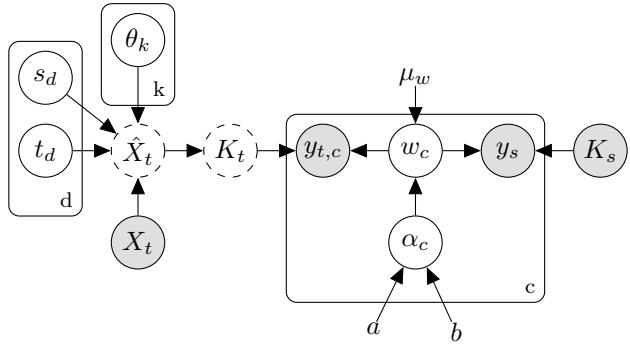
\begin{figure}[t!]
\centering
\begin{tikzpicture}

  \node[latent] (sd) at (0,0) {$s_d$};
  \node[latent, below=0.25cm of sd] (td) {$t_d$};

  \node[latent, dashed, right=0.5cm of td] (Xhat_t) {$\hat{X}_t$};
  \node[obs, below=0.5cm of Xhat_t] (Xt) {$X_t$};
  \node[latent, above=0.75cm of Xhat_t] (theta_k) {$\theta_k$};

  \node[latent, dashed, right=0.5cm of Xhat_t] (Kt) {$K_t$};
  \node[obs, right=0.5cm of Kt] (ytc) {$y_{t,c}$};
  \node[latent, right=0.5cm of ytc] (wc) {$w_c$};
  \node[latent, below=0.5cm of wc] (ac) {$\alpha_c$};
  \node[obs, right=0.5cm of wc] (ys) {$y_{s}$};
  \node[obs, right=0.5cm of ys] (Ks) {$K_{s}$};
  
  \node[const, above=0.5cm of wc] (wcp) {$\mu_w$};
  \node[const, below left=0.8cm and 0.2cm of ac ] (acp1) {$a$};
  \node[const, below right= 0.8cm and 0.2cm of ac] (acp2) {$b$};

  \edge {sd, td} {Xhat_t};
  \edge {Xt} {Xhat_t};
  \edge {theta_k} {Xhat_t};

  \edge {Xhat_t} {Kt};
  \edge {Kt} {ytc};
  \edge {wc} {ytc};
  \edge {ac} {wc};
  \edge {wc} {ys};
  \edge {Ks} {ys};
  \edge {wcp} {wc};
  \edge {acp1} {ac};
  \edge {acp2} {ac};

  \plate {dims} {(sd)(td)} {d};
  \plate {theta_plate} {(theta_k)} {k};
  \plate {class_plate} {(ytc)(wc)(ac)(ys)} {c};

\end{tikzpicture}
\caption{Graphical model representation of the proposed DA-RVM. Nodes correspond to variables: shaded nodes denote observed variables, solid outlines indicate random variables, and dotted outlines represent deterministic nodes. Arrows without a connected parent node indicate prior distributions. Plates represent replicates over dimensions for the mapping variables and classes for classifier weights.}
\label{fig:dgm}
\end{figure}

The model has two core components -- a classifier that is learnt using both source and (limited) target data, and a linear mapping that is applied to the target data to allow for data in both domains to be classified by a shared classifier. 
To achieve a high likelihood of classification in both domains using a single classifier, domain divergence must be low \cite{Ben-David2007}; as a result, the mapping aims to project target data into a feature space where distribution divergence between $p_s(y|\mathbf{x})$ and $p_t(y|\mathbf{x})$ is reduced. In addition, using a shared classifier to reduce distribution divergence does not rely on assumptions about the underlying generative process of the data or require nonparametric measures between the data distributions, which may be beneficial in sparse data scenarios \cite{hughes2022robust}. A graphical model depicting the proposed model is shown in Figure \ref{fig:dgm}. \\

In this paper, the main objective is to learn a mapping that minimises distribution divergence between the domains using minimal target data. Thus, the mapping is restricted to a linear transformation, which is decomposed into a scale $\mathbf{s} \in \mathbb{R}^d$, translation $\mathbf{t} \in \mathbb{R}^d$, and rotation \( \boldsymbol{\theta} = \{ \theta_{(i,j)} \}_{(i,j) \in \mathcal{I}} \in \mathbb{R}^m \), where \( m = \frac{1}{2}d(d-1) \) and \( \mathcal{I} \) is the set of all index pairs between the features \( (i,j) \) such that \( 1 \leq i < j \leq d \). Compared to more complex mappings found by many popular DA methods, the proposed mapping may require fewer data to estimate \cite{vapnik2013nature}, while the classifier can be defined to be more flexible as it is learnt using larger quantities of source data \footnote{It should be noted that the proposed mapping does imply a strong prior assumption about the form of the shift between domains; in many scenarios this assumption may be too strict and it could be relaxed by kernelising the data prior to finding the mapping or including additional transformation terms.}. The modelling assumptions for the mapping parameters are given by, 

\begin{equation}
    s_i \sim \mathcal{TN}(\mu_{s}, \sigma_{s}, a_{s}, b_{s}) \quad  \text{for } i = 1, \dots, d
\end{equation}
\begin{equation}
    t_i \sim \mathcal{N}(\mu_{t}, \sigma_{t}^2) \quad \text{for } i = 1, \dots, d
\end{equation}
\begin{equation}
    \theta_i \sim \mathcal{TN}(\mu_{\theta}, \sigma_{\theta}, a_{\theta}, b_{\theta})  \quad \text{for } i = 1, \dots, m
\end{equation}

\noindent where $\mathcal{TN}$ is the truncated normal distribution, with parameters $\mu_{s}$ and $\sigma_{s}$ representing the mean and standard deviation defining the prior for each scale component $s_i \in \mathbf{s}$; $\mu_{\theta}$ and $\sigma_{\theta}$ are the corresponding parameters for each rotation angle component $\theta_i \in \boldsymbol{\theta}$; and the upper and lower bounds are defined by $a_s < s_i < b_s$ for the scale parameters, and $a_{\theta} < \theta_i <b_{\theta}$ for the rotation parameters. The prior for the translation components $t_i \in \mathbf{t}$, is defined as a normal distribution $\mathcal{N}$, with mean $\mu_{t}$ and standard deviation $\sigma_{t}$.\\

The rotation angles, scale and translation are assembled into matrix form; where $S$ is the scale matrix, given by $S = \operatorname{diag}(\mathbf{s})$, $T$ is the translation matrix, given by $T = \operatorname{diag}(\mathbf{t})$, where $\operatorname{diag}$ denotes a diagonal matrix elements of a vector placed along its diagonal. The rotation matrix $\Theta$ is constructed using the following expression,
\begin{equation}
 \Theta = \prod_{(i,j) \in \mathcal{I}} G^{(i,j)}(\theta_{(i,j)}),
\end{equation}
where \( G^{(i,j)}(\theta_{(i,j)}) \in \mathbb{R}^{d \times d} \) is a Givens rotation matrix in the \((i,j)\)-plane by angle \( \theta_{(i,j)} \) \cite{golub2013matrix}. The transformation on the target features prior to classification can be expressed by, 
\begin{equation}
    \hat{\mathbf{X}}_t = \mathbf{X_t} \cdot \Theta^T \cdot S + T
\end{equation}
where $\hat{\mathbf{X}}_t$ denotes the transformed target features. \\

As the mapping projects target data into the source feature space, the interpretability of the original feature space is maintained, i.e. in structural terms, increases in natural frequency values can still be interpreted as a stiffness increase. In addition, decomposing the mapping in this way promotes interpretability of the mapping itself, allowing for engineering judgement to be used to define prior mapping values. For the specific case study in this paper, these parameters were restricted to improve learning with sparse target data. The features used were natural frequencies; thus, scale was restricted to be positive ($a_s=0$ and $b_s=\infty$ in equation (4)) to reflect the belief that for a given pair of natural frequencies, the relationship between the reduction in stiffness and damage will not reverse between domains -- damage is almost always expected to result in a reduction in natural frequencies \cite{FarrarC.R.CharlesR.2013Shm:}. Furthermore, natural frequencies were selected using the MAC to ensure they correspond to similar modes  (e.g. bending or torsion etc.); therefore, rotation was limited to $[-\frac{\pi}{4}, \frac{\pi}{4}]$ to both reflect the fact the relationship between stiffness does not invert, and features should not be reordered.\\

The classifier used in this model is a relevance vector machine (RVM), a sparse vector learner first proposed by Tipping \emph{et al.} \cite{tipping1999relevance}, and later extended to a multi-class setting in \cite{damoulas2008inferring}. In the RVM, data are projected into a reproducing kernel Hilbert space (RKHS) via a kernel embedding that represents each data point in terms of its similarity to other training data via a kernel function $k(\cdot)$, allowing for more expressive boundaries to be found using linear-in-the-weights models. The RVM also induces sparsity over the weights, whereby most weights are zero; therefore, effectively reducing the number of model parameters and reducing the likelihood of overfitting \cite{Murphy2014}. Thus, test data are classified given their similarity (via the kernel function) to the samples corresponding to non-zero weights; these samples are referred to as \emph{relevance vectors}. First data are projected into RKHS yielding a kernel matrix $K$, which is found by,  
\begin{equation}
    K = \left[ k(\mathbf{x}_i, \mathbf{x}_{s,j}) \right]_{i \in n,
    j \in n_s}
\end{equation}
where $n$ is the total number of labelled samples $n = n_s + n_{t,l}$. Note that the target data are projected into the kernel space after they are mapped to the source feature space via equation (8). Thus, the mapping remains linear, while the classifier can be specified as a flexible nonlinear classifier with a suitable kernel function. Here, relevance vectors are restricted to the source samples, which forces the mapping to align the data such that these relevance vectors are representative of both the source and target domains, implying that divergence between the domains must be low. This choice was to prevent the potential solution of finding domain-specific relevance vectors, which may lead to an arbitrary mapping, as target data must be classified based on their similarity to the set of source relevance vectors. \\

The probabilistic modelling assumptions for the classifier are given by,
\begin{equation}
    w_{c,i} \mid \alpha_{c,i}  \sim \mathcal{N}(0, \alpha_{c,i}^{-1}) \quad \text{for } i = 1, \dots, n_s
\end{equation}

\noindent where $\mathbf{w}_{c} \in \mathbb{R}^{n_s}$ is a vector of weights with components defined as $w_{c,i} \in \mathbf{w}_{c}$, for class $c \in \{1, \dots, C\}$, with $C$ classes. Prior precision values are specified by $\boldsymbol{\alpha}_{c} \in \mathbb{R}^{n_s}$, where each component $\alpha_{c,i} \in \boldsymbol{\alpha}_c$ is assumed to be conditionally independent of the others. The precision values for each component are assumed to follow a Gamma distribution $\Gamma(\cdot)$, as follows,

\begin{equation}
    \alpha_{c,i} \sim \text{Gamma}(a, b) \quad \text{for } i = 1, \dots, n_s
\end{equation}

where are $a$ and $b$ shape and rate parameters. By specifying $a$ and $b$ so the gamma distribution results in a weakly-informative prior on the precision, this prior promotes large precision values $\alpha_{c,i}$, driving the corresponding weights $w_{c,i}$ towards zero under the posterior. As a result, while the $n_s$ weights are estimated, the effective number of parameters should be much lower; more details on this hierarchical prior structure for inducing sparsity can be found in \cite{tipping1999relevance}. These weights are used to obtain the class membership scores $\gamma_c$ for class $c$ by,
\begin{equation}
    \gamma_c = \mathbf{k}_{i} \mathbf{w}_c^T \label{eq:logit} 
\end{equation}
 $\mathbf{k}_i \in K$ represents a kernelised sample. To achieve multi-class classification, the softmax function normalises the unbounded membership values into a valid categorical distribution over the classes; therefore, class probabilities are given by,
\begin{align}
P(y = c \mid \mathbf{k}_i) &= \frac{e^{\gamma_c}}{\sum_{j=1}^{C} e^{\gamma_j}} \label{eq:softmax}
\end{align}
\noindent Normalising the membership scores in this way allows a categorical likelihood function to be used. Since both source and target data are used to learn a classifier, this model is also related to multi-task learning \cite{gonen2014kernelized}, which mainly differs from TL in that it aims to equally improve the performance across multiple domains, instead of prioritising the target domain. \\

\begin{figure}[b!]
    \centering
        \includegraphics[width=0.8\textwidth]{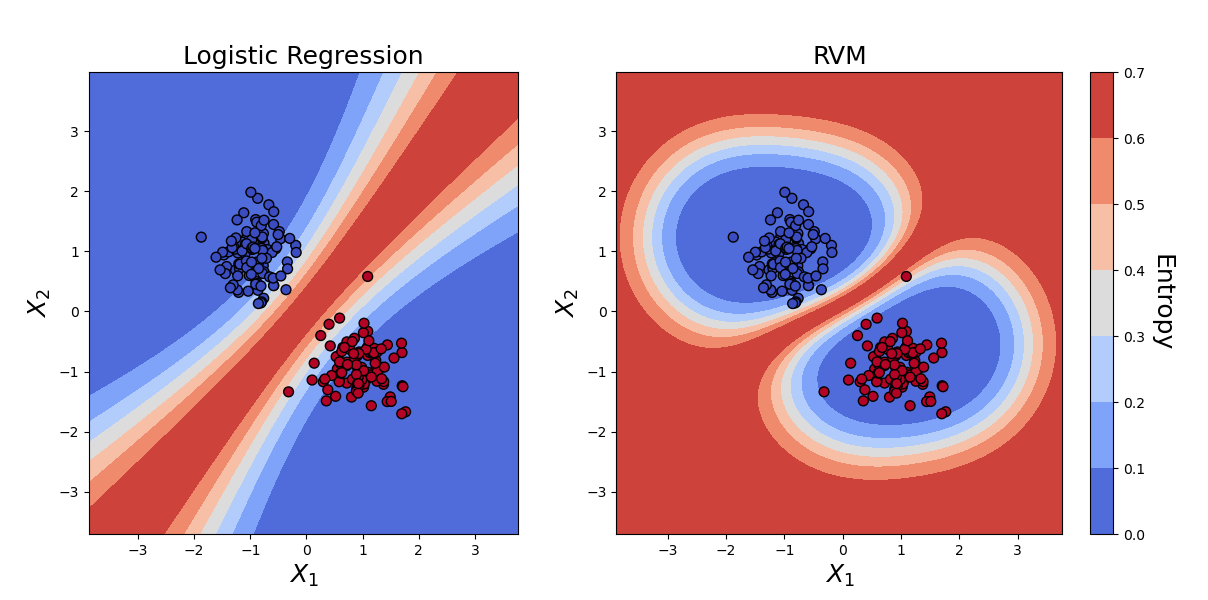}
    \caption{Toy example with shaded regions showing the entropy in label predictions produced by a Bayesian logistic regression model (left) and an RVM (right).}
    \label{fig:rvm_log}
\end{figure}

While there are a variety of suitable classifiers that could be used in this framework, the RVM was chosen for three core reasons. First, the RVM is a flexible nonparametric classifier which has been shown to learn efficiently with sparse datasets \cite{tipping1999relevance, damoulas2008inferring}. The RVM produces tight decision boundaries, and when a Gaussian kernel is used, prediction probabilities converge to a uniform distribution as test points move further from the relevance vectors because of their similarity to the relevance vectors approaching zero, i.e. $k(\mathbf{x}_i, \mathbf{x}_{s,j}) \approx 0 \quad \forall j \in \{1, \dots, n_s\}$. To demonstrate the behaviour of the RVM, a toy example showing the entropy for a Bayesian logistic regression model and an RVM is presented in Figure \ref{fig:rvm_log}. From the perspective of transfer, high classification likelihood would be achieved when data are mostly distributed in low entropy regions, on the correct side of the classification boundaries; thus, more restrictive
boundaries restrict the possible mappings significantly -- potentially leading to better alignment when only a few classes are labelled in the target domain. In the example in Figure \ref{fig:rvm_log}, for the RVM, it can be seen the region of low entropy is significantly smaller compared to logistic regression. There is also an additional benefit when considering this model for MES. As previously discussed, MES typically leads to sampling near the boundaries, but often will not sample novel data at the extremities of the model. However, the RVM only has high confidence in data near the relevance vectors, as can be seen in Figure \ref{fig:rvm_log}; thus, it will also assign high entropy to observations at the extremities of the model, combining the benefits of both MES and low-likelihood sampling \cite{hughes2022robust}.\\

The model was implemented in a general-purpose probabilistic programming language -- \texttt{numpyro} \cite{bingham2019pyro}. The posterior over the weights if intractable because of the categorical likelihood and (potentially) nonlinear transformation via the kernel embedding. Thus, the parameters of the model are inferred via MCMC using the no-U-turn  (NUTS) implementation of Hamiltonian Monte Carlo \cite{hoffman2014no}. The parameters for the RVM were initialised using only the source data with the RVM$_2$ expectation-maximisation algorithm outlined in \cite{damoulas2008inferring}. Weights with values below $10^{-5}$ were pruned from the initial model to reduce the computational complexity of learning this model via sampling. \\

While HMC via NUTS enables approximation of the full posterior -- given a sufficient sample size -- it may face limitations when scaling to high-dimensional problems. To address this limitation, alternative inference strategies may be more appropriate. For example, variational inference or an expectation-maximisation scheme could offer more efficient inference, while importance sampling may allow parallel computational resources to be leveraged more effectively \cite{Murphy2014}. Investigating these approaches represents an interesting direction for future work.\\

Nevertheless, the current approach remains applicable to many SHM scenarios, which often rely on a small number of damage-sensitive features - such as natural frequencies - as demonstrated in the case study presented in this paper. Moreover, since the aim of the proposed method is to update the DA-RVM online as new labels are acquired, reducing the computational cost of inference would also help minimise ongoing resource demands. However, in many SHM scenarios, labels are acquired only periodically; therefore, the DA-RVM would also only require periodic retraining, which could feasibly be performed offline before redeployment for online monitoring.

\subsection{Inferring a prior mapping with distribution alignment} 

In practice, to learn a discriminative classifier, it is required that the underlying conditional distributions of the training and testing data are the same, i.e. $p_s(y|\mathbf{x}) = p_t(y|\mathbf{x})$. However, the lack of labels and limited samples of data means that learning a mapping that directly aligns the conditional distributions is often not possible. As previously discussed, unsupervised DA generally assumes that the underlying conditional distributions can be aligned by minimising a distribution distance metric between a sample of data -- often these approaches aim to minimise marginal-distribution distance, assuming labels are unavailable or sparse \cite{Zhuang2021}. Such mappings will only result in invariant conditional distributions if both the domains are sufficiently related, and there is a suitable DA method to find a mapping with available data \cite{wang2019characterizing}. In addition, testing the outcomes of transfer is challenging, as in many cases labelled data will not be sufficient in the target domain to perform conventional validation, such as cross-validation. As such, current approaches to DA must be applied to testing data prior to full validation.\\

Determining when these assumptions apply, without traditional model validation, is a critical challenge for the practical application of DA. Without direct validation, assessing the prediction quality on target test data is challenging, highlighting the importance of research into validation and prediction of transfer outcomes for PBSHM \cite{hughes2024quantifying}. This paper proposes that these mappings can be treated as a ``prior mapping'' in the DA-RVM, where prior uncertainty is reflect the trust in unsupervised DA\footnote{In practice, defining this prior uncertainty may depend on similarity assessment performed between domains and engineering judgement.}.In this way, assumptions made when estimating the initial DA mapping can be considered as a prior, where the posterior mapping parameters are updated with data directly relating to the quantity of interest -- the likelihood of classification in the target domain. A consequence of the Bayesian formulation is, as labelled data become more abundant in the target, the posterior mapping becomes less influenced by the prior mapping, meaning it relies less on the strict assumptions made by unsupervised DA.\\

In this paper, the mapping is composed of scale, translation and rotation parameters. Thus, an appropriate set of DA techniques would be \emph{statistic alignment}, a branch of DA which aim to directly align the lower-order statistics \cite{Zhuang2021}. Since engineering datasets are prone to class imbalance, a method called \emph{normal condition alignment} (NCA) was used \cite{poole2023statistic}. NCA was used to mitigate issues related to class imbalance by selecting a subset of data from the beginning of the monitoring campaign, where it is typically assumed that the structure was undamaged and is a standard practice in damage detection \cite{FarrarC.R.CharlesR.2013Shm:, dervilis2014damage}. This method first standardises the source domain, and the normal conditions are then aligned by, 
\begin{equation}   \boldsymbol{z}_{t,i}=\left(\frac{\boldsymbol{x}_{t,i}-\boldsymbol{\mu}_{t,n}}{\boldsymbol{\sigma}_{t,n}} \right)\boldsymbol{\sigma}_{s,n}+\boldsymbol{\mu}_{s,n}
\end{equation}
\noindent where $\boldsymbol{z}_{t,i}$ is the transformed target data, $\boldsymbol{\mu}_{s,n}$, $\boldsymbol{\mu}_{t,n}$ and $\boldsymbol{\sigma}_{s,n}$, $\boldsymbol{\sigma}_{t,n}$ are the means and standard deviations of the normal-condition data for the source and target respectively.  Using NCA, the prior scale mean parameters (equation 4), can be defined by $\boldsymbol{\mu}_s = \frac{\boldsymbol{\sigma}_{s,n}}{\boldsymbol{\sigma}_{t,n}} $, and the prior mean translation parameters (equation 5), as $\boldsymbol{\mu}_t = \boldsymbol{\mu}_{s,n}-\boldsymbol{\mu}_{t,n}\frac{\boldsymbol{\sigma}_{s,n}}{\boldsymbol{\sigma}_{t,n}}$. In this paper, prior rotation was assumed to be zero; however, methods such as correlation alignment \cite{Sun2015}, or Procrustes analysis \cite{conti2023physics} could be used to define a prior rotation.\\

While in some cases these approaches may lead to better generalisation by regularising the mapping, if these mappings are not appropriate for a specific transfer problem, they could still cause negative transfer. 
In general, prior mappings should only be used when they are suitable for a given transfer task, highlighting the need for similarity-quantification methods -- an area of ongoing research \cite{Gosliga2021}. Nevertheless, NCA was used in this case as it has been shown to perform successful transfer in a number of previous case studies in PBSHM \cite{poole2023statistic, giglioni2024domain}.\\

\subsection{Active sampling scheme}

\begin{figure}[h!]
    \centering
\includegraphics[width=0.8\textwidth]{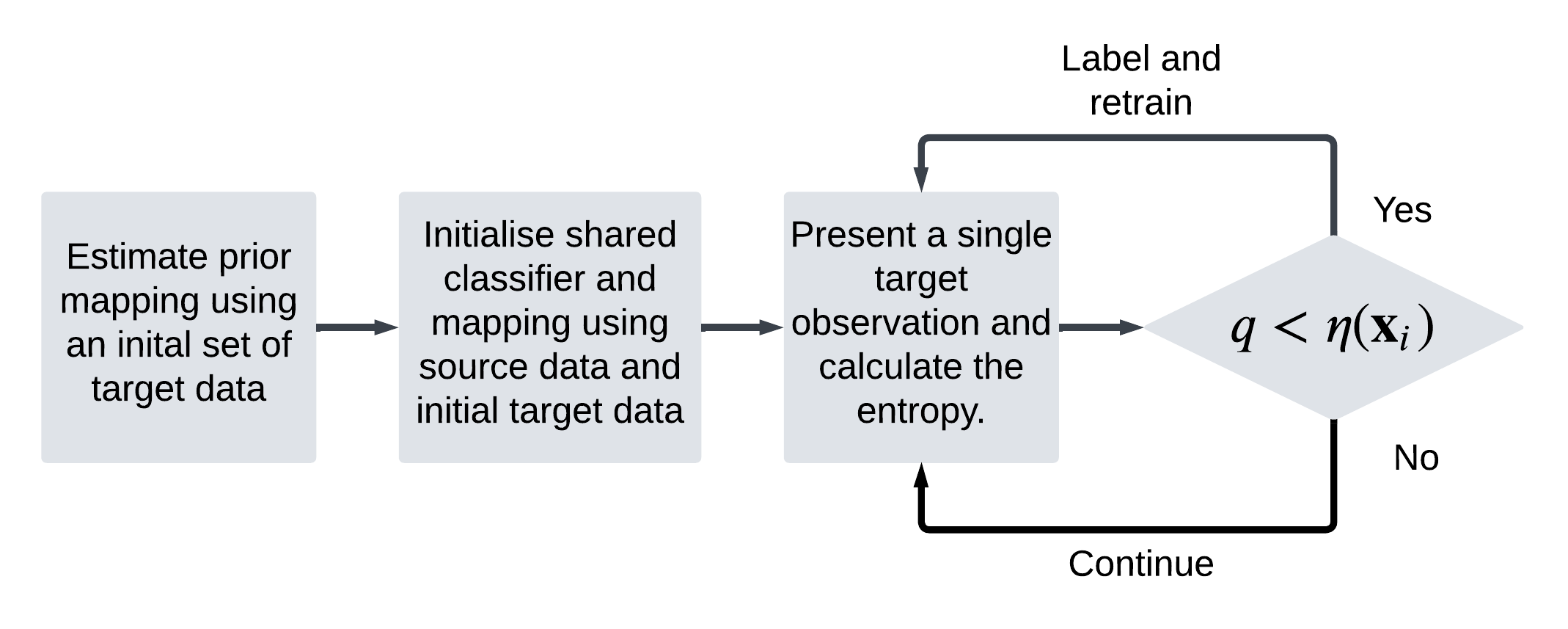}
    \caption{Flow chart to illustrate the active-learning process with DA.}
    \label{fig:flow}
\end{figure}

While incorporating labels into a DA framework may be beneficial, it is pertinent that the number of samples are minimised to reduce the associated cost of the monitoring system. To this end, an active-sampling strategy is proposed to ensure that the most informative data are labelled. This paper utilises a MES strategy first proposed in \cite{bull2022sampling}. Sampling is performed in a stream-based setting following the procedure outlined in Figure \ref{fig:flow}. \\

To decide when to query a sample, first entropy is obtained for test data using equation (3). To constrain the mapping prior to the acquisition of damage labels, the classifier is trained to discriminate between ambient and freezing undamaged data, where data collected above $0\degree C$ was considered to be generated under ``ambient'' conditions and below to be ``freezing''. To prevent unnecessary labelling at this boundary the probabilities of these classes are combined. In general, different weightings could be given to each class based on their importance to decision making \cite{hughes2021probabilistic}. The information efficiency \cite{mackay2003information} is then used to normalise entropy between zero and one,

\begin{equation}
    \eta (\mathbf{x}_i) = \frac{H(\hat{y}_i)}{\log(C)}
\end{equation}

The information efficiency  $\eta (\mathbf{x}_i)$, reflects the confidence in the label prediction compared to a uniformly distributed label prediction. Following \cite{bull2022sampling}, $\eta (\mathbf{x}_i)$ can be treated as a pseudo-probability that observation $i$ should be labelled. An observation is then labelled if a random draw $q$ from a uniform distribution $q \sim  \mathcal{U}(0,1)$ is less than $\eta (\mathbf{x}_i)$. Since the probability of sampling any observation will never be zero, this sampling scheme provides some protection against sampling bias.\\

By considering uncertainty on the DA mapping, the proposed DA-RVM model has an important advantage for MES in comparison to approaches that find deterministic mappings. As a result of mapping uncertainty, prediction uncertainty will generally be higher in the target domain to reflect the reduced trust in predictions caused by the reliance on source data. In the context of MES, accounting for this source of uncertainty will also have implications on the sampling procedure, potentially preventing overconfident predictions leading to ignoring informative target data.

\section{Transfer between lab-scale bridges}

This section presents an experimental dataset collected to investigate the active transfer learning approach for damage classification using a population of lab-scale beam and slab bridges. Specifically, data for three bridges with varying span lengths were obtained across changing temperatures, and the same four pseudo-damage states. \\

The inspection and maintenance for populations of bridges presents a major challenge, and there are signifiant safety concerns as bridges are operated towards the end of their design life. In addition, the scale and cost of these structures will often limit available SHM data to streaming data obtained throughout the operation of the structure. While it is uncommon for two bridges to have a nominally-identical design, there exists many examples of large heterogeneous populations with slight variations in geometry (i.e. with different lengths and support locations), managed by a single asset manager. For example, the main highways agency in the UK, National Highways, was responsible for managing 9,392 bridges in 2020 \cite{HighwaysEngland2020}. This motivates the application of the proposed active transfer-learning framework to bridge monitoring applications.\\

\subsection{Experimental dataset}

\begin{figure}[h!]
    \centering
    \begin{subfigure}{0.7\textwidth}
         \centering
        \includegraphics[width=\textwidth]{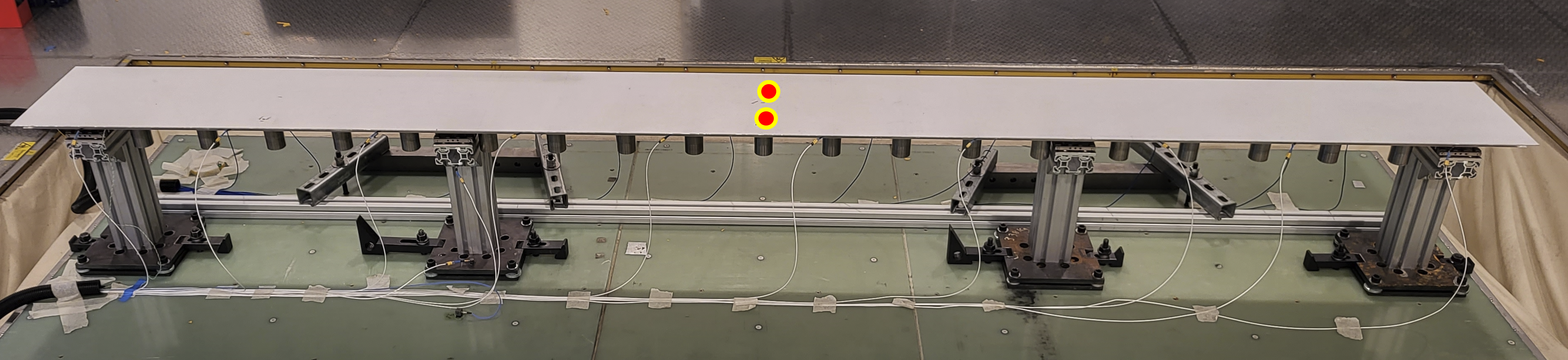}
         \caption{}
     \end{subfigure}
    \begin{subfigure}{0.285\textwidth}
         \centering
         \includegraphics[width=\textwidth]{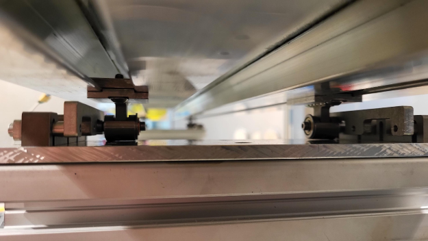}
         \caption{}
     \end{subfigure}
    \caption{The experimental set-up to perform modal testing for one configuration (B1), showing the full bridge (a) and the connection between the deck and supports via roller bearings (b). The red dots in (a) indicate the locations masses were added. }
    \label{fig:lab bridge}
\end{figure}

A population of three bridges, each with three spans, was constructed using a bespoke modular bridge kit that facilitates changing of the deck length, number of supports and support locations; thus, allowing for controlled variation between structures\footnote{All testing was conducted at the laboratory for verification and validation (LVV) at the University of Sheffield.}. Figure \ref{fig:lab bridge}(a) presents an example of one of the configurations used in these experiments. The kit consist of a set of four supports and a deck, supported by two I-beams, which are connected via a pair of roller bearings at each support, shown in Figure \ref{fig:lab bridge}(b). The bearings at one end are locked, such that they behave as pin joints. The location of the supports was varied between bridges to produce a heterogeneous population. The locations of the supports for each bridge are presented in Table \ref{tab:geo}. The bridges are referred to as B1, B2, and B3; this abbreviation will be used for the remainder of the paper. \\

\begin{table}[h!]
\caption{Summary of the configuration for each experimental bridge structure. Support locations indicate the position of the bearings connecting the deck and the supports.}
\centering
\setlength{\tabcolsep}{4pt}
\begin{tblr}{
  hline{1-2} = {-}{},
}
           & Deck length (m) & \shortstack{Support 1\\location (m)} & \shortstack{Support 2\\location (m)} & \shortstack{Support 3\\location (m)} & \shortstack{Support 4\\location (m)} \\
B1          & 3.00     & 0.14       & 0.725        & 2.28        & 2.86       \\
B2 & 3.00     & 0.14       & 0.82        & 2.19        & 2.86      
 \\
B3 & 3.00     & 0.14      & 0.86       & 2.15       & 2.86    
\end{tblr}
\label{tab:geo}
\end{table}

The bridges were attached to a six-axis shaker table via bolts at the base of each support, within an environment chamber. The bandwidth of excitation from the shaker table is approximately 90Hz; therefore, a set of masses were uniformly distributed along the underside of the deck, shown in Figure \ref{fig:lab bridge}(a), to reduce the natural frequencies and aid modal identification. Modal testing was conducted by applying a continuous white-noise random excitation, via the shaker table. Data were collected via twenty uniaxial 100 mV/g accelerometers, organised in two rows of ten on each edge of the underside of the deck, and the response was measured at a sample rate of 256Hz.\\

To investigate challenges presented by changing environmental conditions, the first two bridges, B1 and B2, were subjected to a range of temperature effects; B3 was only tested at ambient temperatures. Specifically, the response of the bridges was measured across two temperature cycles: from 15°C down to -15°C for B1, and from 15°C to -5°C for B2. A thermocouple was attached to the deck surface to monitor its temperature. To emulate a bi-linear stiffness relationship, which can be observed in concrete bridges \cite{maeck2001damage}, a fabric sheet was attached to the surface of the deck and saturated with water for the second temperature cycle, such that when frozen, its stiffness would sharply increase. Data were also acquired at ambient temperatures (between 23$\degree$C and 31$\degree$C), as well as for four pseudo-damage states which correspond to two masses (damage extents), 21.6g and 64.4g masses, placed in the centre of the central span in two locations, indicated by the red circles in Figure \ref{fig:lab bridge}. These masses were chosen as they represent a relatively small change in mass in comparison the the deck, which had a mass of 30kg\cite{papatheou2010use}.\\

To obtain natural frequencies for use as features, output-only modal analysis (OMA) was performed using covariance stochastic-subspace identification (SSI); natural frequencies were extracted, based on a reference set, selected via an automated pole-selection algorithm, using the software presented in \cite{garcia2020mova}. Several samples were unidentified and a few experimental outliers were removed from the normal condition data. Table \ref{tab:samps} shows the number of samples per class for each dataset following modal analysis. As is often characteristic of SHM datasets, the data are imbalanced, with larger quantities of undamaged data. \\

The full experimental dataset is openly available. For more details, the interested reader may refer to \cite{Poole2024, giglioni2025transfer}.

\begin{figure}[h!]
    \centering
    \begin{subfigure}{0.32\textwidth}
         \centering
        \includegraphics[width=\textwidth]{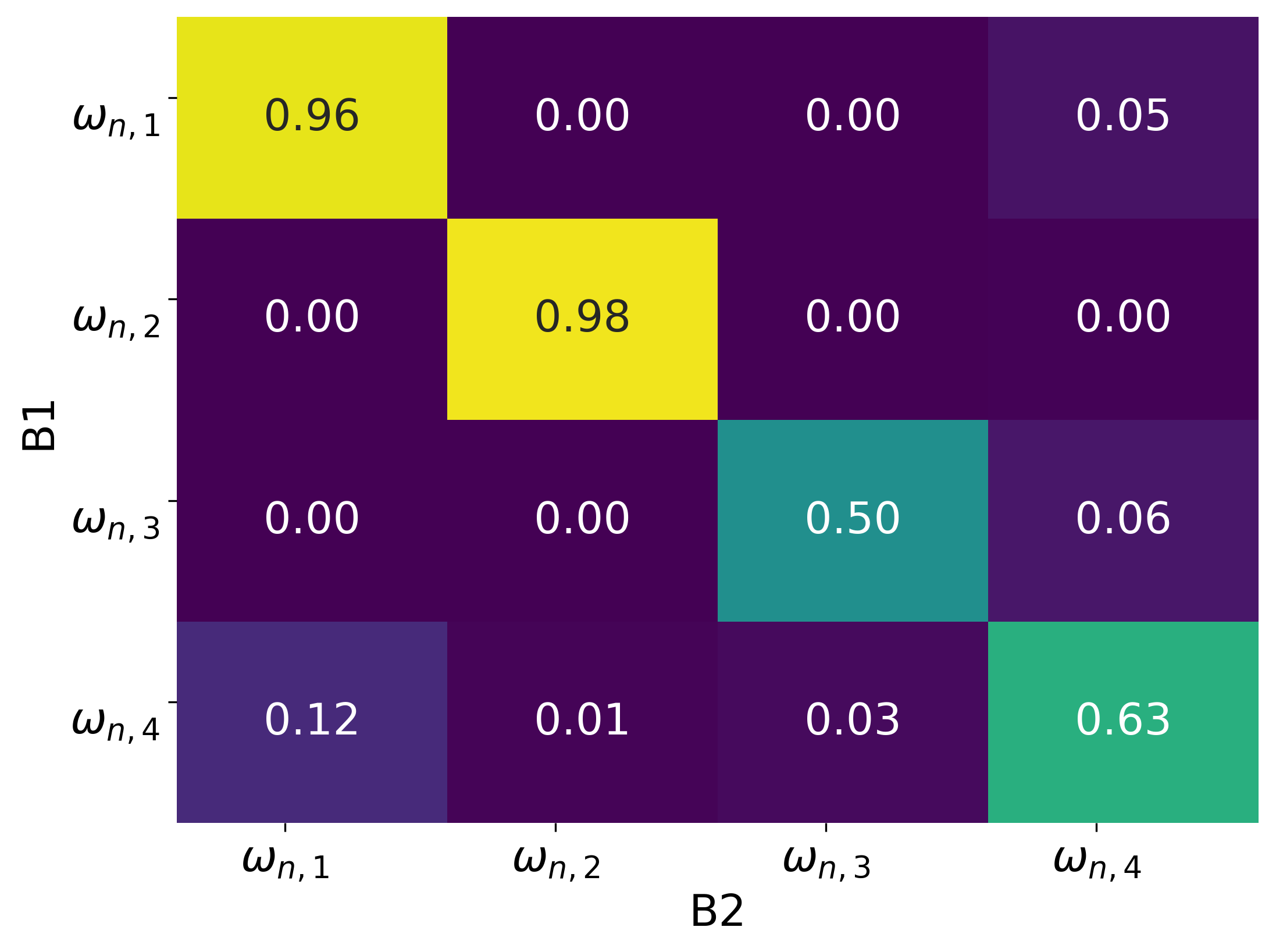}
         \caption{}
     \end{subfigure}
    \begin{subfigure}{0.32\textwidth}
         \centering
         \includegraphics[width=\textwidth]{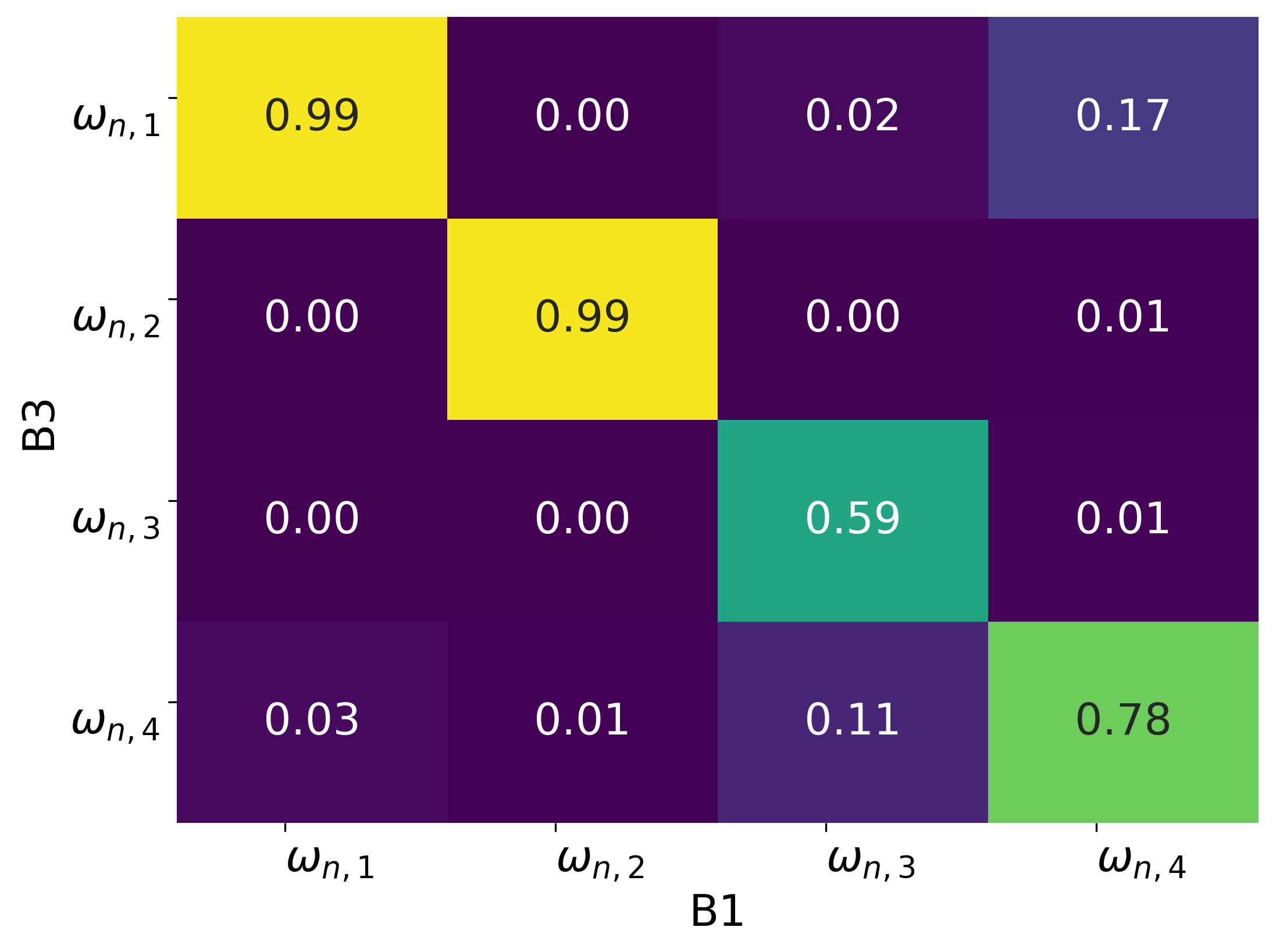}
         \caption{}
     \end{subfigure}
     \begin{subfigure}{0.32\textwidth}
         \centering
         \includegraphics[width=\textwidth]{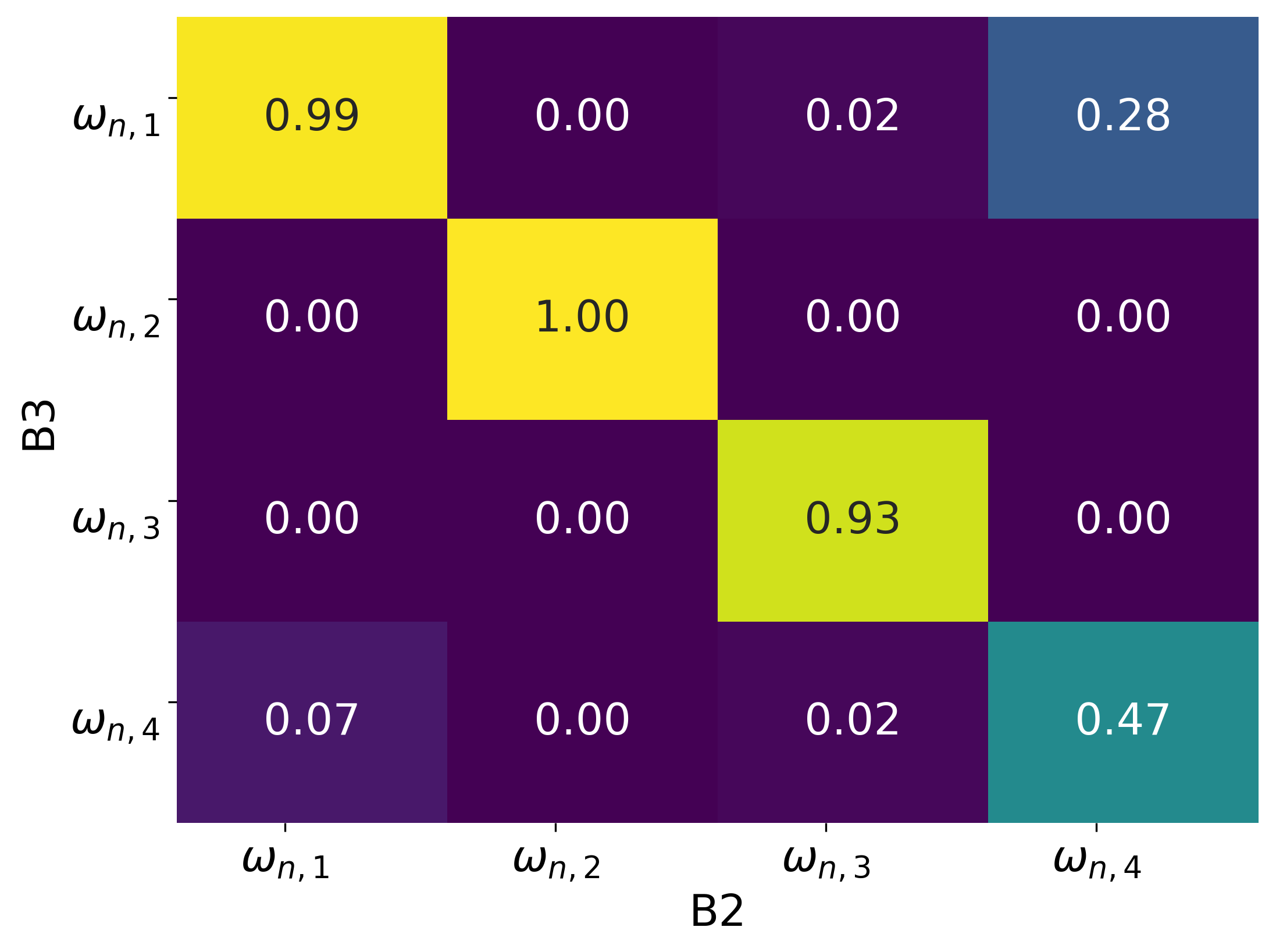}
         \caption{}
     \end{subfigure}
    \caption{Visualisation of the MAC scores between each pair of structures used as a source/target pair. }
    \label{fig:MAC}
\end{figure}

\begin{table}
\caption{Number of samples available after SSI per class for each bridge.}
\centering
\begin{tblr}{
  hline{1-2} = {-}{},
}
            & Ambient & Freezing & 21.6g off-centre & 21.6g centre& 64.4g off-centre & 64.4g centre\\
B1          & 179     & 138      & 10       & 9       & 10  & 10    \\
B2 & 129     & 54       & 5       & 5        & 7    & 10
 \\
B3 & 36     & 0       & 5       & 5        & 10    & 10
\end{tblr}
\label{tab:samps}
\end{table}

\subsection{Transfer tasks and methodology}

Four transfer tasks were investigated in this paper, which are split into two case studies. The first case study investigates transfer between structures under changing temperatures using the B1 and B2 datasets, considering each structure as a source and target, resulting in two transfer tasks; there tasks will be referred to as B1$\rightarrow$B2 and B2$\rightarrow$B1. The second case study investigates transfer from datasets with comprehensive temperature data and a target with limited data, considering B1 and B2 as source domains, and using B3 as a target, resulting in another two transfer tasks, referred to as B1$\rightarrow$B3 and B2$\rightarrow$B3. \\

The results from modal analysis were used to select features for transfer via the modal assurance criterion (MAC) \cite{allemang2003modal}, following \cite{gardner2022domain, poole2023towards}. The MAC matrices for the first four identified natural frequencies are presented in Figure \ref{fig:MAC}. In this paper, the first two natural frequencies were selected as features, as they have high MAC scores for each pair of structures; however, the third mode also has a high MAC value between B2 and B3, and may be utilised for transfer in future work.  \\

In each transfer task, the objective was to transfer a damage classifier capable of predicting the normal condition, and the four mass-states. The location of the 21.6g masses was not discriminative using the identified natural frequencies; thus, these two locations were considered as a single class, resulting in three damage classes. Furthermore, to constrain the mapping in the initial model (before damage is observed), the healthy data were split by ambient ($T > 0\degree C$) and freezing temperatures ($T < 0\degree C$). Therefore, the classifier was trained to discriminate between five classes - ambient and freezing normal condition data, pseudo-damage caused by adding a 21.6g mass (damage 1) and pseudo-damage resulting from a 64.4g mass placed off-centre (damage 2) and at the centre (damage 3) of the central span.\\

\begin{figure}[h!]
    \centering
    \begin{subfigure}[b]{\textwidth}
         \centering
        \includegraphics[width=0.62\textwidth]{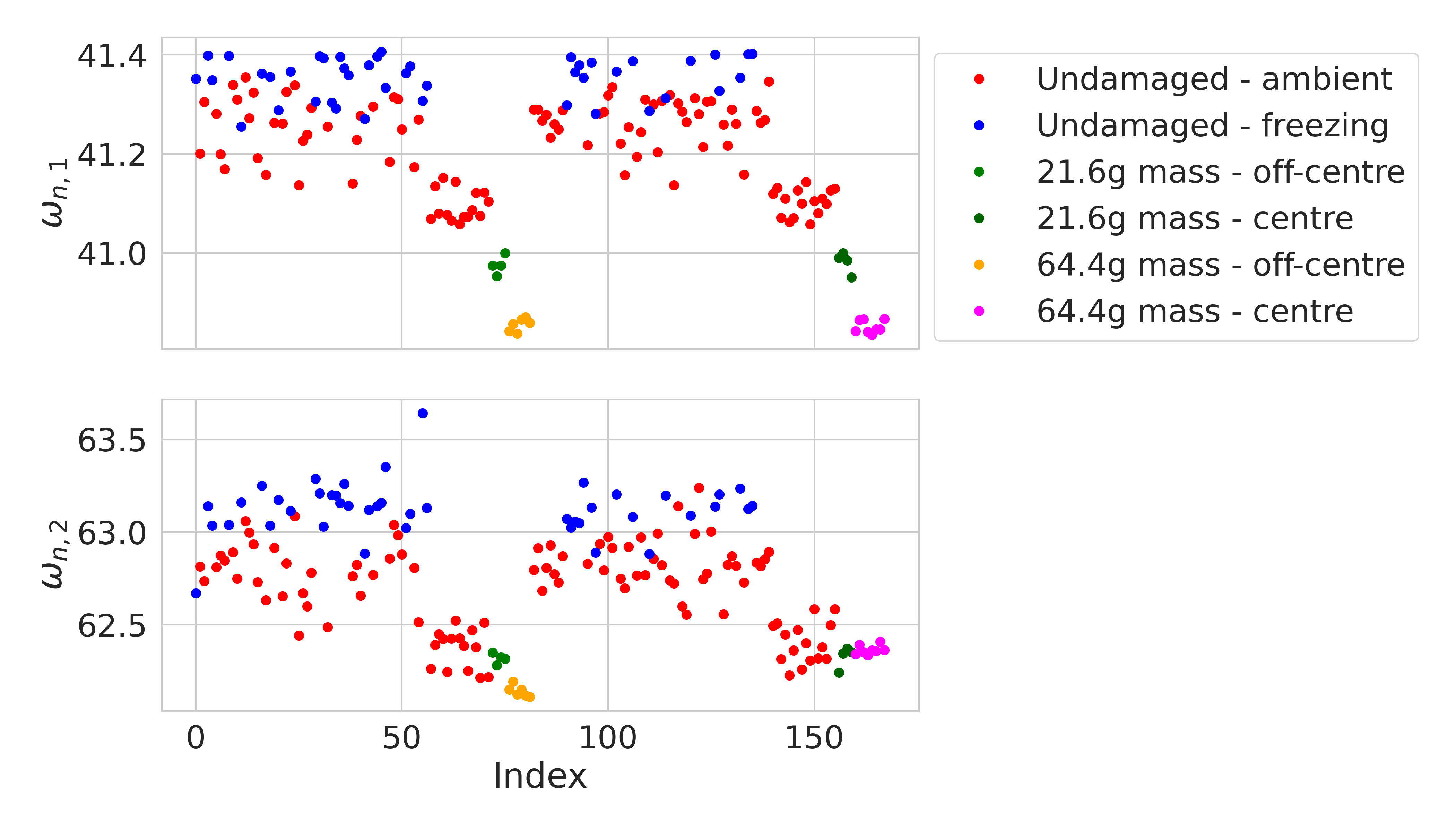}
         \caption{}
     \end{subfigure}
    \begin{subfigure}[b]{\textwidth}
         \centering
         \includegraphics[width=0.62\textwidth]{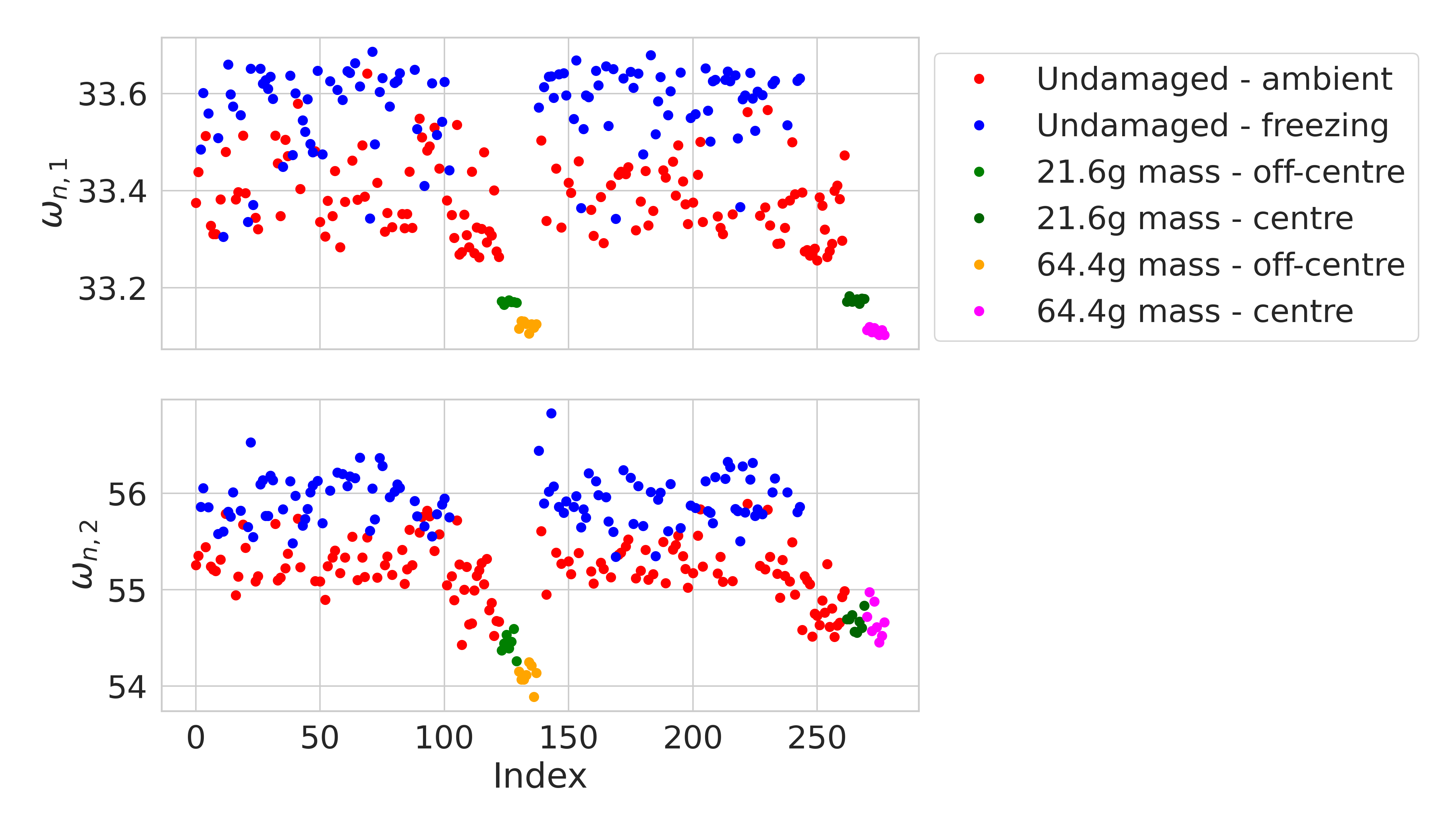}
         \caption{}
     \end{subfigure}
     \begin{subfigure}[b]{\textwidth}
         \centering
         \includegraphics[width=0.62\textwidth]{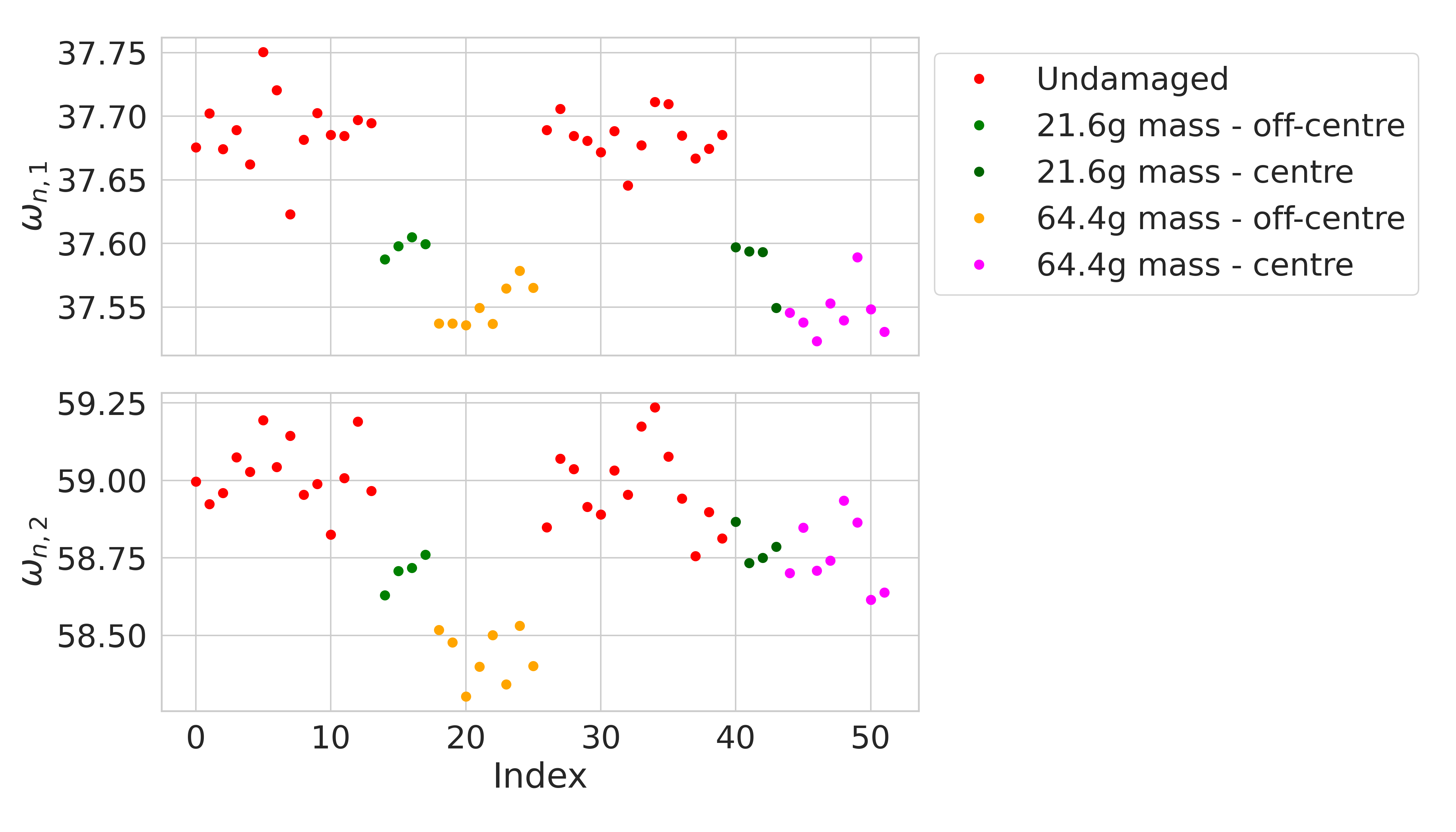}
         \caption{}
     \end{subfigure}
    \caption{Example of the ordered training data used for the active sampling process when B1, B2, and B3 are considered as target domains, presented in (a), (b) and (c) respectively.}
    \label{fig:example train}
\end{figure}

To emulate an active-sampling process for SHM, with the structure's state gradually transitioning from undamaged to damaged states, the target data were presented to the model as follows. First, data were split into training and testing datasets at a ratio of 80:20 using stratified sampling to ensure that the proportion of damage and undamaged data was consistent; the dataset was randomly shuffled and 100 training/testing datasets were generated in this way to test for differences in initial data used to learn the NCA mapping and the effect of presenting streaming data in different orders. The damage-states were organised into two damage scenarios, where in a single location, damage is initialised with minor damage (the 21.6g mass) and progresses to more severe damage (the 64.4g mass). Data were then ordered to present undamaged data collected at changing temperatures, followed by undamaged data collected under ambient conditions, subsequently by a damaged scenario; thus, each target domain includes two cycles of normal condition data, followed by a damage scenario. Figure \ref{fig:example train} presents an example of a single repeat of the training data (the first two natural frequencies), for each target domain considered. It can be seen that the expected range of values for both the first and second natural frequencies do not overlap between domains, motivating the application of mapping-based DA for transfer. \\

At the start of the active sampling process, the model was initialised using the source training dataset and only a subset of target training data, representing data corresponding to the undamaged structure. Following \cite{poole2023statistic}, in each case, the initial data used for NCA was selected to correspond to similar temperatures. Specifically, for B1$\rightarrow$B2 and B2$\rightarrow$B1, 70 initial data were used, and the ambient undamaged class was learned to learn the NCA mapping, as B1 contains data corresponding to lower temperatures. For B1$\rightarrow$B3 and B2$\rightarrow$B3, only 14 initial data were used as there were fewer normal condition data. In addition, data in B3 were only collected at room temperature; therefore, NCA was learn using data above 23$\degree$C in both domains. \\

The remaining data were presented sequentially, being labelled using the probabilistic sampling strategy discussed in Section 3.3; uncertainty between the ambient and freezing classes was not considered to prevent unnecessarily sampling normal condition data. Since there is significant class imbalance, the macro F1-score was used to assess classification performance on the entire test data, which included data from each class (for details see \cite{Gardner2020}).\\

The Gaussian kernel was utilised in the RVM as it is flexible and well-studied \cite{Murphy2014, damoulas2008inferring}. Following  \cite{damoulas2008inferring, hughes2022robust}, the bandwidth was defined as $\frac{1}{d}$. To demonstrate the benefits of incorporating transfer into the active-learning procedure, results were obtained for an active learner, using the RVM$_2$ algorithm from \cite{damoulas2008inferring}, trained solely with target data. Since the target-only RVM requires multiple target classes to be initialised, three random samples from the first damage scenario were selected to initialise the classifier. The specification of variances on the prior mapping parameters reflect the confidence in the NCA mapping, given the initial quantity of data. Thus, variances for all mapping parameters were set as $\sigma_t = \sigma_s = \sigma_{\theta} = 0.1$ in the first case study, and the variance for translation and scale were increased to $\sigma_t = \sigma_s = 1$ for the second case study, since only very few data were used to learn the NCA mapping, leading to large discrepancies in mean and scale between domains, as discussed in Section 4.5.   

\subsection{Case study: active transfer learning under changing temperatures}

\begin{figure}[t!]
    \centering
    \begin{subfigure}[b]{0.8\textwidth}
         \centering
        \includegraphics[width=\textwidth]{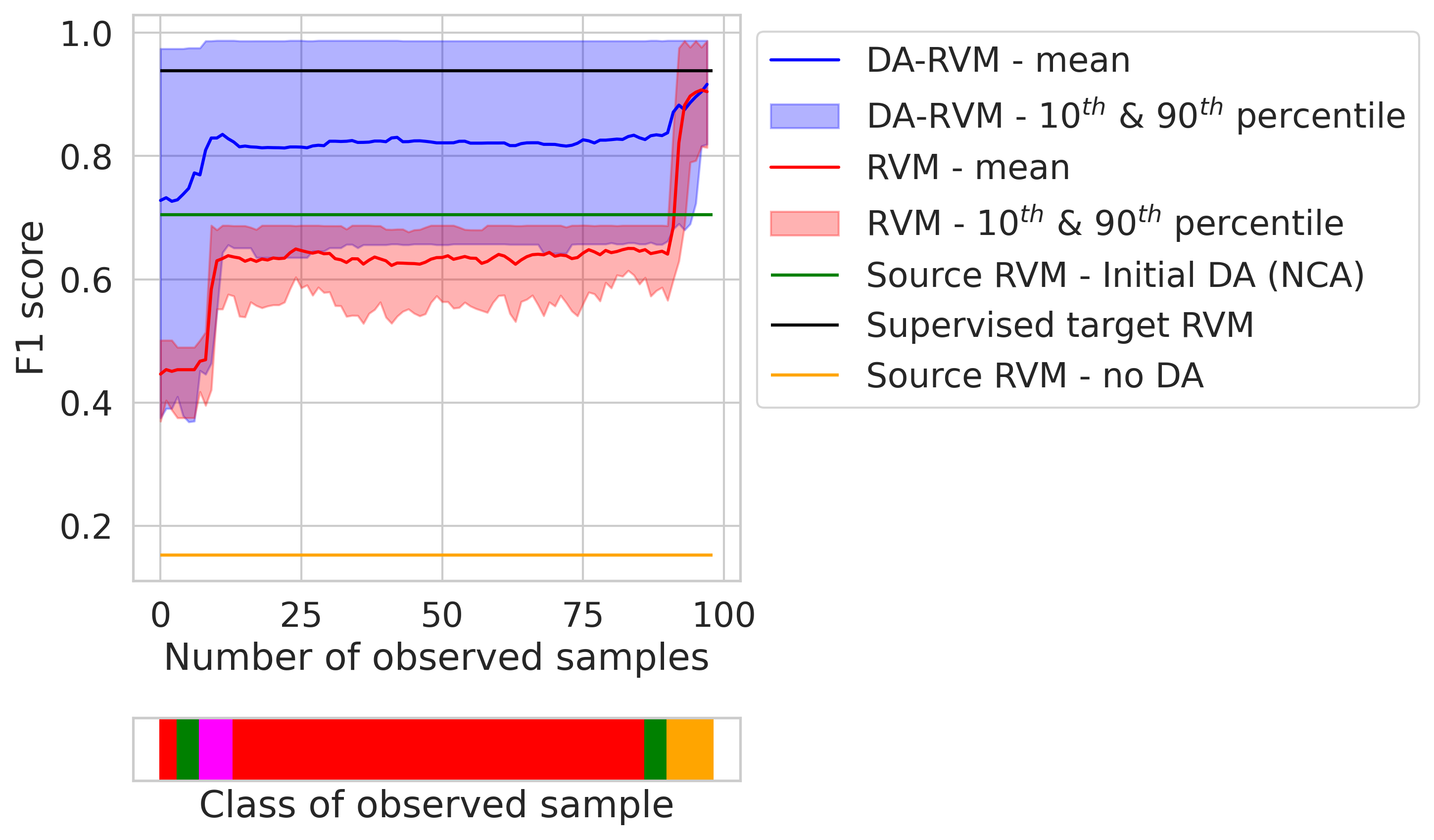}
         \caption{}
     \end{subfigure}
    \begin{subfigure}[b]{0.8\textwidth}
         \centering
         \includegraphics[width=\textwidth]{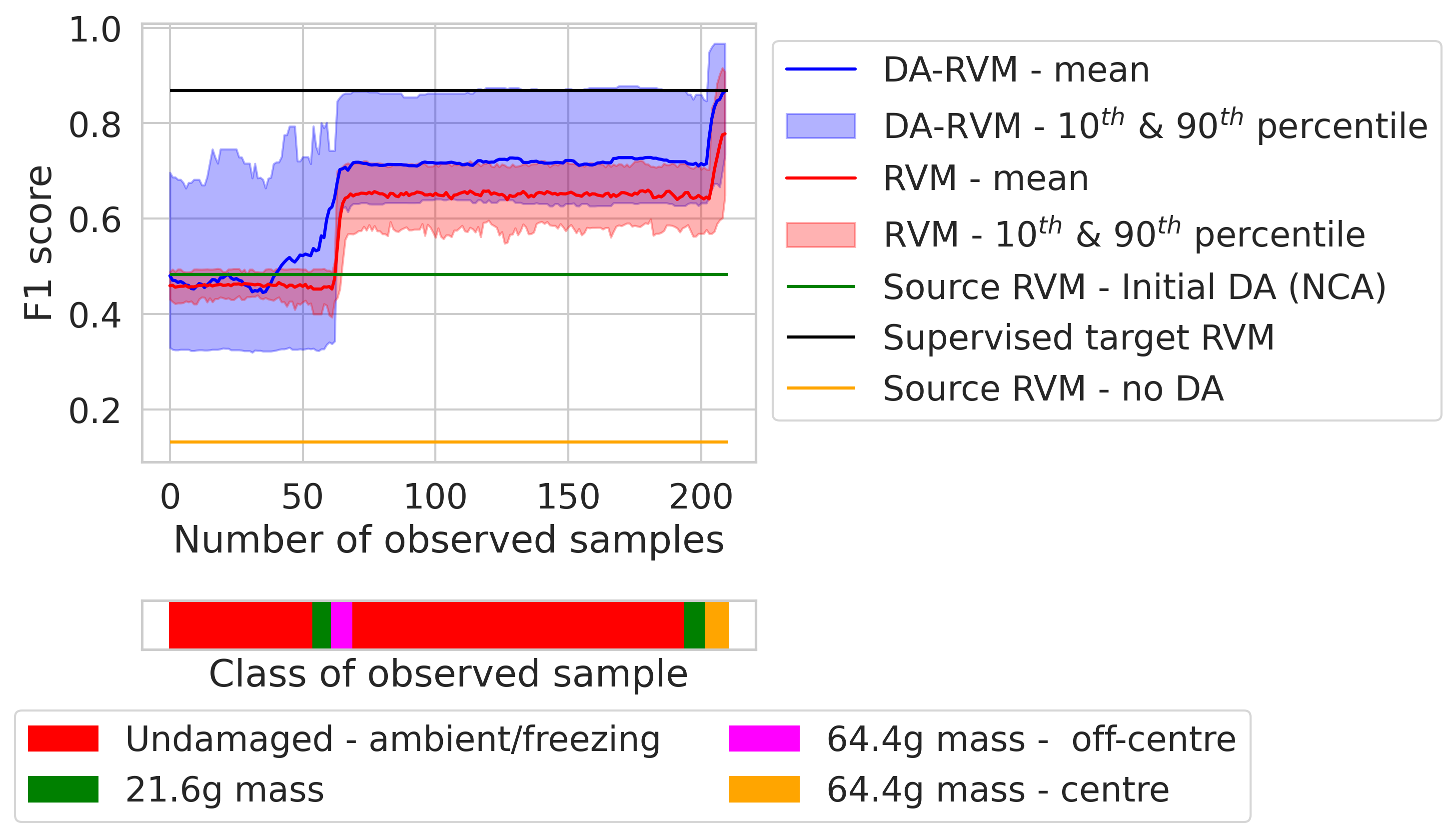}
         \caption{}
     \end{subfigure}
    \caption{The test F1 score vs the number of unlabelled samples presented to the active learners for B1$\rightarrow$B2, shown in (a), and B2$\rightarrow$B1 presented in (b). Mean F1 scores are shown by solid lines, and the region between the 10th and 90th percentile are shown by the shaded region, with the DA-RVM given in blue and the target-only RVM in red.}
    \label{fig:f1 active 1}
\end{figure}

\begin{figure}[b!]
    \centering
        \begin{subfigure}[b]{0.40\textwidth}
         \centering
        \includegraphics[width=\textwidth]{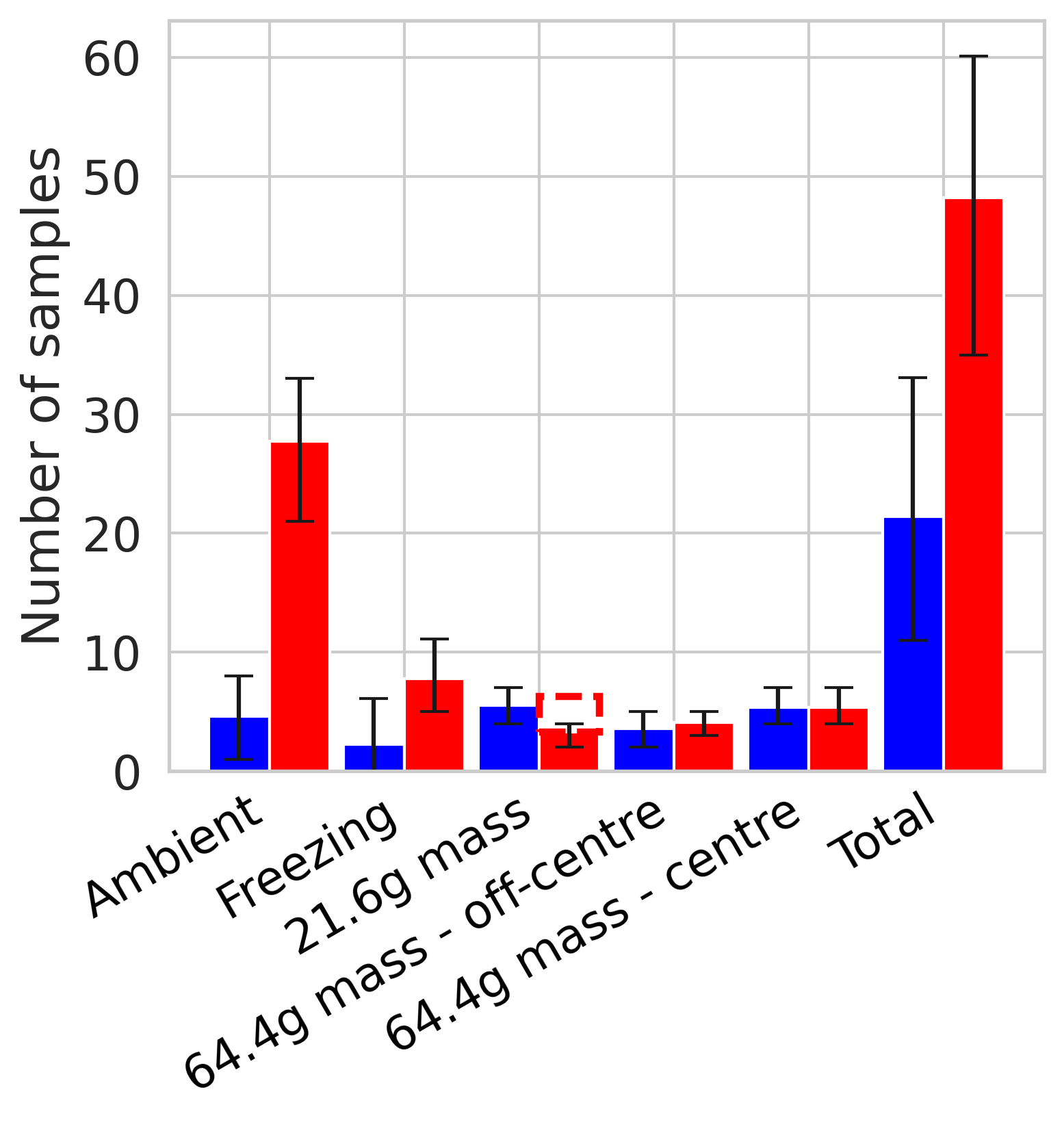}
         \caption{}
     \end{subfigure}
    \begin{subfigure}[b]{0.54\textwidth}
         \centering
\includegraphics[width=\textwidth]{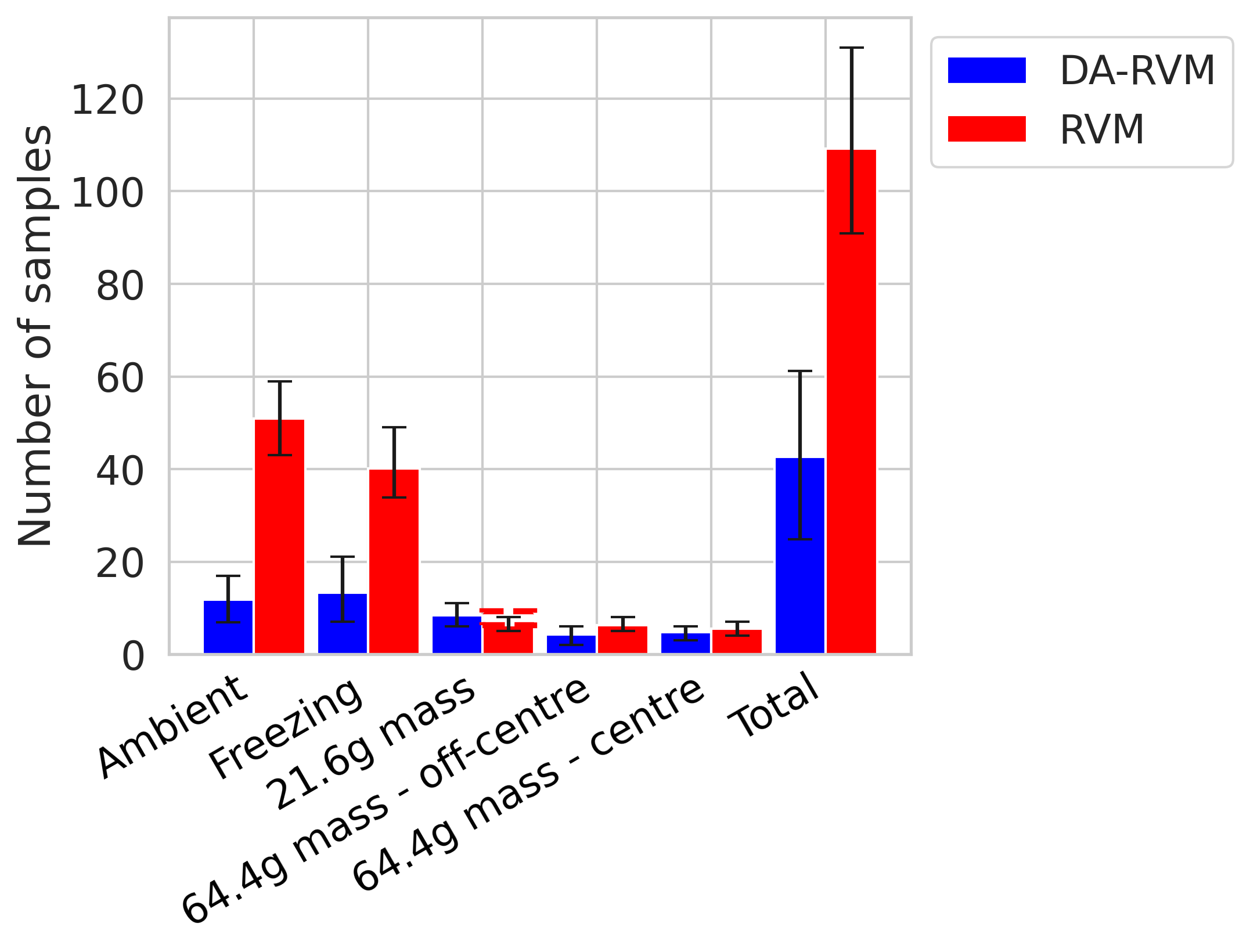}
         \caption{}
     \end{subfigure}
    \caption{The number of observations queried via active sampling using the DA-RVM (blue) and target-only RVM (red), for B1$\rightarrow$B2, shown in (a), and B2$\rightarrow$B1 presented in (b). The black lines indicate the range of samples, showing the 10th and 90th percentiles, while the red dashed lines above the 21.6g mass bar represent additional samples used to initialise the target-only RVM.}
    \label{fig:bars 1}
\end{figure}

\begin{figure}[b!]
    \centering
        \begin{subfigure}[b]{0.37\textwidth}
         \centering
        \includegraphics[width=\textwidth]{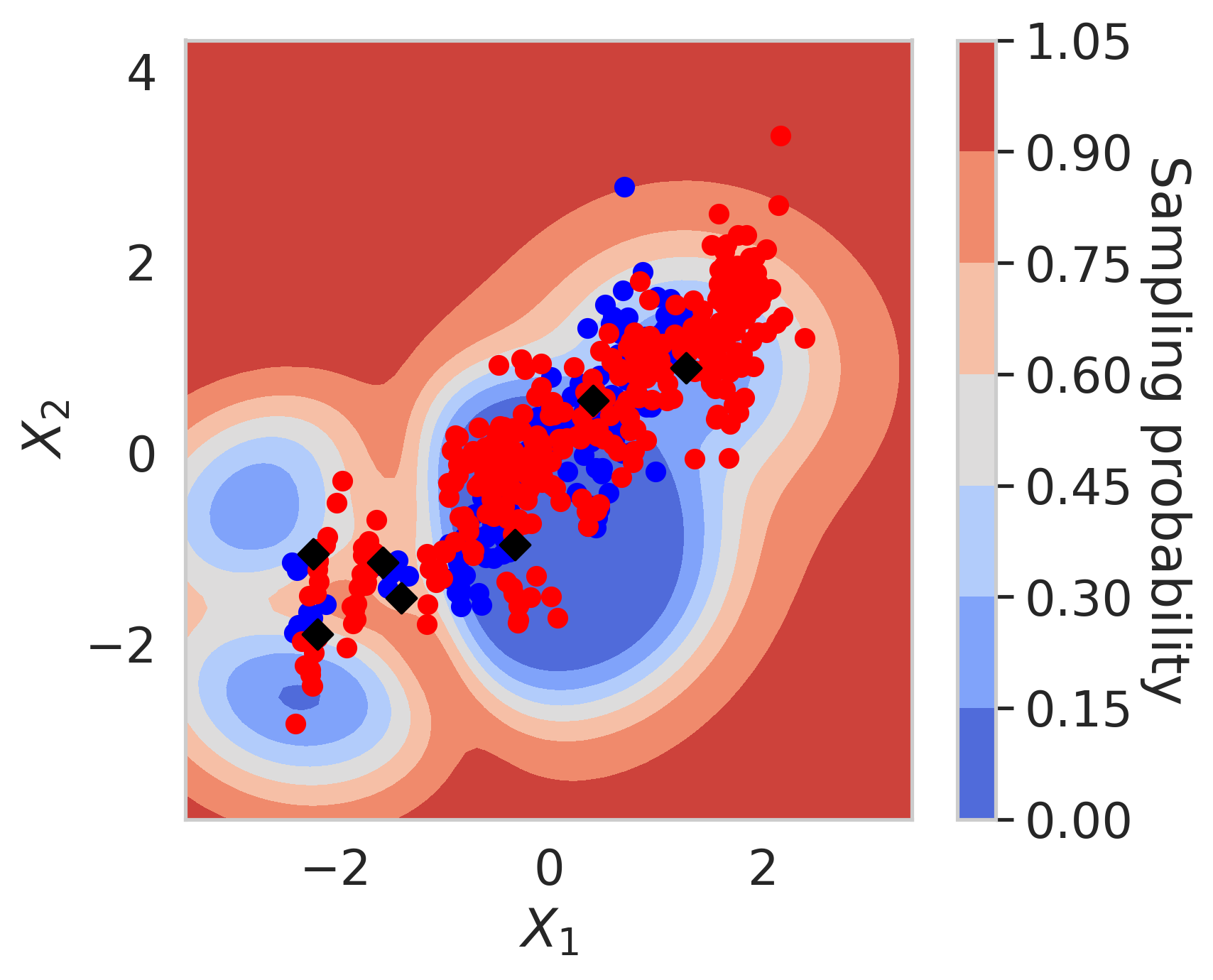}
         \caption{}
     \end{subfigure}
    \begin{subfigure}[b]{0.58\textwidth}
         \centering
\includegraphics[width=\textwidth]{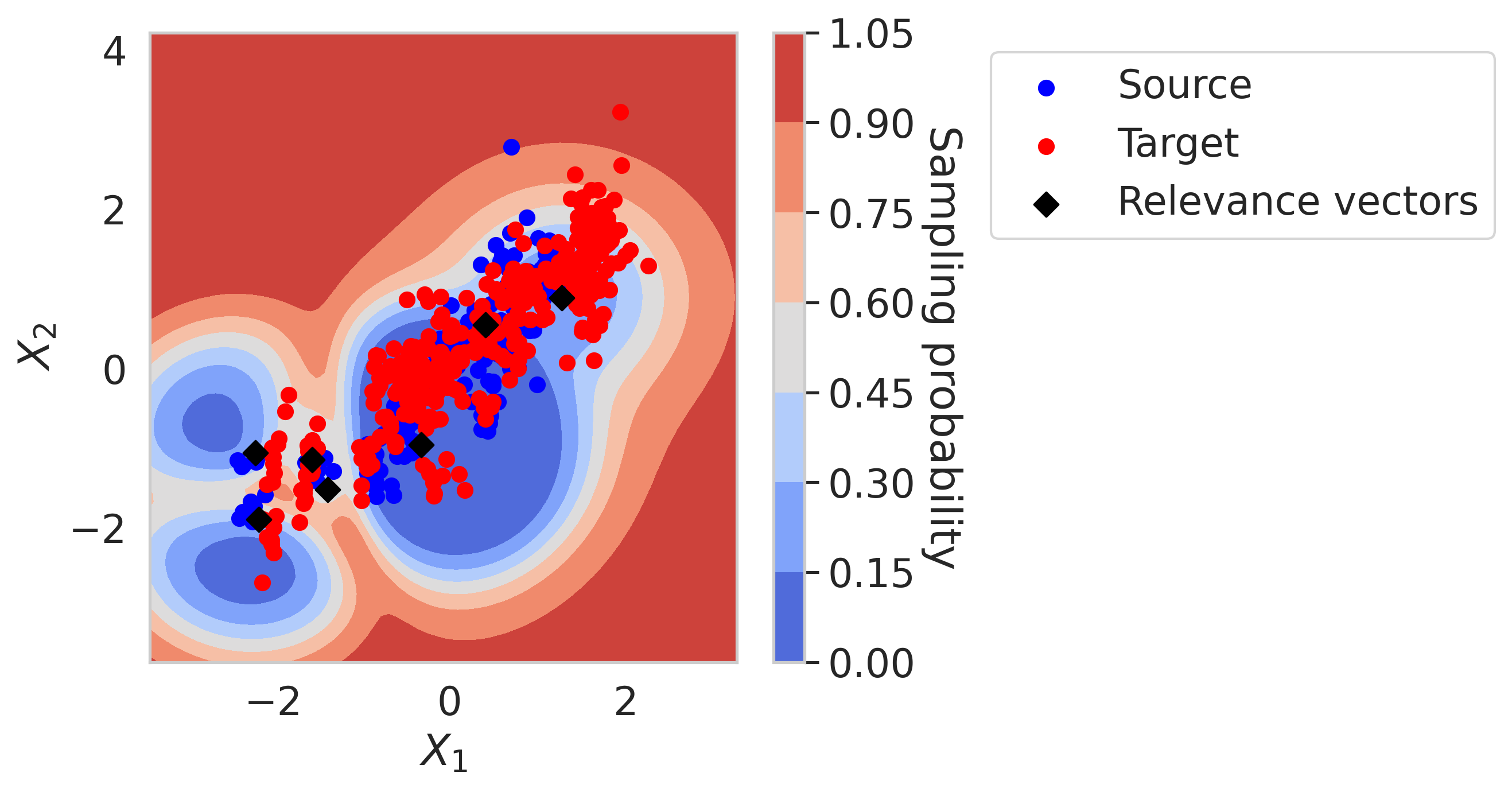}
         \caption{}
     \end{subfigure}
    \caption{An example of the sampling probability across the feature space (shown by the shaded regions) for a single test repeat in B2$\rightarrow$B1, showing training and testing data. The target data and sampling probabilities are mapped to the source domain using the expected posterior mapping. The mapping before the active-sampling process is shown in (a), and (b) shows the mapping after the DA-RVM was presented with all data. }
    \label{fig:entropy 1}
\end{figure}

Figures \ref{fig:f1 active 1}(a) and \ref{fig:f1 active 1}(b) show the F1 scores across the test set after each unlabelled observation was presented, with solid lines representing the mean F1 scores from 100 repeats and the shaded region indicating the 10th to 90th percentiles. It can be seen in both cases, naively applying a source-only classifier led to poor generalisation in the target domain, indicated by the orange line. This motivates the application of transfer learning; it can be seen applying NCA improves generalisation of the source classifier to the target domain (the mean F1 score is indicated by the green line). However, NCA still exhibits significantly worse performance compared to a fully supervised classifier learnt using target data, indicated by the black line in Figure \ref{fig:f1 active 1}, motivating the incorporation of additional information to further improve the initial NCA mapping.\\

In both cases, updating the NCA mapping with the DA-RVM using labelled target data significantly improved the F1 scores. While before any damage was observed, the DA-RVM produces similar results to NCA (the expected initial mapping), after a few observations from the first damage scenario (green and magenta regions on the colour bar), the mean F1 scores improve significantly. In addition, it consistently produces better classification than the target-only RVM prior to observing all classes, indicating that leveraging both an informative initial mapping and a few labels, the DA method is able to learn a classifier that can extrapolate to yet to be observed classes in the target. Moreover, the 10th percentile does not generally produce lower F1 scores than the mean result of the target-only RVM, demonstrating robustness to negative transfer. Finally, after all data are presented to the DA-RVM, it achieves a mean F1 score close to the fully supervised RVM, matching the target-only RVM in B1$\rightarrow$B2 (Figure \ref{fig:f1 active 1}(a)) and exceeding it in B2$\rightarrow$B1 (Figure \ref{fig:f1 active 1}(b)).\\

Although the DA-RVM achieves similar performance to the fully supervised target model at the end of the active-sampling procedure, it is able to achieve this result using fewer labelled observations, as shown in Figure \ref{fig:bars 1}\footnote{The dashed line indicates the samples used to initialise the target-only RVM; these were not selected during the sampling process.}. The DA-RVM used 10.2\% and 12.0\% of samples for B1$\rightarrow$B2 and B2$\rightarrow$B1, respectively, compared to 23.4\% and 31.5\% for the target-only RVM. While both methods reduce the number of labels required compared to a fully-supervised RVM, the DA-RVM results in fewer queries and a smaller reduction in classification performance. In addition, the DA-RVM resulted in far less normal condition data being labelled by increasing confidence in predictions of the undamaged state by leveraging source data, which in practice would reduce unnecessary inspections.\\

Compared to conventional active learning, leveraging source data not only presents the opportunity to classify data as yet to be observed classes, but also facilitates more efficient querying behaviour. This improvement can be explained by inspecting the sampling probability of the DA-RVM; an example of the sampling probability in the initial classifier for the B2$\rightarrow$B1 task is given in Figure \ref{fig:entropy 1}(a), with the target data and entropy mapped to the source feature space via the expected posterior mapping. It can be seen there are already regions where sample probability is low when the DA-RVM is initialised, particularly for samples from the largest cluster, which represent the undamaged data. As such, it appears that the source data allows for the initial model to have better defined boundaries, guiding the labelling process to prioritise damage classes, where classes are less separable and there are fewer data in the source domain. It can also be seen that by using an RVM as a classifier the model would effectively sample novel data, as the DA-RVM effectively produces low-entropy regions near observed data, whereas the extremities of the model have a sampling probability near unity. The sampling probability for the DA-RVM after the active sampling process is shown in Figure \ref{fig:entropy 1}(b), where it can be seen that obtaining labels in the target has led to a further reduction in sampling probability in some regions. This reduction in sampling probability is particularly evident for the damage classes -- the three smaller clusters.\\

\begin{figure}[h!]
    \centering
     \begin{subfigure}[b]{0.35\textwidth}
         \centering
\includegraphics[width=\textwidth]{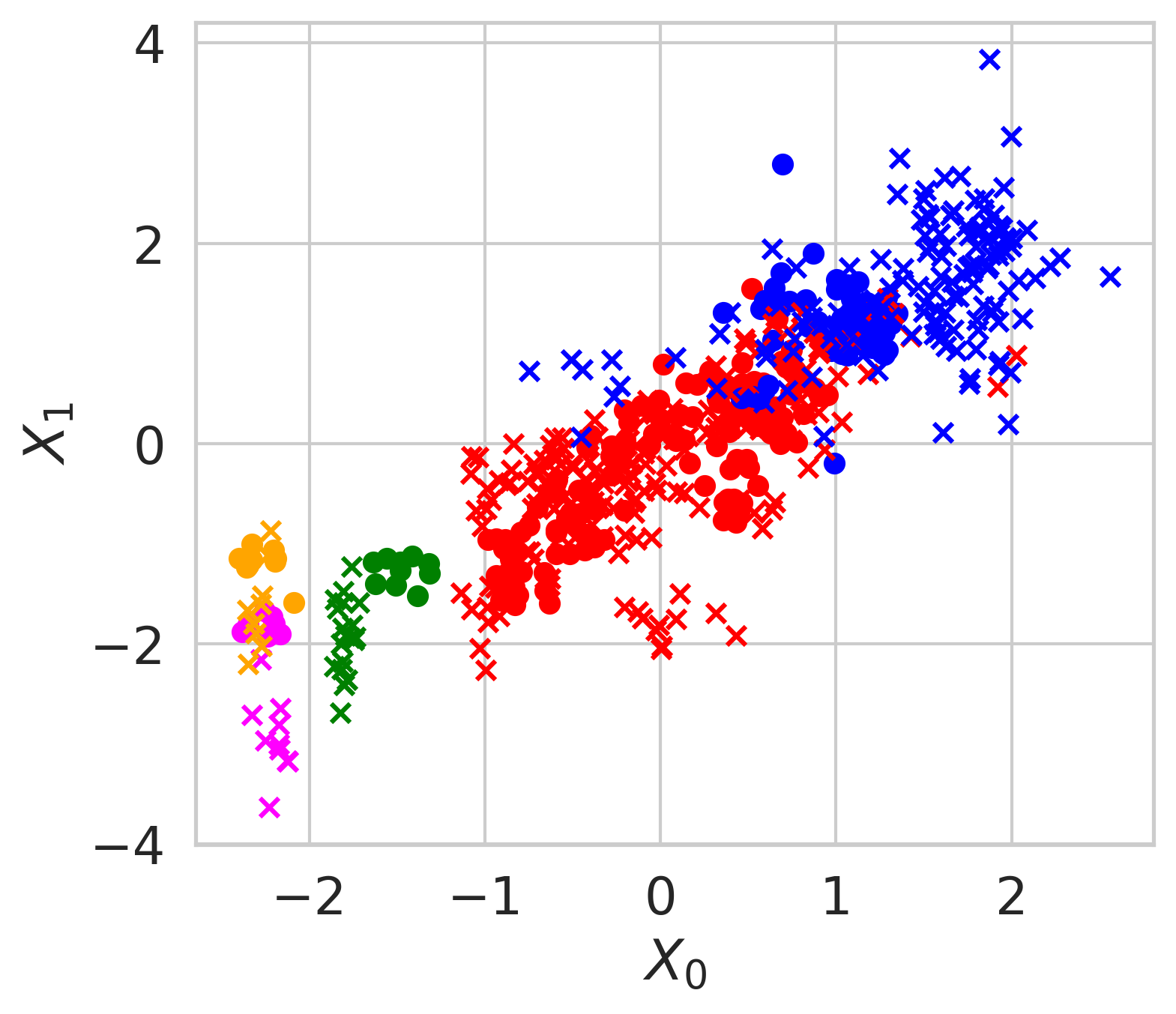}
         \caption{}
     \end{subfigure}
   \begin{subfigure}[b]{0.605\textwidth}
         \centering
\includegraphics[width=\textwidth]{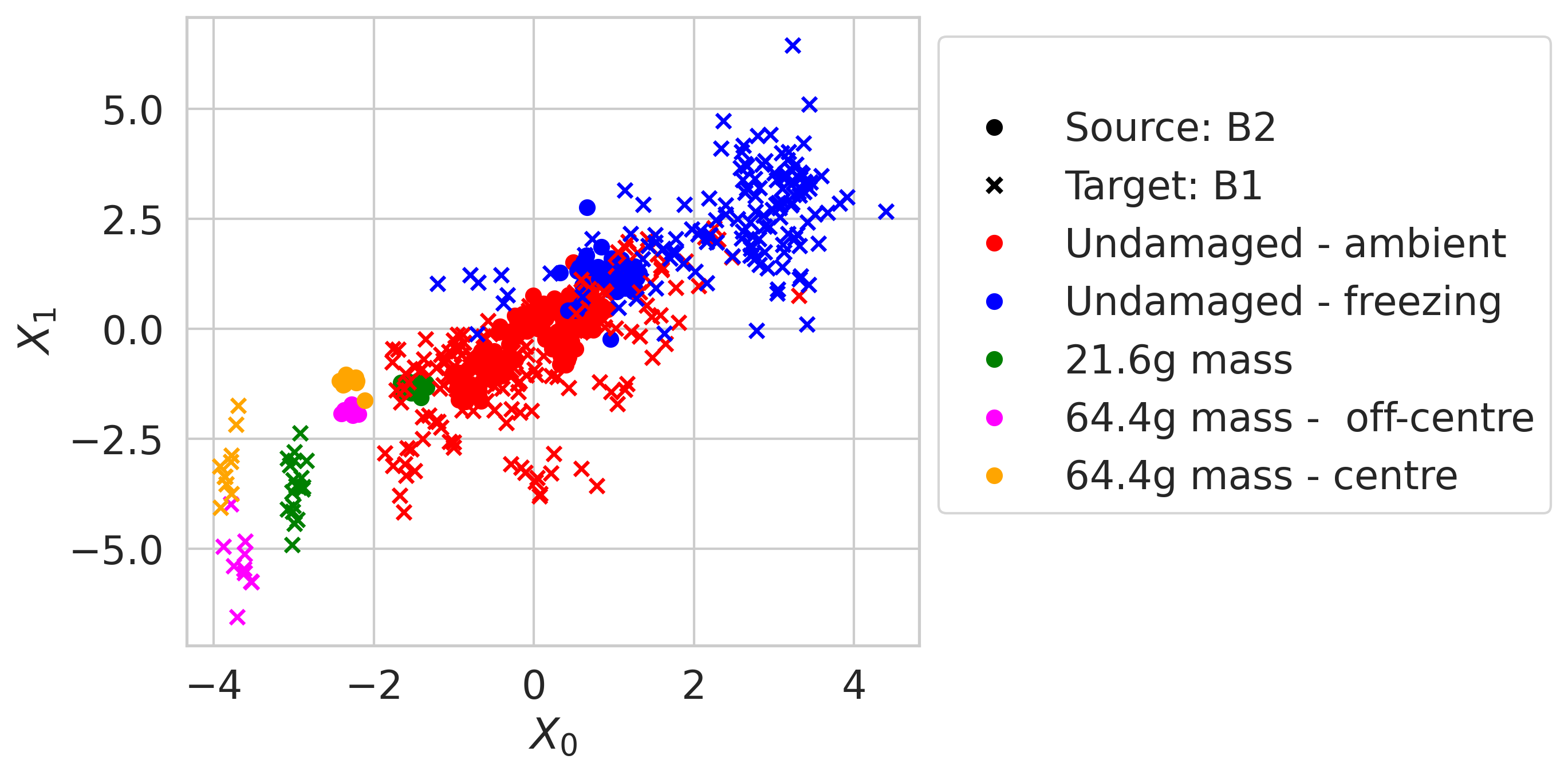}
         \caption{}
     \end{subfigure}
    \caption{An example of the training and testing data after the NCA mappings that resulted in the highest and lowest JMMD values. The NCA mappings for B2$\rightarrow$B1, shown in (a) and (b), for the lowest and highest JMMD values, respectively.}
    \label{fig:nca feats 1}
\end{figure}

\begin{figure}[h!]
    \centering
     \begin{subfigure}[b]{0.35\textwidth}
         \centering
\includegraphics[width=\textwidth]{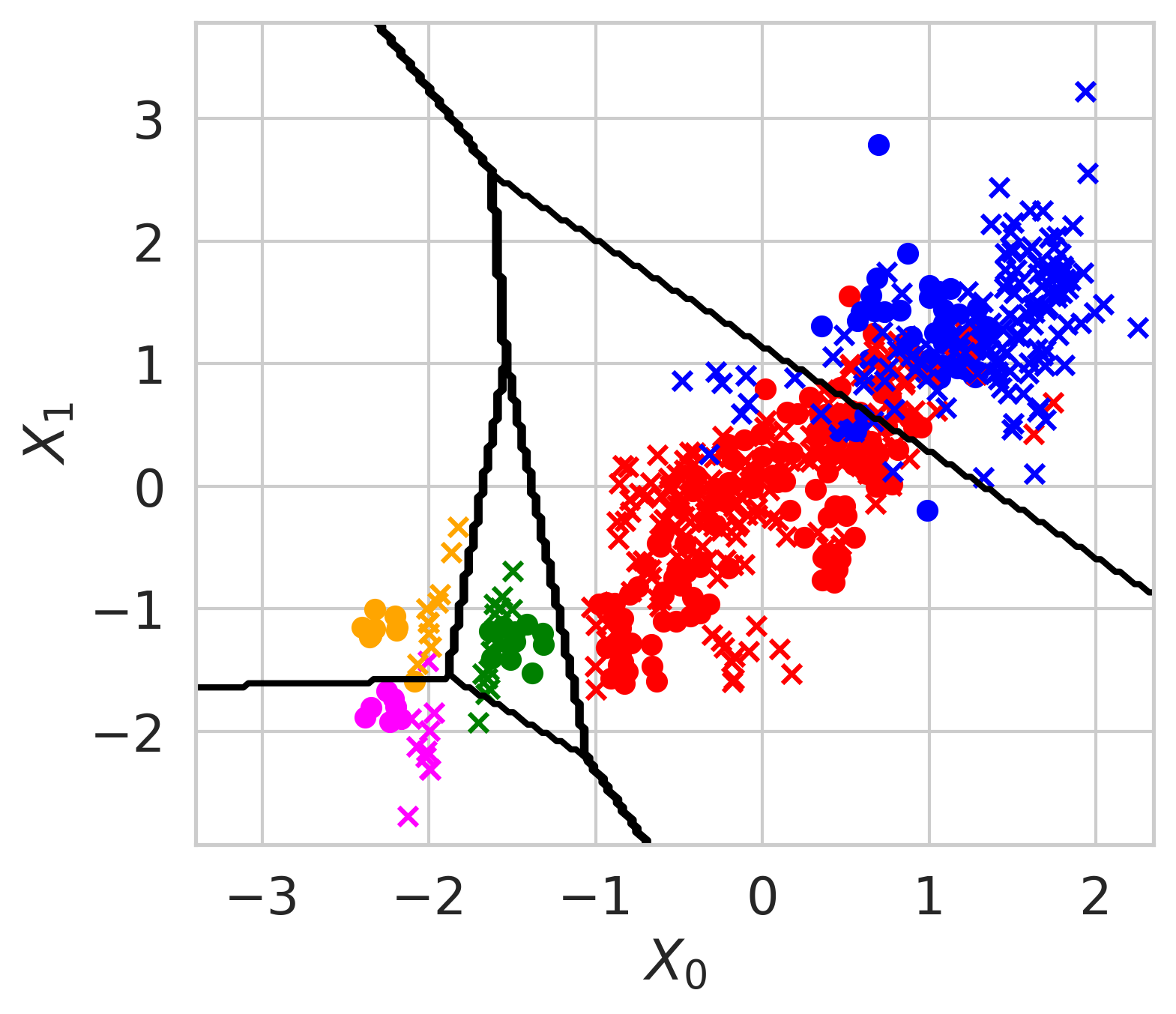}
         \caption{}
     \end{subfigure}
   \begin{subfigure}[b]{0.605\textwidth}
         \centering
\includegraphics[width=\textwidth]{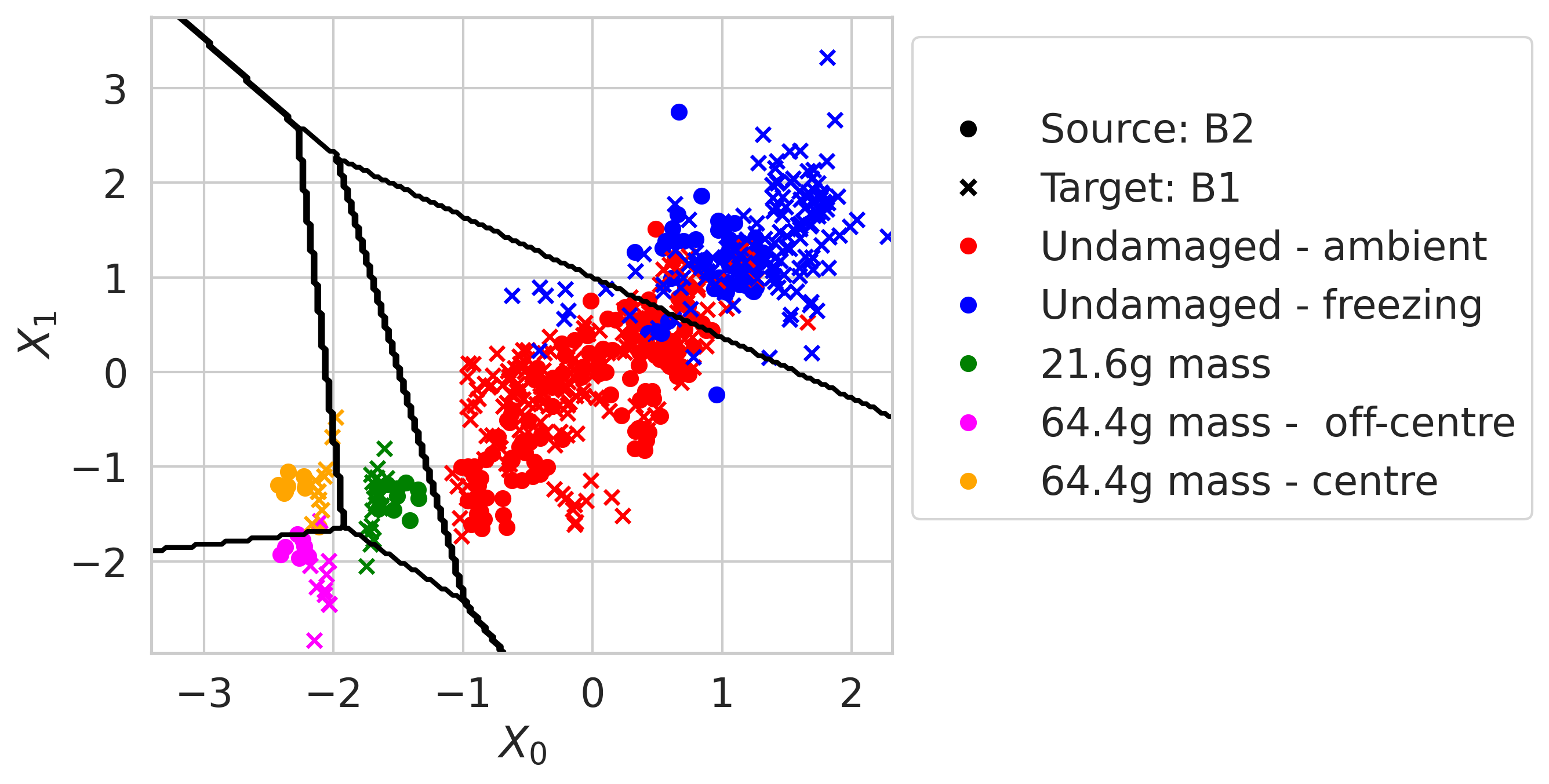}
         \caption{}
     \end{subfigure}
    \caption{An example of the training and testing data from B2$\rightarrow$B1, with the expected final DA-RVM mappings for the repeats where the NCA produced the lowest and highest JMMD values shown in (a) and (b), respectively. Solid black lines represent the decision boundary.}
    \label{fig:post feats 1}
\end{figure}

A core advantage of the DA-RVM over conventional DA is its ability to incrementally correct poor initial alignment using labels, which is particularly useful when sampling bias prevents accurate estimation of distribution divergence or when domain similarity is insufficient for unsupervised DA. This advantage is demonstrated in this case study, since the small initial random sample across various temperatures used to learn the NCA mapping resulted in varying alignment quality. To demonstrate this impact, the joint-maximum mean discrepancy (JMMD) -- a nonparametric distribution divergence measure (see \cite{Long2013} for details)-- was used to identify the ``best'' and ``worst'' NCA mappings, corresponding to the lowest and highest joint distribution distances, respectively. Features from the ``best'' mapping for B2$\rightarrow$B1 (Figure \ref{fig:nca feats 1}(a)) show the target classes closely aligned with the corresponding classes in the source domain. In contrast, the ``worst'' mapping (Figure \ref{fig:nca feats 1}(b)) shows differences in scale, with source and target damage data occupying distinct regions in the feature space. This variation variation is likely because the statistics for NCA were calculated with a small sample of normal condition data distributed across a range of temperatures. It would be expected with larger normal condition samples NCA mappings would be more consistent. These differences in initial alignment quality potentially contribute to the higher variability observed in the DA-RVM results compared to the target-only RVM (Figure \ref{fig:f1 active 1}).\\

The features found following the active-sampling process for these same ``best'' and ``worst'' repeats are presented in Figure \ref{fig:post feats 1}. It can be seen that alignment could be improved in both cases, resulting in shared classifiers that can predict samples from both domains, indicated by the classification boundaries shown in black. This result supports the idea that labels from a few damage classes can be used to improve poor initial mappings, mitigating the likelihood of negative transfer later in the monitoring campaign.  In addition, in the final feature space, the freezing temperature data for both bridges can be observed as an increase in the values of these features, following an increase in stiffness, and each damage (mass) class produces a reduction in the value of these features, showing the features derived via this DA process maintain their physical interpretability, while also facilitating shared visualisation of both datasets. 

\subsection{Case study: active transfer to a target domain with limited data}

\begin{figure}[b!]
    \centering
    \begin{subfigure}[b]{0.8\textwidth}
         \centering
        \includegraphics[width=\textwidth]{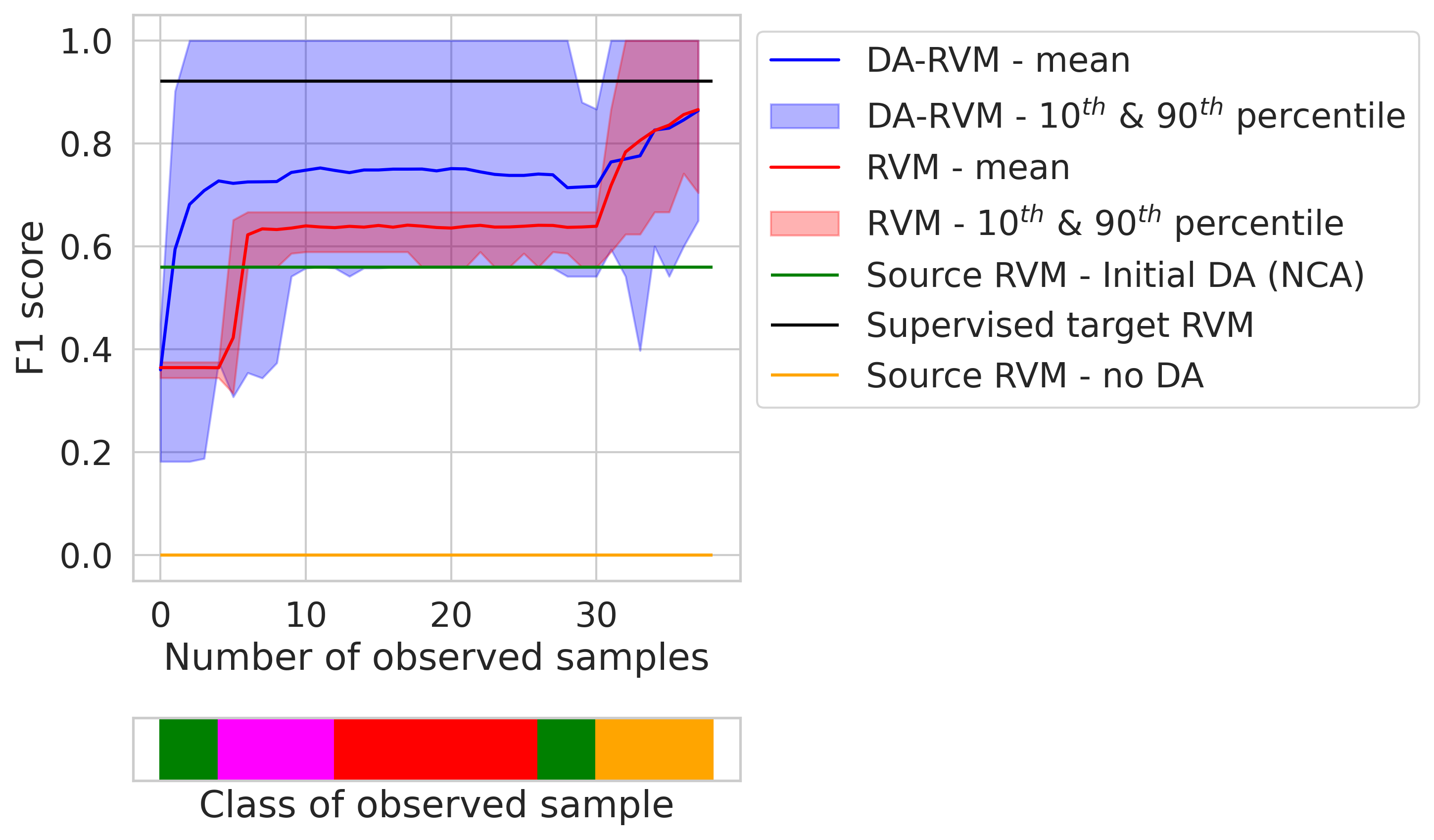}
         \caption{}
     \end{subfigure}
    \begin{subfigure}[b]{0.8\textwidth}
         \centering
         \includegraphics[width=\textwidth]{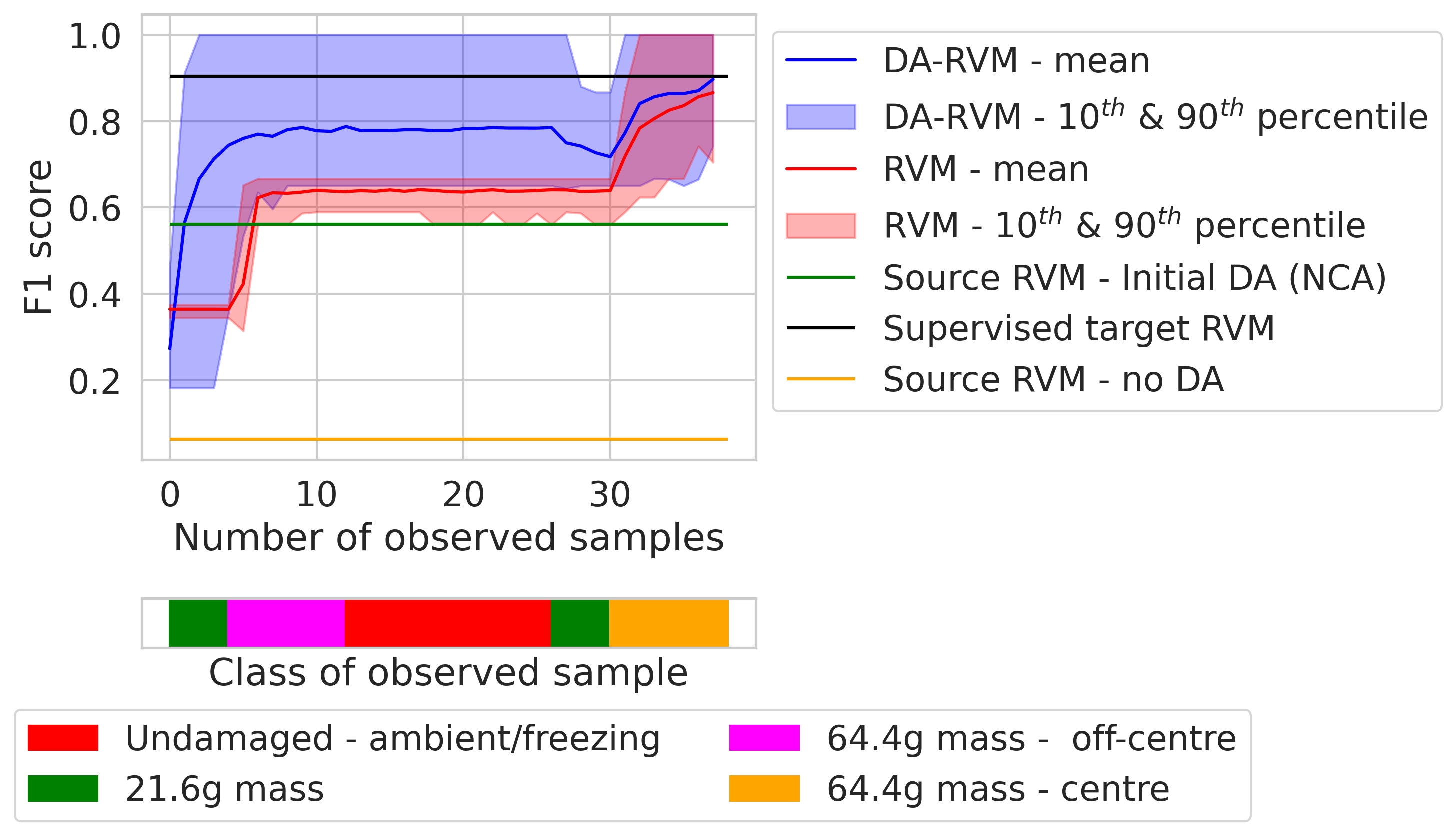}
         \caption{}
     \end{subfigure}
    \caption{The test F1 scores for the DA-RVM against the number of labelled samples selected via active sampling (shown in blue) and random sample (shown in red); B1$\rightarrow$B3 is shown in (a), and B2$\rightarrow$B3 is given in (b). }
    \label{fig:f1 active 2}
\end{figure}

\begin{figure}[h!]
    \centering
        \begin{subfigure}[b]{0.37\textwidth}
         \centering
        \includegraphics[width=\textwidth]{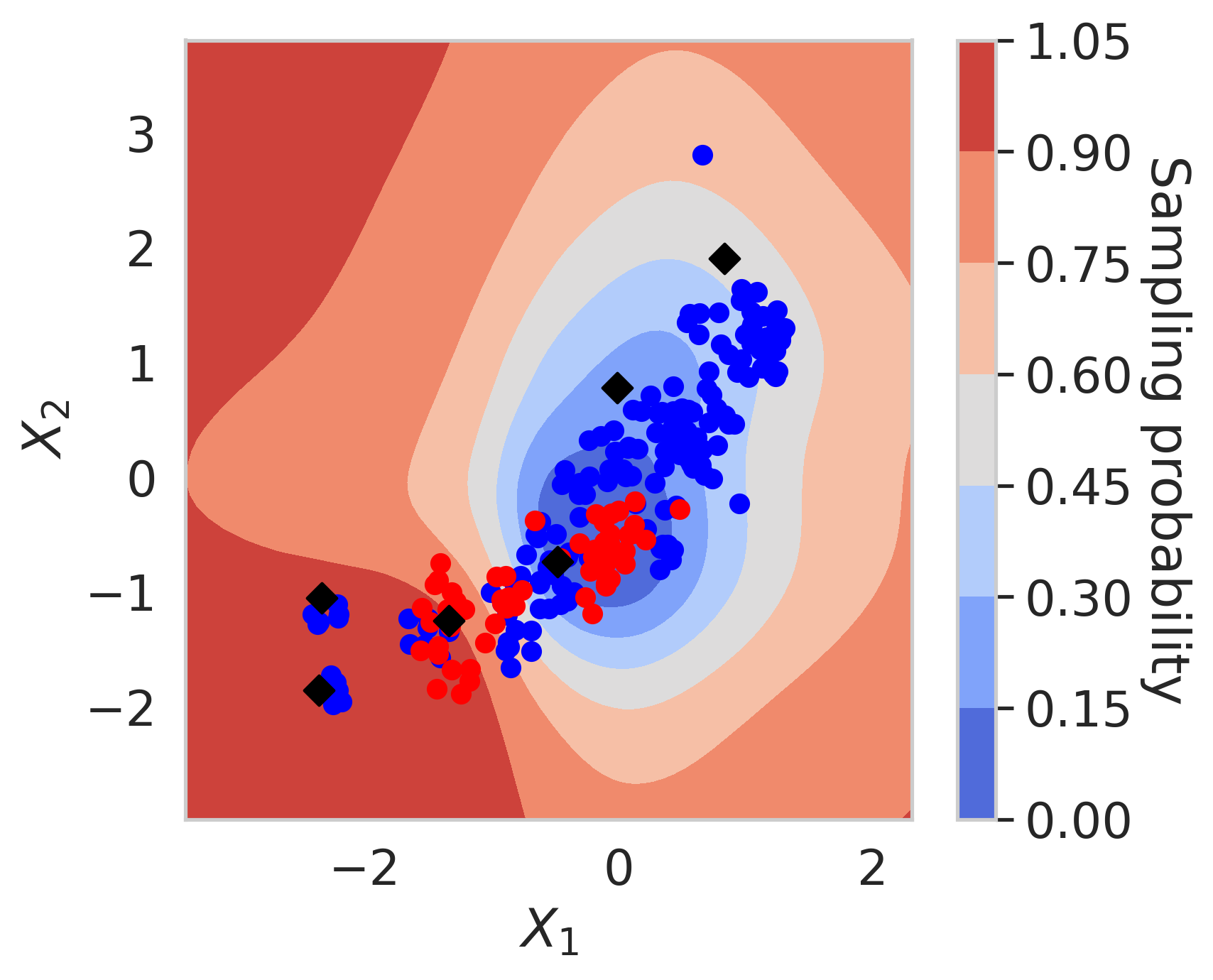}
         \caption{}
     \end{subfigure}
    \begin{subfigure}[b]{0.58\textwidth}
         \centering
\includegraphics[width=\textwidth]{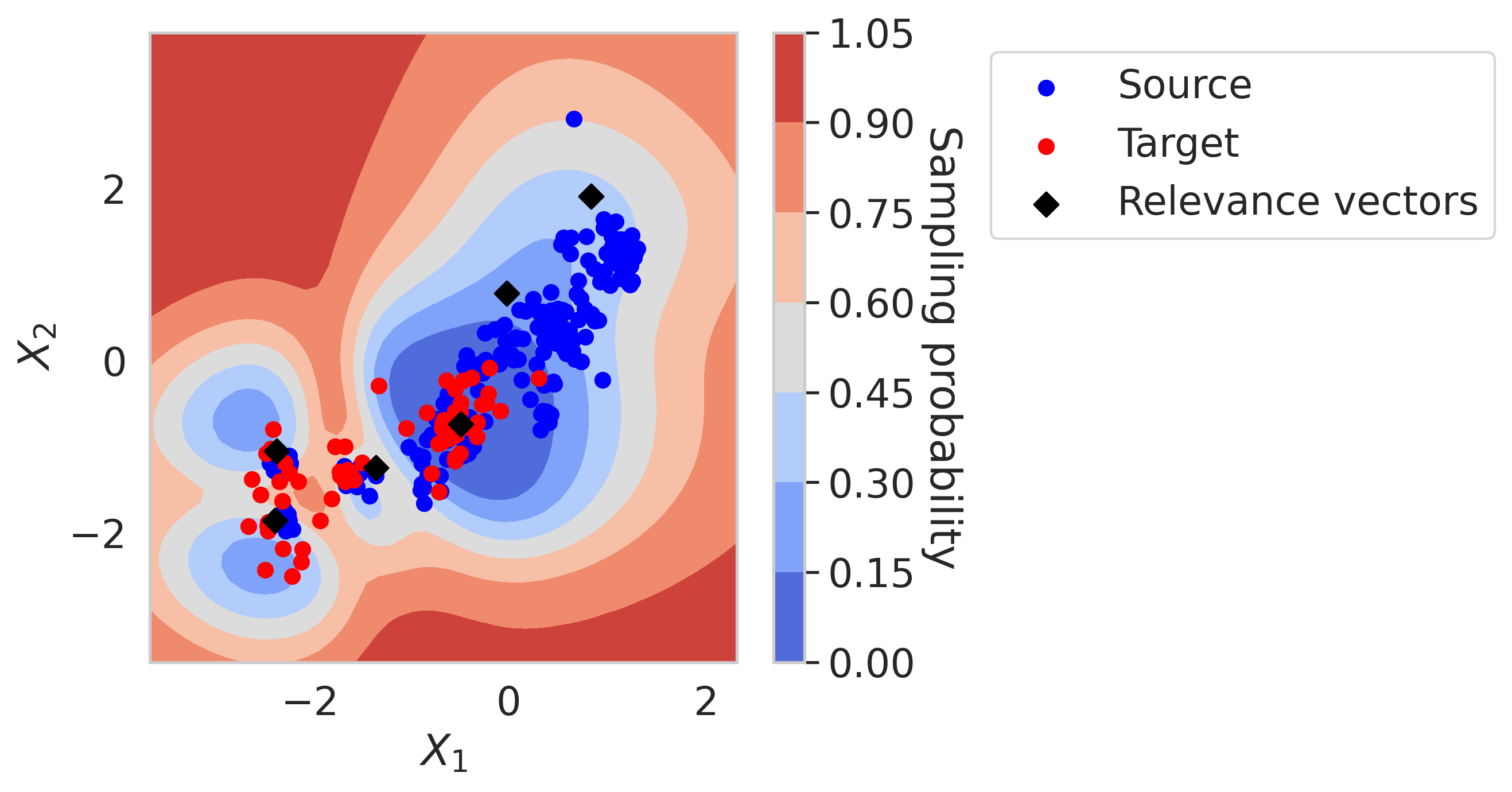}
         \caption{}
     \end{subfigure}
    \caption{An example of the sampling probability (shown by the shaded regions) for a single test repeat in B2$\rightarrow$B3, showing training and testing data. The target data and sampling probabilities are mapped to the source domain using the expected posterior mapping. The mapping before the active-sampling process is shown in (a), and (b) shows the mapping after the DA-RVM was presented with all data.}
    \label{fig:entropy 2}
\end{figure}

The second case study presents a scenario where target data to estimate initial DA mappings are extremely sparse, and not representative of the same environmental effects as the source dataset. It can be seen in this case study,  the initial classification rate of the DA-RVM is worse than NCA in both cases, shown in Figure \ref{fig:f1 active 2}. There are two potential reasons for these results. First, in this case study, only one class is available at the start of the sampling process, meaning there is limited information to constrain the posterior mapping parameters. Second, the prior variance on the translation and scale parameters were increased to $\sigma_t = \sigma_s = 1$. Thus, since the initial information available to reduce the posterior variance is limited, this choice of priors seems to have increased the uncertainty of prediction for all the damage classes to near uniform; this can been seen by the initial sampling probabilities in Figure \ref{fig:entropy 2}(a), where it can be seen the regions of the feature space away from the normal condition (the large cluster at the origin), has a sampling probability of near one (corresponding to a uniform labelling probability). This behaviour would often be desirable, and it highlights the importance of considering both classification and mapping uncertainty, as this increased prediction uncertainty could limit the impact of incorrect classifications on decision-making in scenarios where trust in initial transfer is low. Note that in B2$\rightarrow$B3, the target-only RVM achieves a slightly higher initial F1 score, likely due to being initialised with three data points from the 21.6g mass state, enabling classification of this class prior to observation of these data during the active-sampling process; however, in practice, such data would be unavailable at this stage.\\

\begin{figure}[b!]
    \centering
        \begin{subfigure}[b]{0.40\textwidth}
         \centering
        \includegraphics[width=\textwidth]{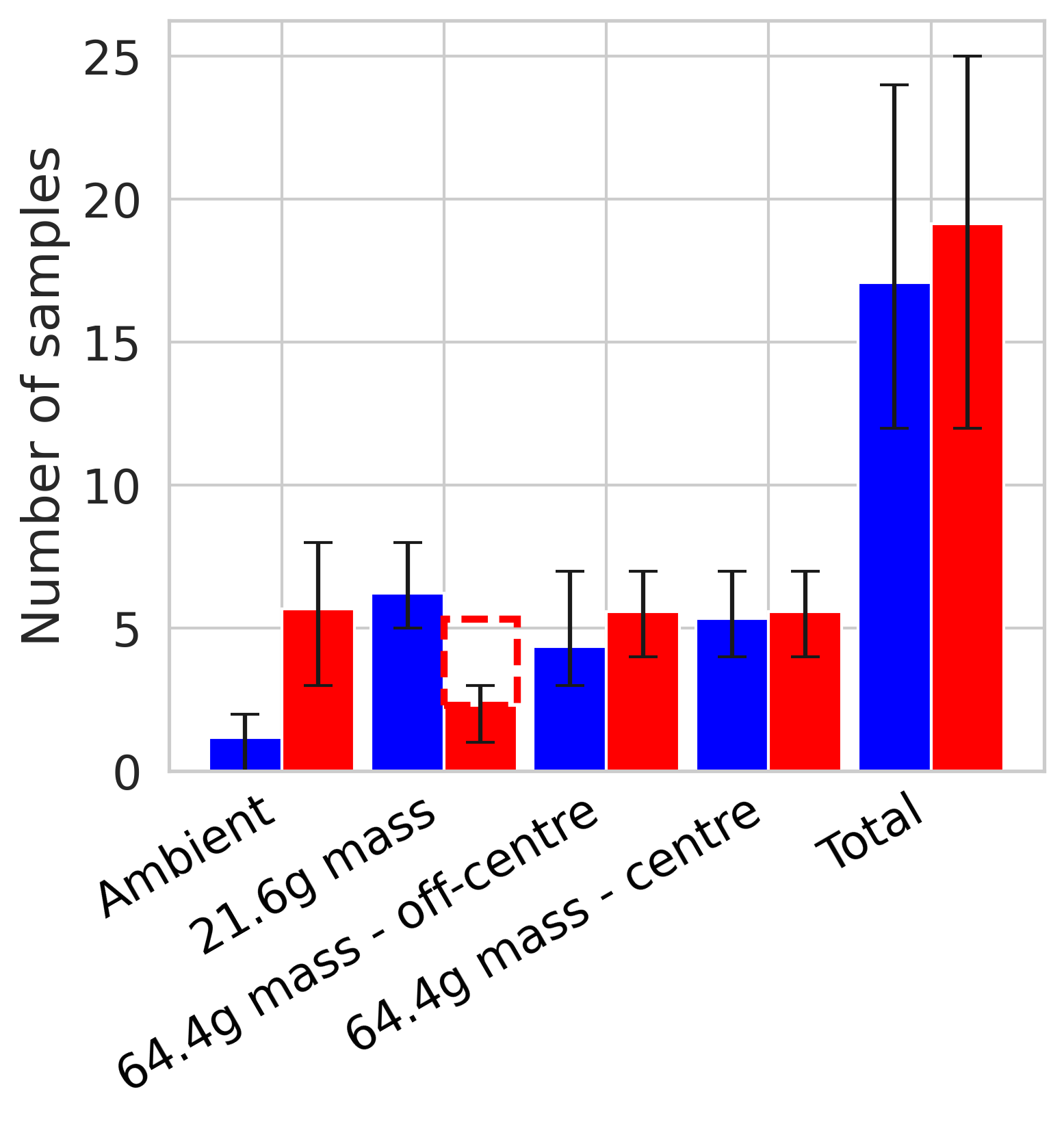}
         \caption{}
     \end{subfigure}
    \begin{subfigure}[b]{0.54\textwidth}
         \centering
\includegraphics[width=\textwidth]{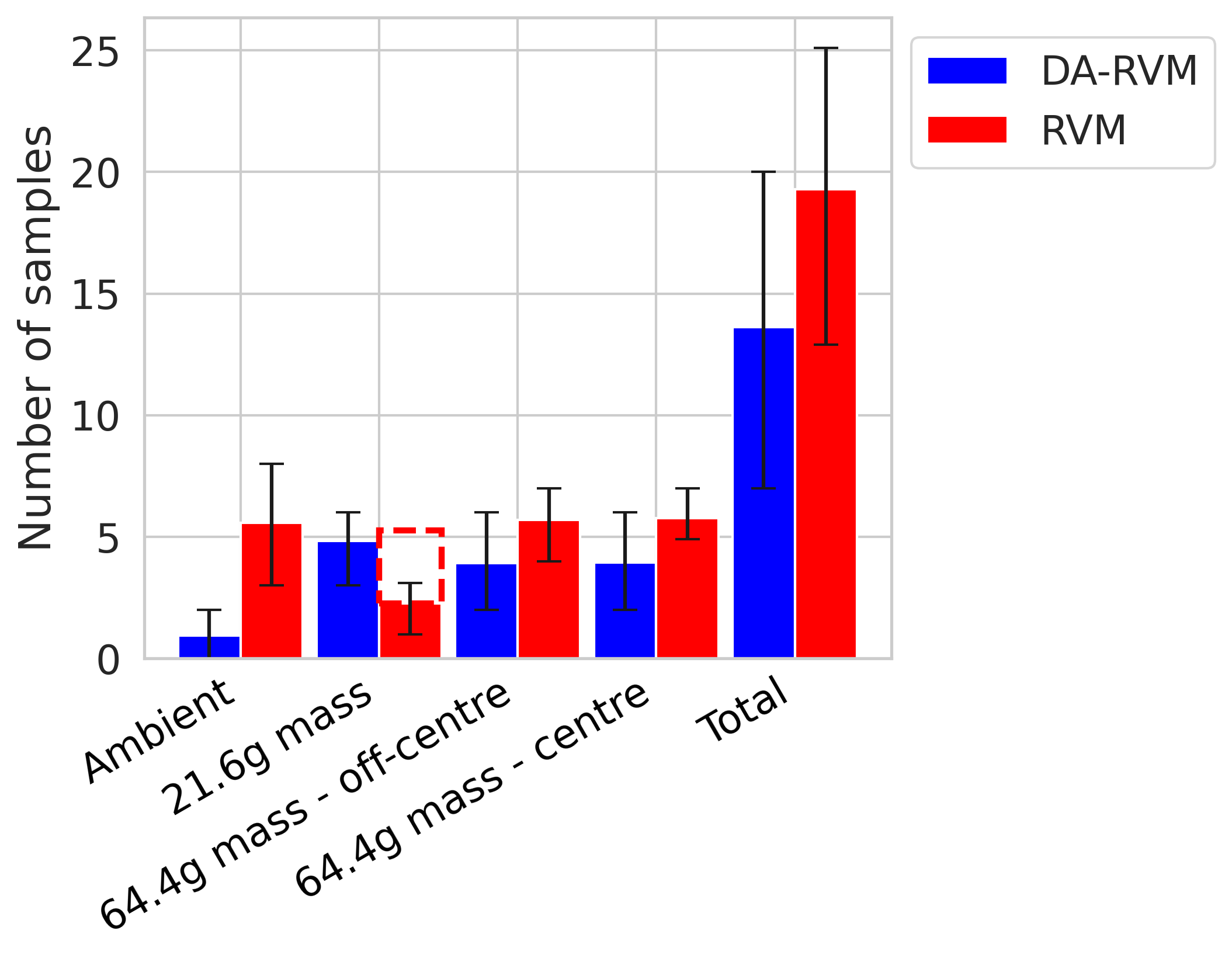}
         \caption{}
     \end{subfigure}
    \caption{The number of observations queried via active sampling using the DA-RVM (blue) and target-only RVM (red) is shown for B1$\rightarrow$B3 in (a) and for B2$\rightarrow$B3 in (b). The black lines indicate the range of samples, showing the 10th and 90th percentiles, while the red dashed lines above the 21.6g mass bar represent additional samples used to initialise the target-only RVM.}
    \label{fig:samples 2}
\end{figure}

Similarly to the previous case study, it can be seen that observing small quantities of labelled target data allowed for significant improvements in the F1 score, as shown in Figure \ref{fig:f1 active 2}. Following only a few observations of the 21.6g mass-state, the rise mean F1 score for classification of all classes is particularly pronounced in this case study, and the target-only RVM only achieves similar F1 scores after observing data from all classes. This result provides further evidence that by using only data from a minor damage extent in one location, the mapping parameters can be updated to allow for classification of classes where labelled data are only available in the source domain. Furthermore, after only observing a few observations from the 21.6g mass-state, the 90th percentile reaches an F1 score of unity, showing that in some cases, only a few data from a minor damage state allows for the effective transfer of all the labelled data in the source domain. In addition, at the end of the sampling process, both the DA-RVM and the target-only RVM achieved similar mean F1 scores, which are approaching the result of a fully-supervised RVM, as shown in Figure \ref{fig:f1 active 2}, while using far fewer labelled data as shown in Figure \ref{fig:samples 2}.\\

\begin{figure}[h!]
    \centering
     \begin{subfigure}[b]{0.35\textwidth}
         \centering
\includegraphics[width=\textwidth]{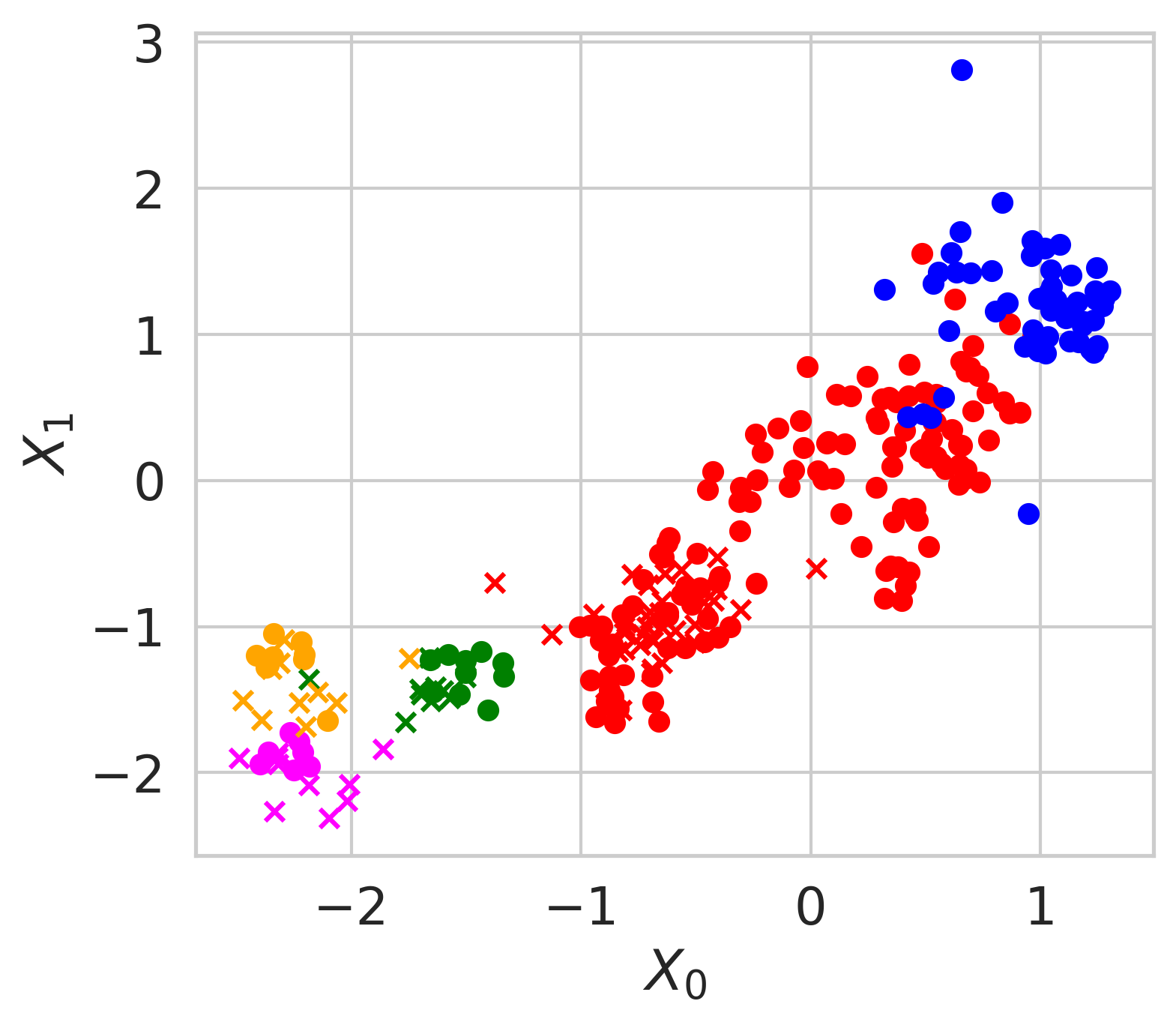}
         \caption{}
     \end{subfigure}
   \begin{subfigure}[b]{0.605\textwidth}
         \centering
\includegraphics[width=\textwidth]{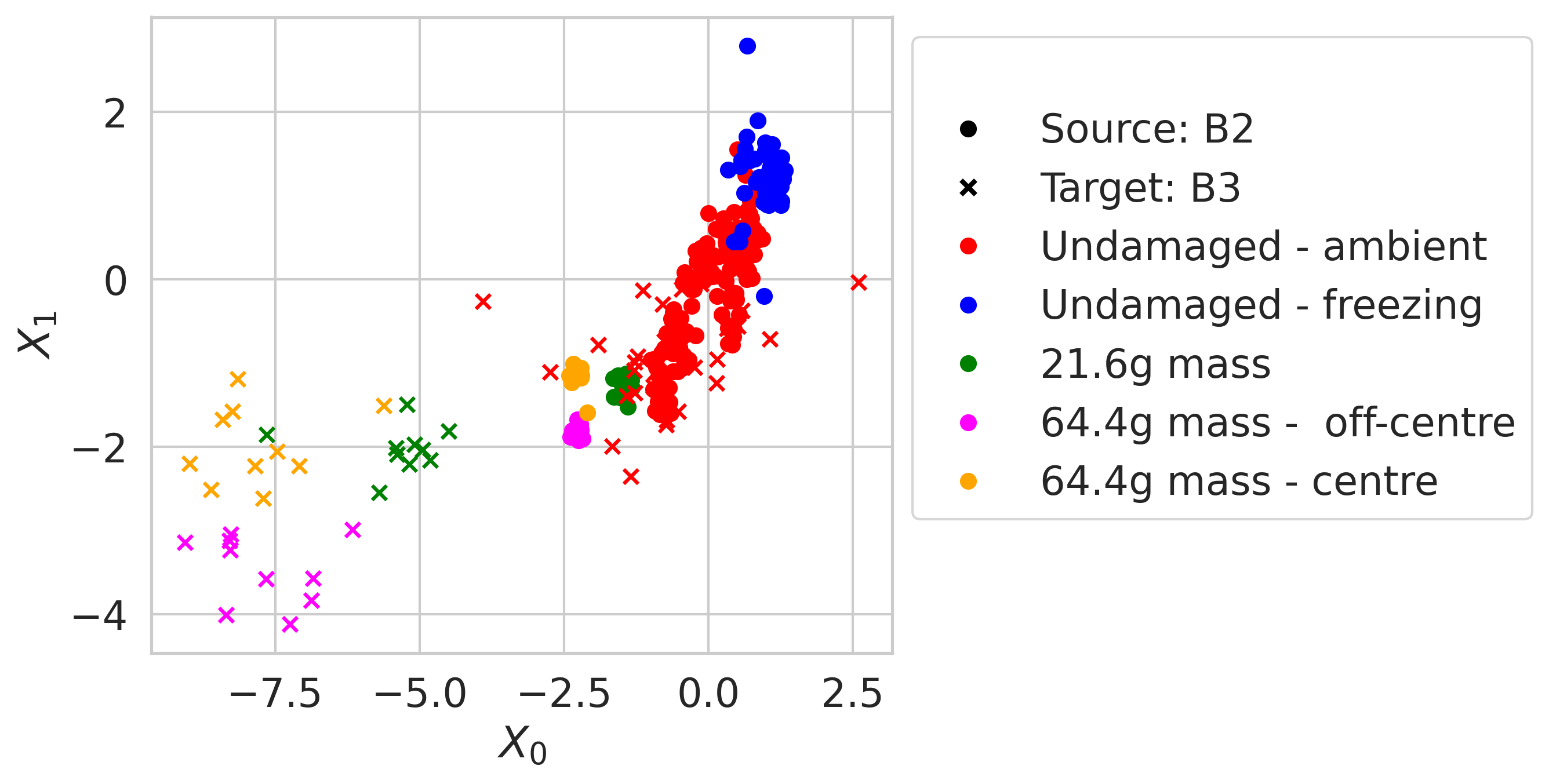}
         \caption{}
     \end{subfigure}
    \caption{An example of the data (training and testing data), after the NCA mappings that resulted in the highest and lowest JMMD values. The NCA mappings for B2$\rightarrow$B3, shown in (a) and (b), for the lowest and highest JMMD values, respectively.}
    \label{fig:nca feats 2}
\end{figure}

\begin{figure}[h!]
    \centering
     \begin{subfigure}[b]{0.35\textwidth}
         \centering
\includegraphics[width=\textwidth]{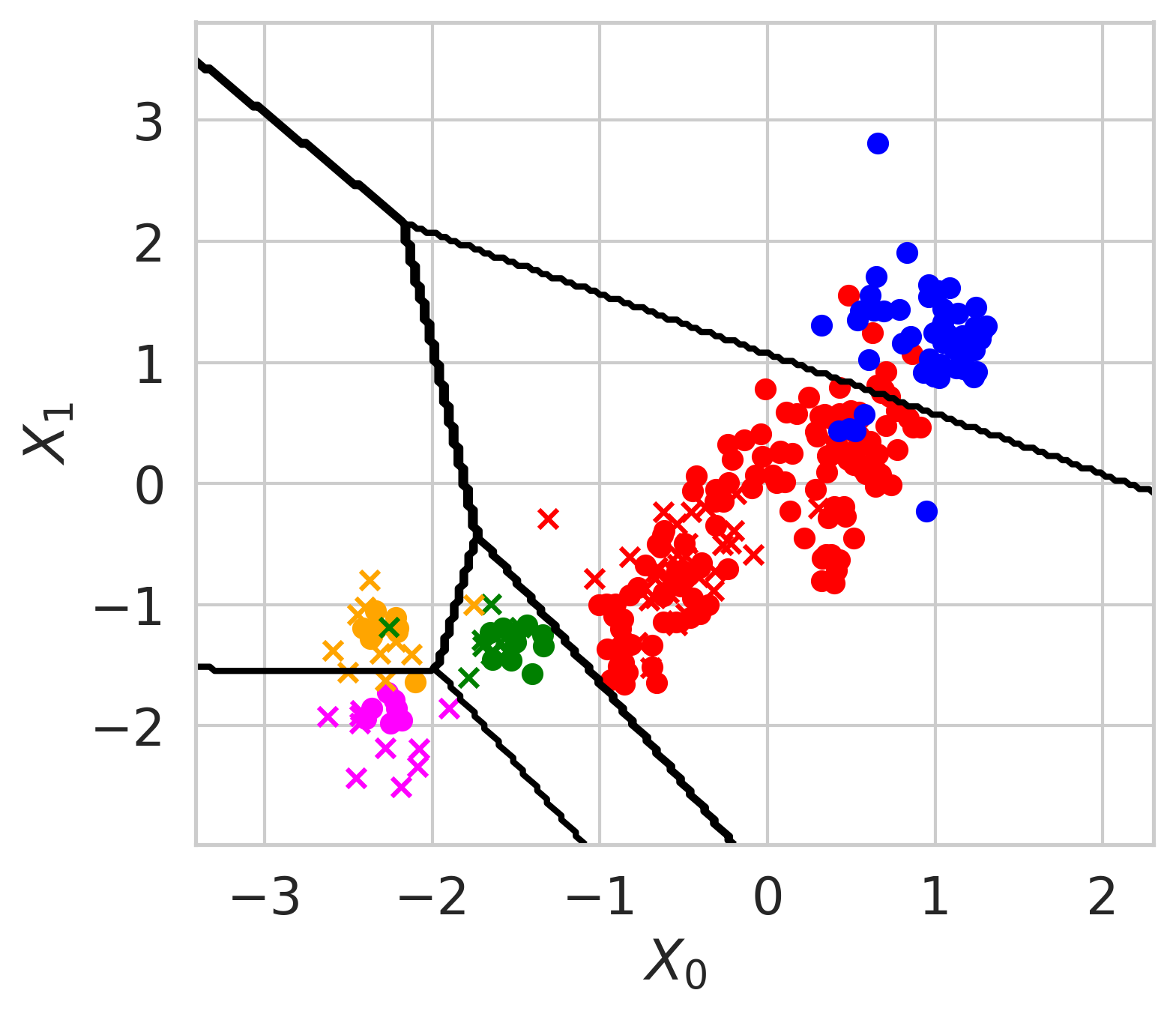}
         \caption{}
     \end{subfigure}
   \begin{subfigure}[b]{0.605\textwidth}
         \centering
\includegraphics[width=\textwidth]{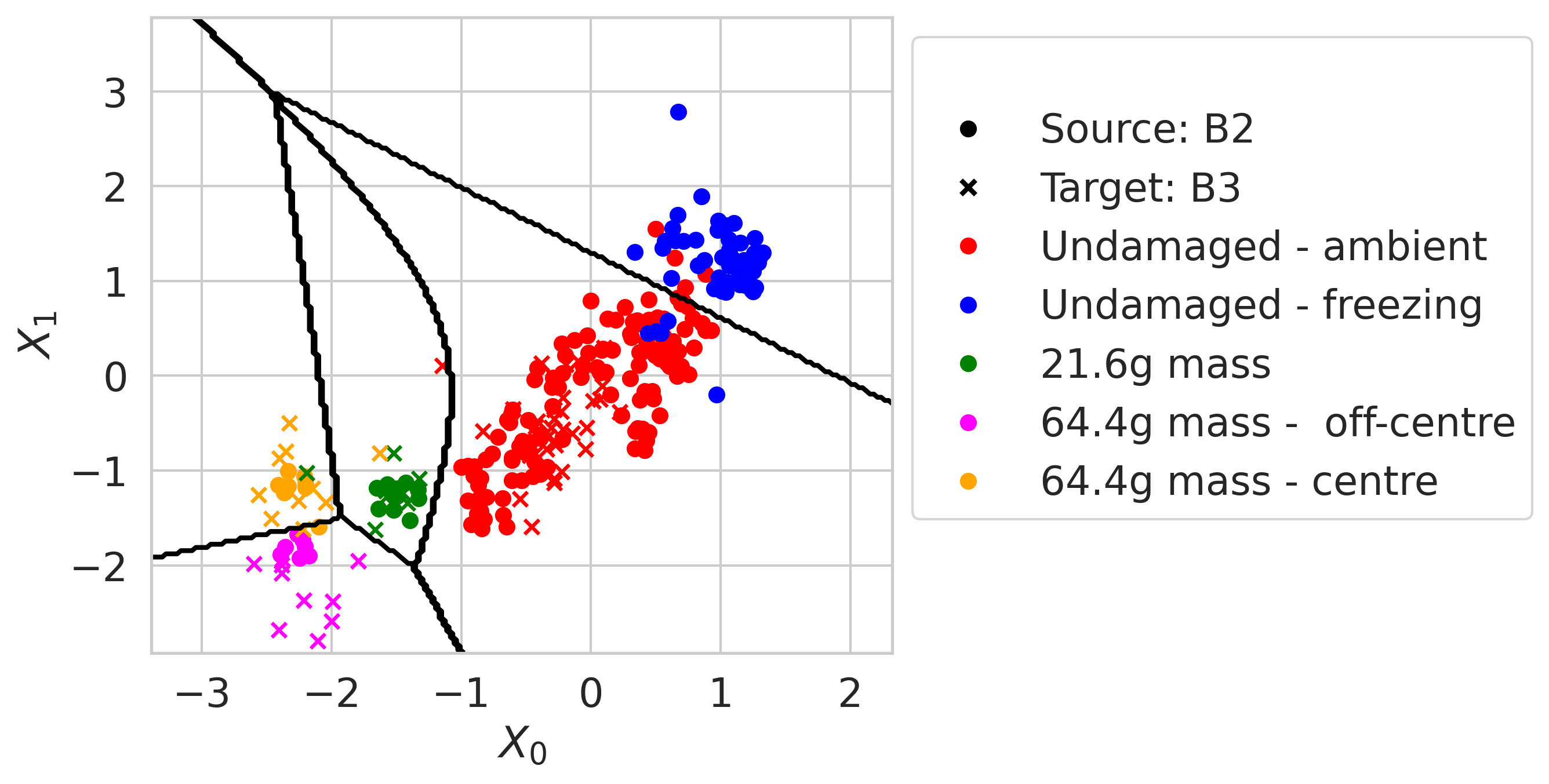}
         \caption{}
     \end{subfigure}
    \caption{An example of the training and testing data from B2$\rightarrow$B3, with the expected final DA-RVM mappings for the repeats where the NCA produced the lowest and highest JMMD values shown in (a) and (b), respectively. Solid black lines represent the decision boundary.}
    \label{fig:post feats 2}
\end{figure}

Examining the ``best'' (Figure \ref{fig:nca feats 2}(a)) and ``worst'' (Figure \ref{fig:nca feats 2}(b)) NCA mappings, selected using the JMMD as in the previous section, it can be seen there are significant discrepancies between the initial mappings. While visually the ``best'' mapping (Figure \ref{fig:nca feats 2}(a)) appears to align the target such that classes in the source are close to the corresponding target class, the ``worst'' mapping seems to have a large difference in scale -- visually, this discrepancy in scale appears to be larger than the ``worst'' example from the previous case study (shown in Figure \ref{fig:nca feats 1}(b)). This result is perhaps caused by the small sample of normal condition data used to estimate the target mean and standard deviation being insufficient to produce unbiased estimates of the statistics. These poor mappings may be a contributing factor for the 10$^{th}$ percentile indicating negative transfer in Figure \ref{fig:f1 active 2}. However, following observation of all data the DA-RVM was able to correct the poor initial mapping in the selected ``worst'' case, as shown by the expected DA-RVM posterior mapping found after observing all data shown in Figure \ref{fig:post feats 2}.\\

It is worth noting here that the specification of the prior variance has an important effect on the final mapping. Given the small number of labelled data, if the mapping variance is assumed to be small and large scale differences are present (as seen in Figure \ref{fig:post feats 2}(a) and Figure \ref{fig:post feats 2}(b)), the DA-RVM may struggle to learn the large-scale values needed to correct this misalignment, as such values are unlikely under the prior. Appendix A presents the same examples where the variance of the mapping parameters was chosen to be $\sigma_t = \sigma_s = 0.1$ to demonstrate this issue, showing that the worst mappings are unchanged, even with labels. This highlights the important balance when defining the prior variance: it must be high enough to avoid over-constraining the mapping parameters, yet low enough to prevent the mapping from overfitting to the limited target data. \\

As with the previous case, the variation in results for the DA-RVM is higher than the target-only RVM. Furthermore, while the DA-RVM introduces the potential of increasing the test F1 score beyond the maximum possible F1 score for a target-only model, in contrast to all other case studies, in B1$\rightarrow$B3 the 10th percentile of the DA-RVM also drops below the target-only RVM at some stages of the active sampling process, as shown in Figure \ref{fig:f1 active 2}(a). The few test repeats producing worse F1 scores with the DA-RVM suggest that although labels can mitigate negative transfer using labels, the negative effect of poor initial mappings can persist even after the inclusion of labels. This result is likely caused by poor initial NCA mappings, and shows that even though leveraging labels may reduce the likelihood of negative transfer, it may still be a critical issue and selection of a suitable source structure, features and data preprocessing are crucial considerations.\\ 

\section{Discussion and Conclusions}

A critical limitation of conventional data-driven approaches to SHM is that supervised machine learning methods require a fully-labelled dataset with examples representing each health state of interest, which is often costly and/or unfeasible. Two technologies for reducing the label requirement of supervised classifiers are transfer learning and active learning. Previous studies have considered these technologies independently; however,  this paper proposes an active transfer-learning strategy to address several key challenges of considering either approach independently, resulting in a practical framework for online learning in PBSHM. \\

Four transfer tasks were used to demonstrate the proposed framework by transferring a damage classifier between lab-scale bridge structures subject to various temperatures and pseudo-damage states. In all cases, leveraging labelled source data enabled the DA-RVM to classify health states that had not been observed in the target, even when the target dataset only contained data corresponding to a subset of the classes in the source dataset -- showing robustness to class imbalance. On the other hand, conventional active-learning approaches can only classify data from previously-observed health-states. The ability to classify health-states prior to their observation in the target domain has significant implications in SHM, as predictions about health-states critical to decision making could be achieved before these health-states are observed in the target structure and without repeating labelling efforts. Furthermore, the active transfer-learning approach resulted in fewer overall queries compared to conventional active learning, which in practice would result in a reduction in inspections; hence, lower operational costs throughout a monitoring campaign. \\

There are several interesting potential directions for future work. One of the main limitations of the current approach is that the number of observations labelled is not directly related to a labelling budget. In practice, an operator would have a limited budget for inspections; thus, the current approach may exhaust the budget prior to observing all data. Ensuring the labelling budget is not depleted early in the sampling process is a common challenge in stream-based active learning \cite{settles2009active}. A potential solution would be to ensure that labels correspond to health-states useful for decision making, i.e. using a decision-based sampling procedure as in \cite{hughes2022risk, hughes2022robust}. While this approach can still request more labels than a budget allows, it could result in fewer queries overall as typically data corresponding to minor damage-states are labelled less, since they are not critical for decision making. \\

This paper assumes that labelled data are available for all classes in the source domain. In practice, obtaining such comprehensive source datasets could be challenging, and a more feasible approach might assume data are distributed across multiple source domains. Previous DA methods are often prone to negative transfer when aligning datasets with only a subset of shared classes, whereas the presented framework can effectively align data using a limited number of shared classes. This capability supports the extension to a multi-source scenario, which could allow for the aggregation of class information from multiple-source monitoring campaigns by aligning the target domain to each source domain using a shared subset of classes. Even with data from multiple structures, novel classes may still arise in the target domain. Given that the proposed approach assigns a high sampling probability to unexplored regions in the feature space, it should facilitate querying and inclusion of new target classes, although this requires validation in future studies.\\

A mapping initialised via unsupervised DA can be used to define initial predictive models for active learning, even when labelled target data are sparse or unavailable. However, poor initial DA mappings may result from using limited or biased data, or from scenarios where domains are not sufficiently related, which may prevent a source model from generalising well to target data. In some cases, as shown in Section~4, such as the examples given in Figure~\ref{fig:nca feats 1}(b) or Figure~\ref{fig:nca feats 2}(b), it may be possible to rectify poor initial attempts at transfer by incorporating target labels via active learning. However, negative transfer is still an important issue, particularly as it can have compounding effects with sampling bias. Thus, methods to avoid negative transfer, such as the development of similarity quantification approaches \cite{pan2020transfer}, remain an important direction for future work.\\

In addition, this paper applies strict assumptions about the mapping form to minimise the risk of overfitting more complex mappings when limited labelled target data are available. A drawback of this approach is that, if these mapping assumptions are too restrictive, it may not be possible to learn a suitable mapping to facilitate label sharing. Further work could investigate ways to extend the proposed approach to accommodate more complex mappings while minimising the risk of overfitting; for example, perhaps a number of transformation operations could be selected from a set of candidates using sparsity-inducing priors. However, in the case studies presented in this paper, it was found that a linear mapping was sufficient, finding the presented approach to be capable of sharing value damage labels between different bridges.\\

Active learning in the context of TL presents several interesting considerations. In some cases, source domains may represent structures still in operation meaning that sampling strategies could consider querying data from multiple structures. An additional extension could involve a multi-task approach with a latent DA mapping, where multiple structures stream data with the objective of enhancing performance across all structures. In addition, this paper leverages a shared damage classifier to learn latent mapping parameters; however, other tasks could be used to learn the mapping, following a multi-task approach. Future work could investigate whether tasks with lower labelling costs, such as temperature data (which can be obtained without manual effort), can be used as target variables to learn a shared regression model, which could allow for a mapping to be inferred with few labels corresponding to inspections.\\

\section{ACKNOWLEDGEMENTS}
The authors would like to acknowledge the support of the UK Engineering and Physical Sciences Research Council via grants EP/R006768/1 and EP/W005816/1. For the purpose of open access, the authors have applied for a Creative Commons Attribution (CC-BY-ND) licence to any Author Accepted Manuscript version arising.

\begin{appendices}

\section{Low prior mapping variance for active transfer to a target without changing temperatures}

\begin{figure}[t!]
    \centering
    \begin{subfigure}[b]{0.8\textwidth}
         \centering
        \includegraphics[width=\textwidth]{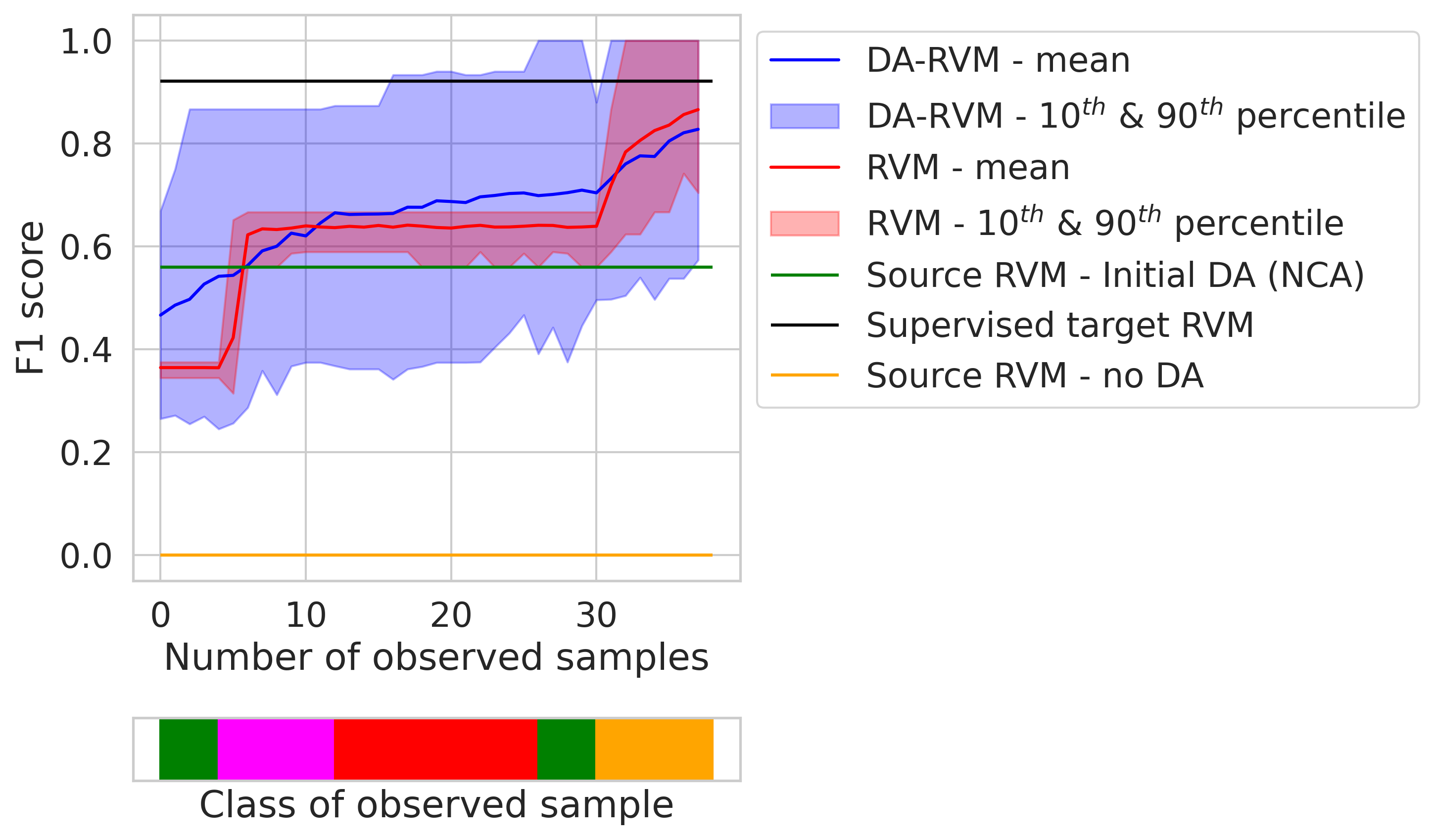}
         \caption{}
     \end{subfigure}
    \begin{subfigure}[b]{0.8\textwidth}
         \centering
         \includegraphics[width=\textwidth]{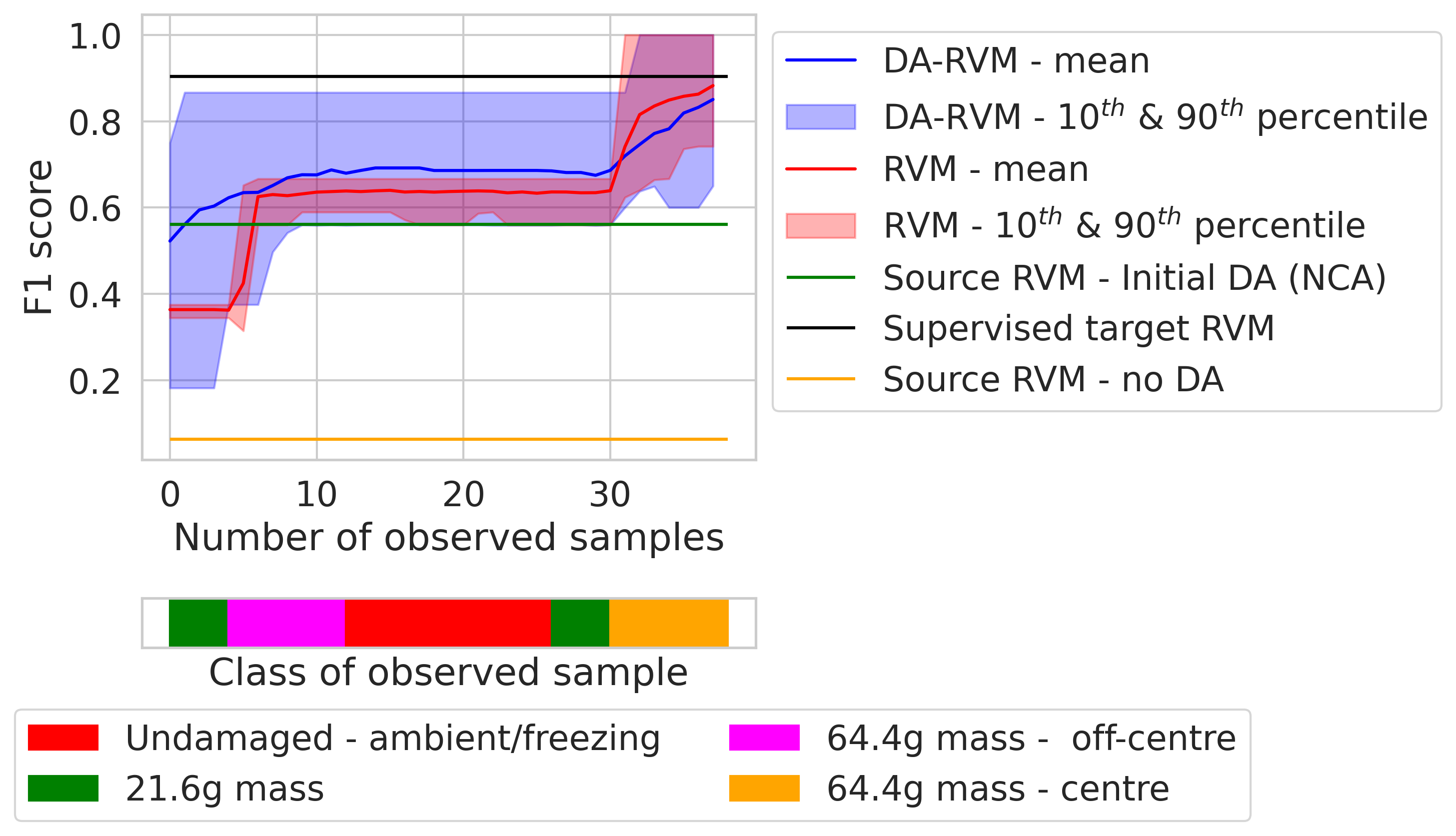}
         \caption{}
     \end{subfigure}
    \caption{Mean test f1 score vs the number of samples previously presented to the active learners for B1$\rightarrow$B3, shown in (a), and B2$\rightarrow$B3 presented in (b). }
    \label{fig:f1 active 22}
\end{figure}

This section provides the results for the second case study when prior variance on the scale and translation is defined to be more restrictive: $\sigma_t = \sigma_s = 0.1$. The F1 scores throughout the active learning process are presented in Figure \ref{fig:f1 active 22}. It can be seen in this case that, although the initial F1 scores are higher when defining lower prior variance, the mean F1 score was not improved to the same extent after observing the first damage scenario, and the final F1 score is slighly lower than the target-only RVM. This result is likely caused by the low prior variance preventing the mapping from learning large enough scale and translation values to update poor initial mappings found in some test repeats.\\

\begin{figure}[t!]
    \centering
        \begin{subfigure}[b]{0.35\textwidth}
         \centering
        \includegraphics[width=\textwidth]{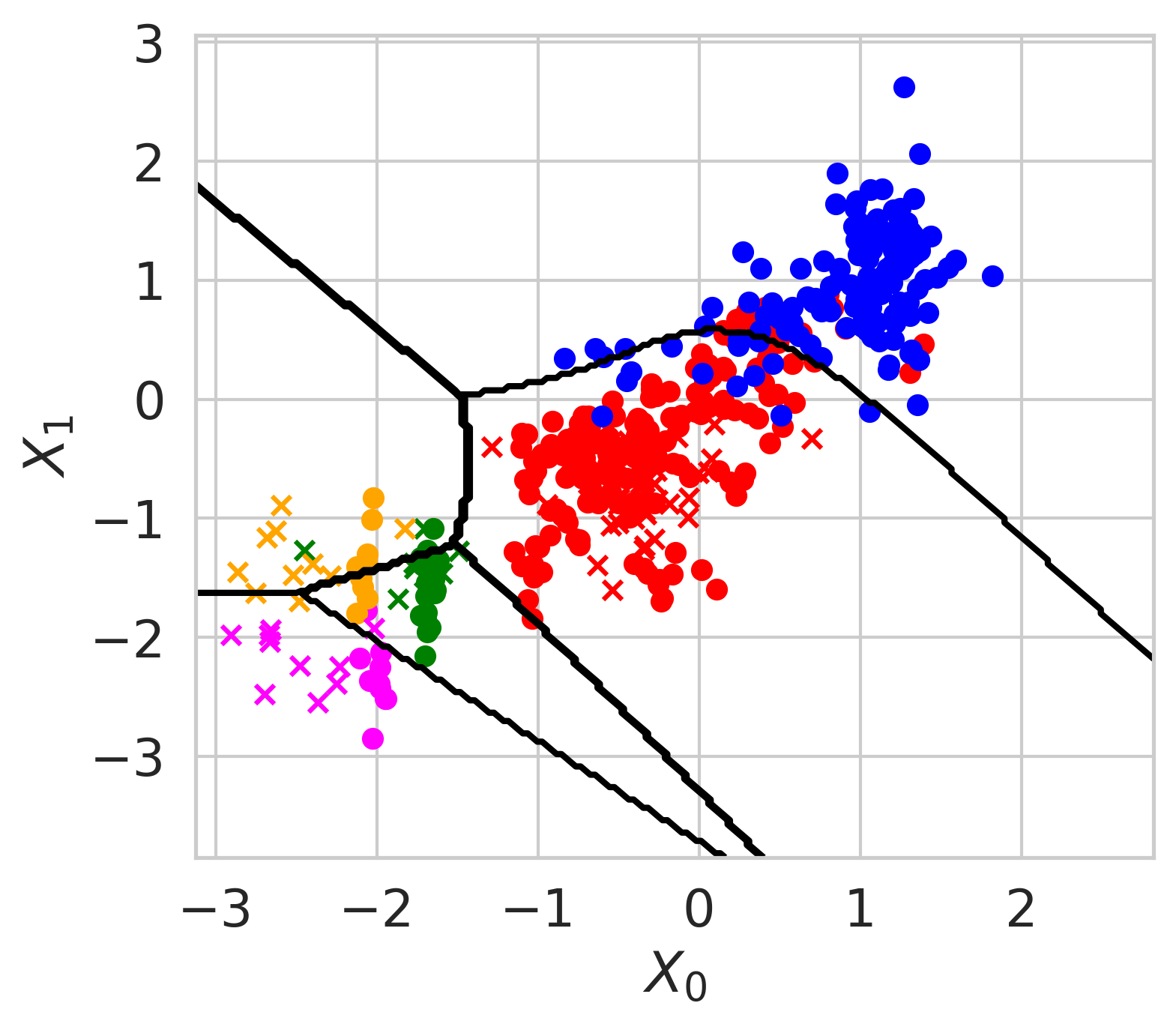}
         \caption{}
     \end{subfigure}
    \begin{subfigure}[b]{0.605\textwidth}
         \centering
\includegraphics[width=\textwidth]{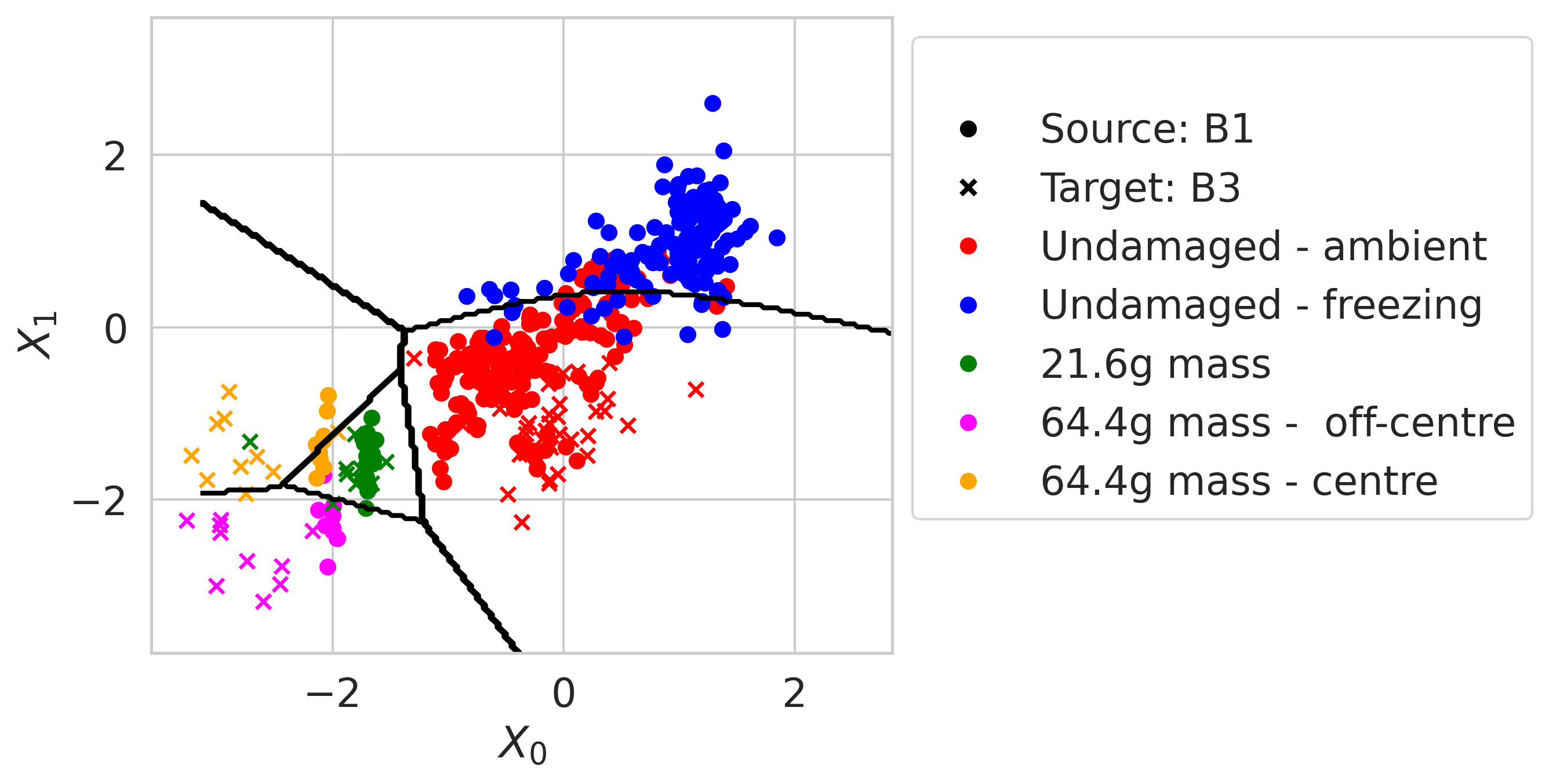}
         \caption{}
     \end{subfigure}
     \begin{subfigure}[b]{0.35\textwidth}
         \centering
\includegraphics[width=\textwidth]{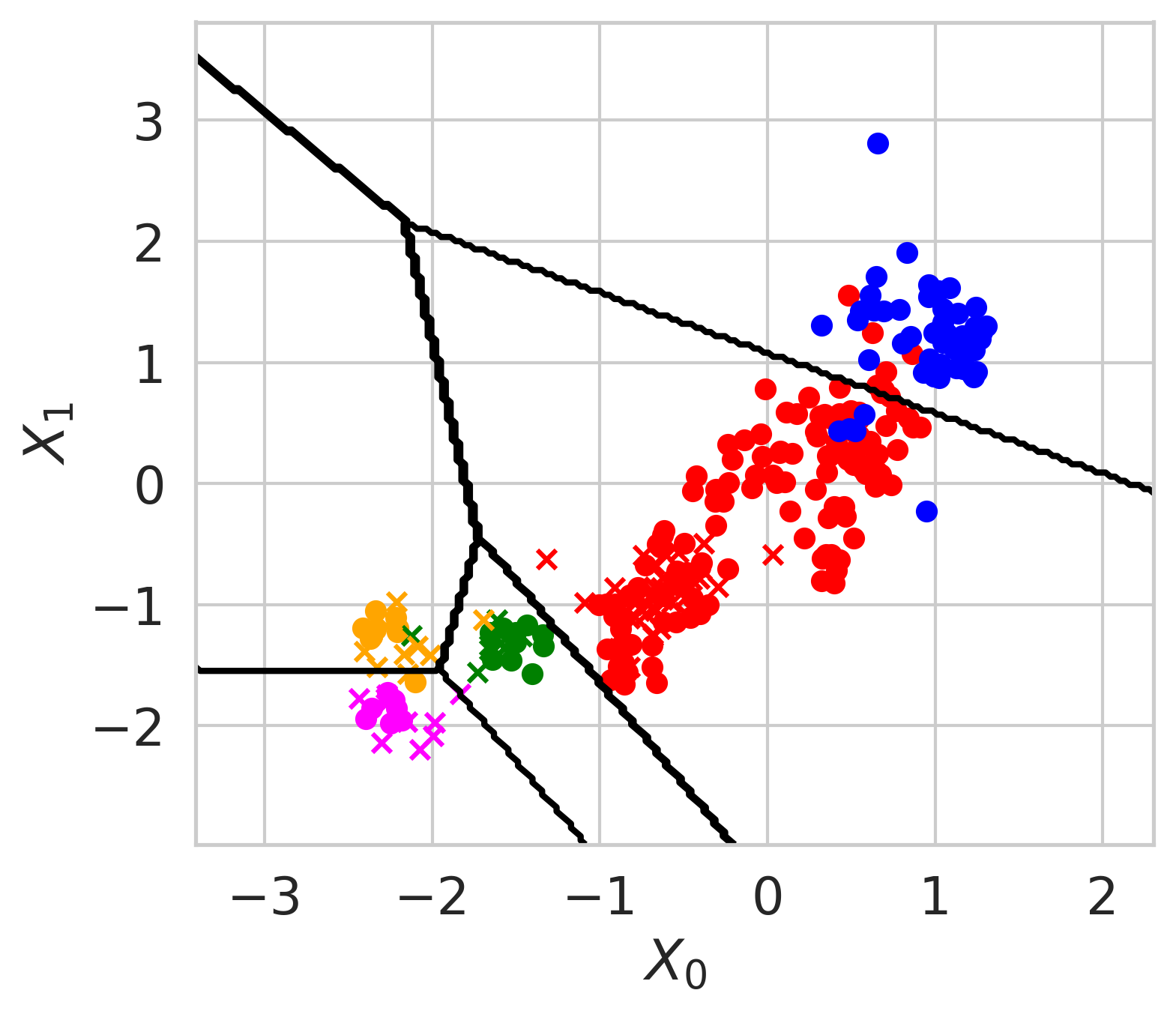}
         \caption{}
     \end{subfigure}
   \begin{subfigure}[b]{0.605\textwidth}
         \centering
\includegraphics[width=\textwidth]{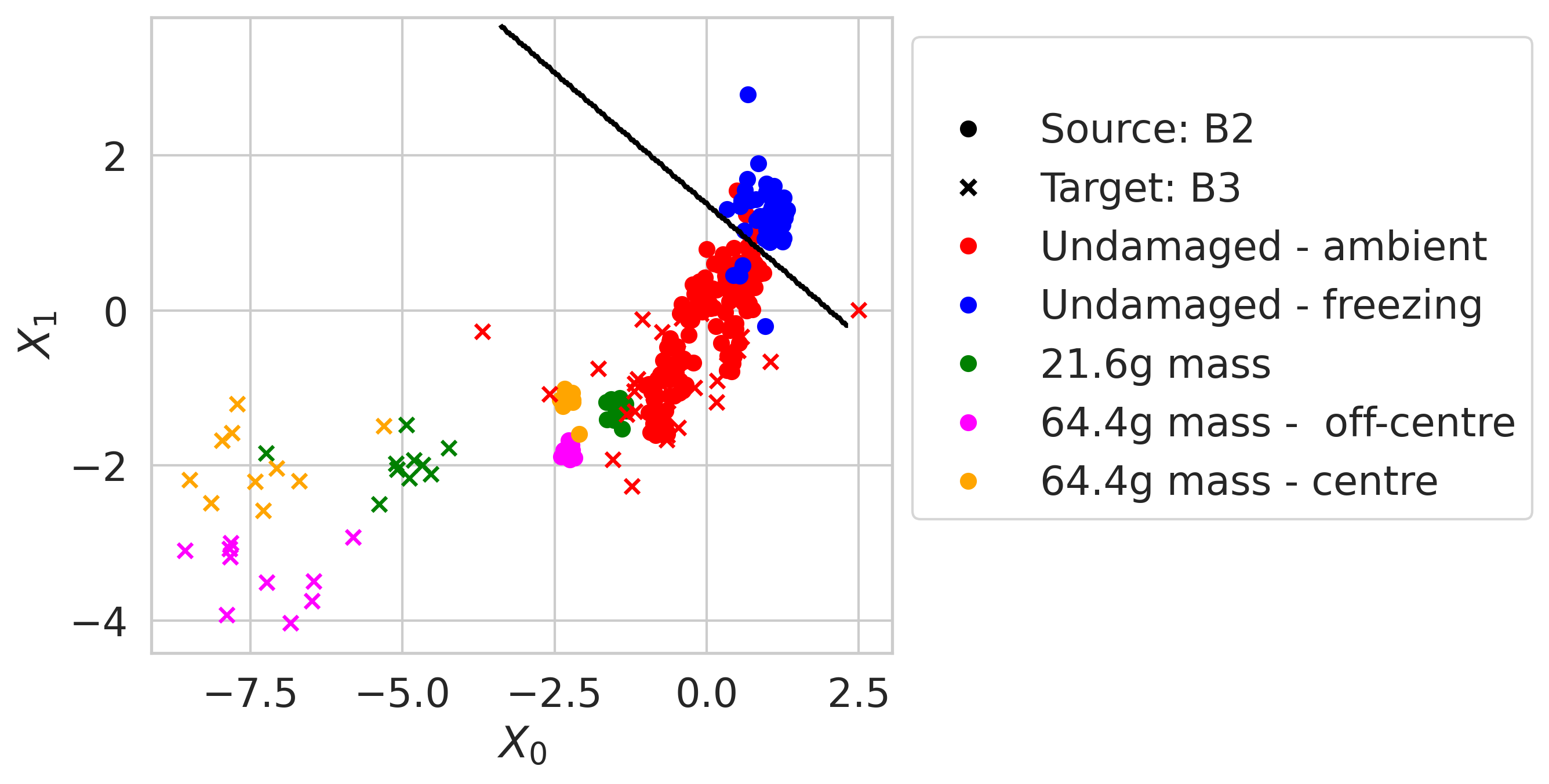}
         \caption{}
     \end{subfigure}
    \caption{Visualisation of the data (training and testing data), mapped via the expected posterior mapping after being presented with all data.  Figures (a) and (b) present the results from the ``best'' and ``worst'' prior mappings for B1$\rightarrow$B3, respectively, and results from the ``best'' and ``worst'' prior mappings for B2$\rightarrow$B3 presented in (c) and (d), respectively.}
    \label{fig:post feats 22}
\end{figure}

The features found via the expected posterior mapping after observing all data for the same ``best'' and ``worst'' test repeats as Figure \ref{fig:nca feats 2}, are presented in Figure \ref{fig:post feats 22}. It can be seen while the ``best'' mappings resulted in close alignment in the final feature space, shown in Figure \ref{fig:post feats 22}(a) and Figure \ref{fig:post feats 22}(c). The ``worst'' test repeat for B2$\rightarrow $B3, shown in Figure \ref{fig:post feats 22}(d) remains largely unchanged from the original NCA mapping  (Figure \ref{fig:nca feats 2}(b)). This result suggests that the prior mapping was too restrictive. It can also be seen that there is only a boundary between the ambient and freezing data in Figure \ref{fig:post feats 22}(d); this is because high mapping variance means data in most regions of the feature space were assigned a uniform label probability.\\

\section{Case study: active transfer learning under changing temperatures -- comparison with random sampling}

\begin{figure}[h!]
    \centering
        \begin{subfigure}[b]{0.31\textwidth}
         \centering
        \includegraphics[width=\textwidth]{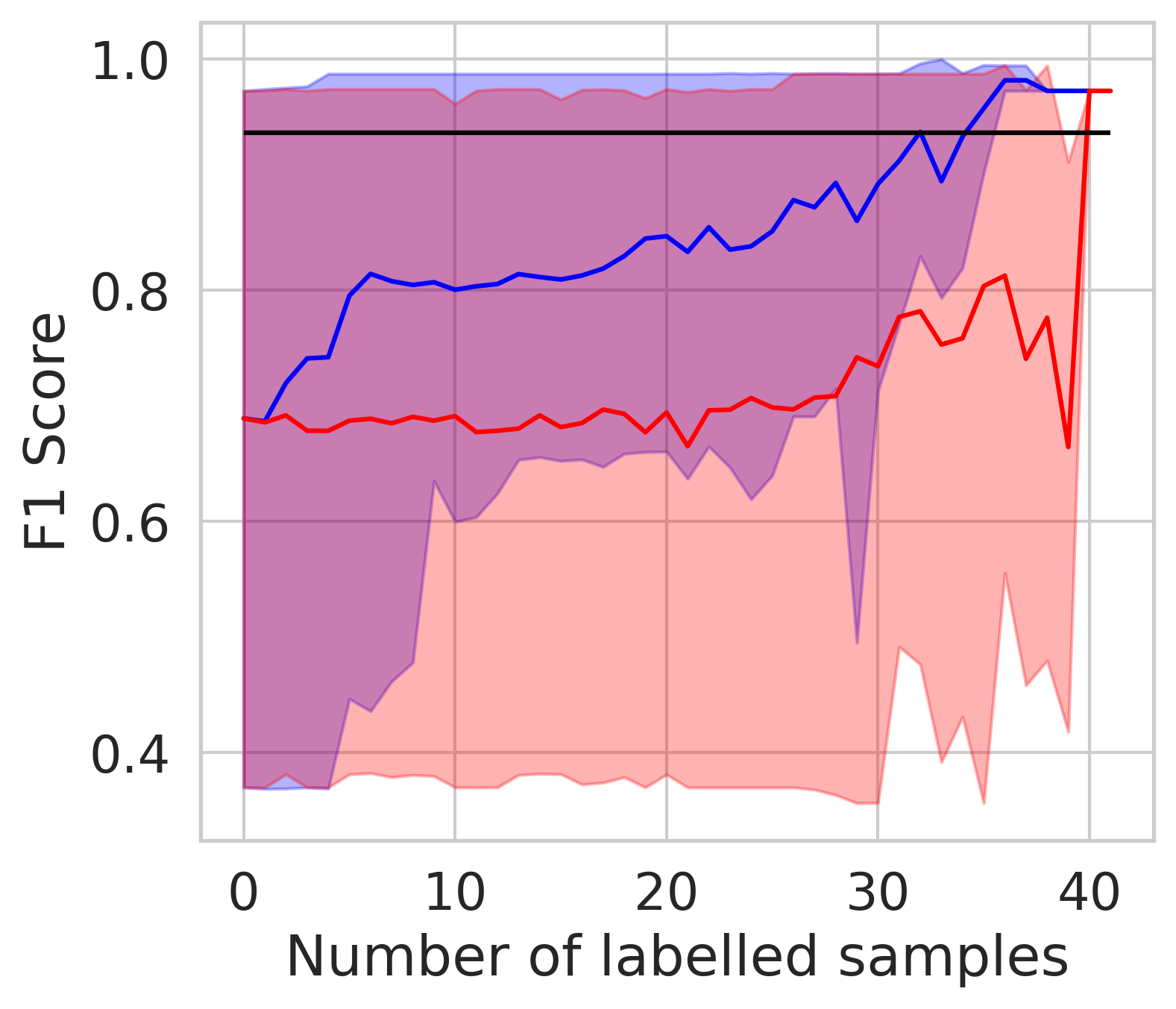}
         \caption{}
     \end{subfigure}
    \begin{subfigure}[b]{0.645\textwidth}
         \centering
\includegraphics[width=\textwidth]{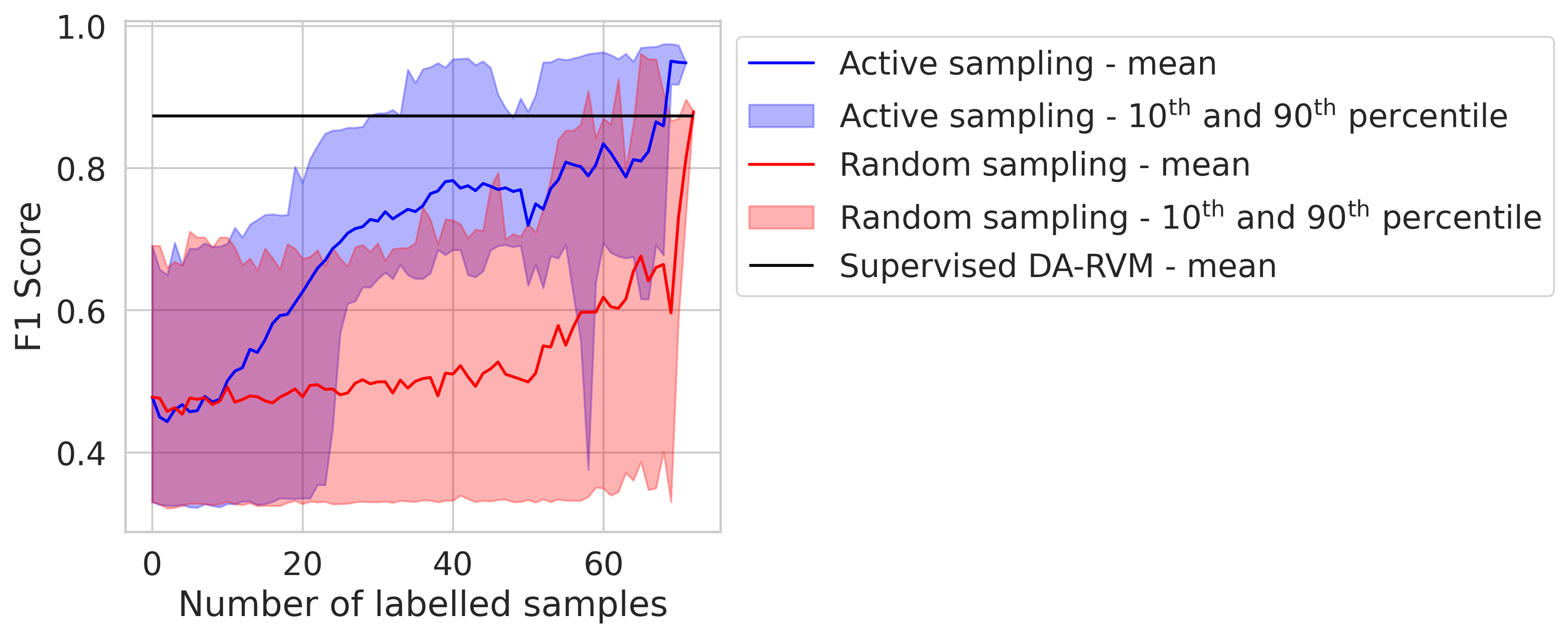}
         \caption{}
     \end{subfigure}
    \caption{The test F1 scores for the DA-RVM against the number of labelled samples selected via active sampling (shown in blue) and random sample (shown in red); B1$\rightarrow$B2 is shown in (a), and B2$\rightarrow$B1 is given in (b).}
    \label{fig:random 1}
\end{figure}

To verify the effectiveness of the active-sampling strategy, a comparison with random sampling is presented in Figure \ref{fig:random 1}(a) and Figure\ref{fig:random 1}(b), for B1$\rightarrow$B2 and B2$\rightarrow$B1 respectively. Random sampling results were generated by selecting samples at random from the entire target training dataset, selecting the number chosen by the active-sampling strategy for the given test repeat. In both cases, uncertainty sampling caused a sharper rise in F1 score and higher performance in the final model. There is a significant increase in the F1 score with random sampling and a reduction in the inter-percentile range at the higher end of the labelled sample count; however, this occurs because only a few test repeats led to this many queries, and these test repeats correspond to those with high F1 scores.\\ 

\section{Case study: active transfer to a target domain with limited data -- comparison with random sampling}

\begin{figure}[h!]
    \centering
        \begin{subfigure}[b]{0.31\textwidth}
         \centering
        \includegraphics[width=\textwidth]{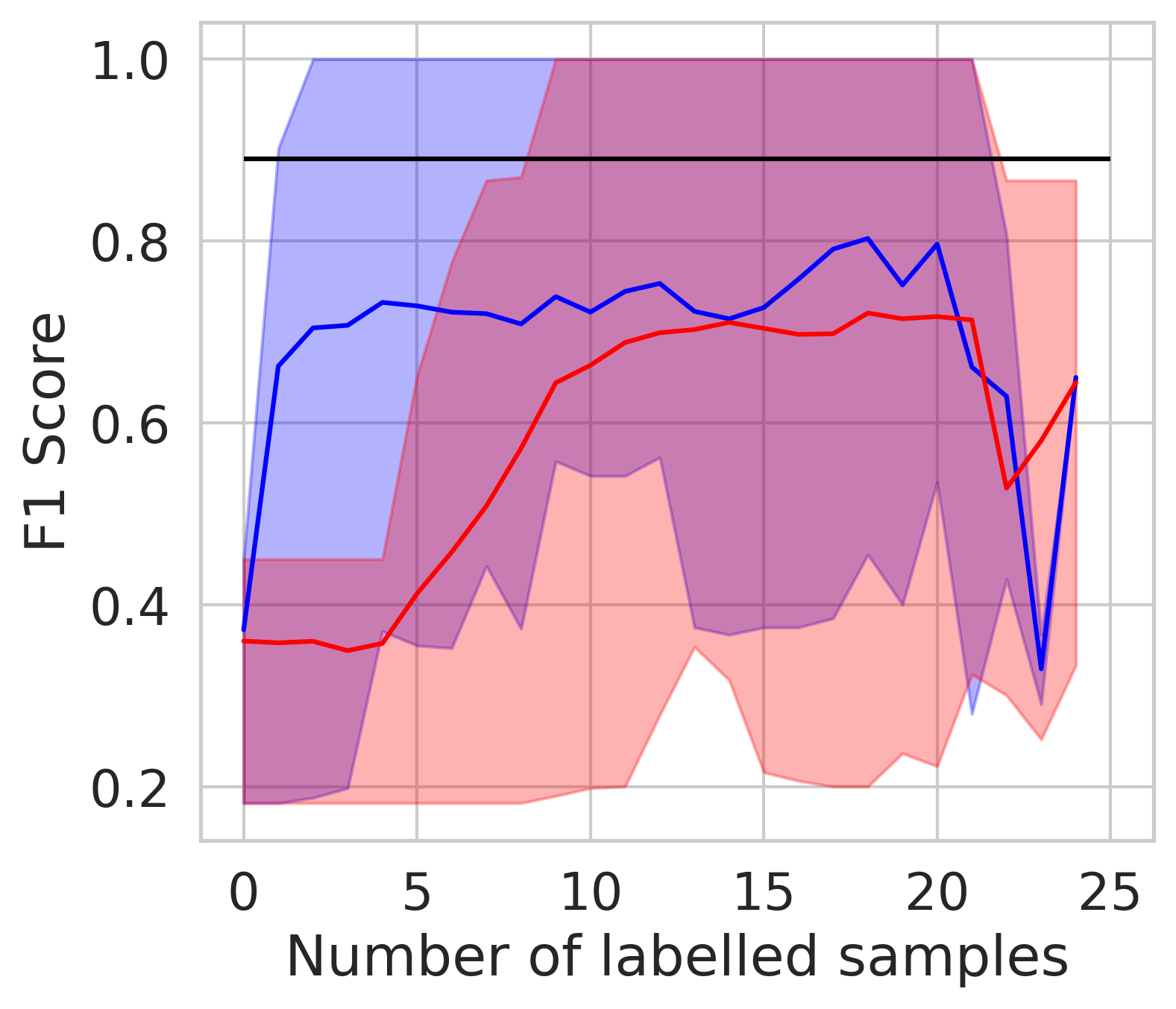}
         \caption{}
     \end{subfigure}
    \begin{subfigure}[b]{0.645\textwidth}
         \centering
\includegraphics[width=\textwidth]{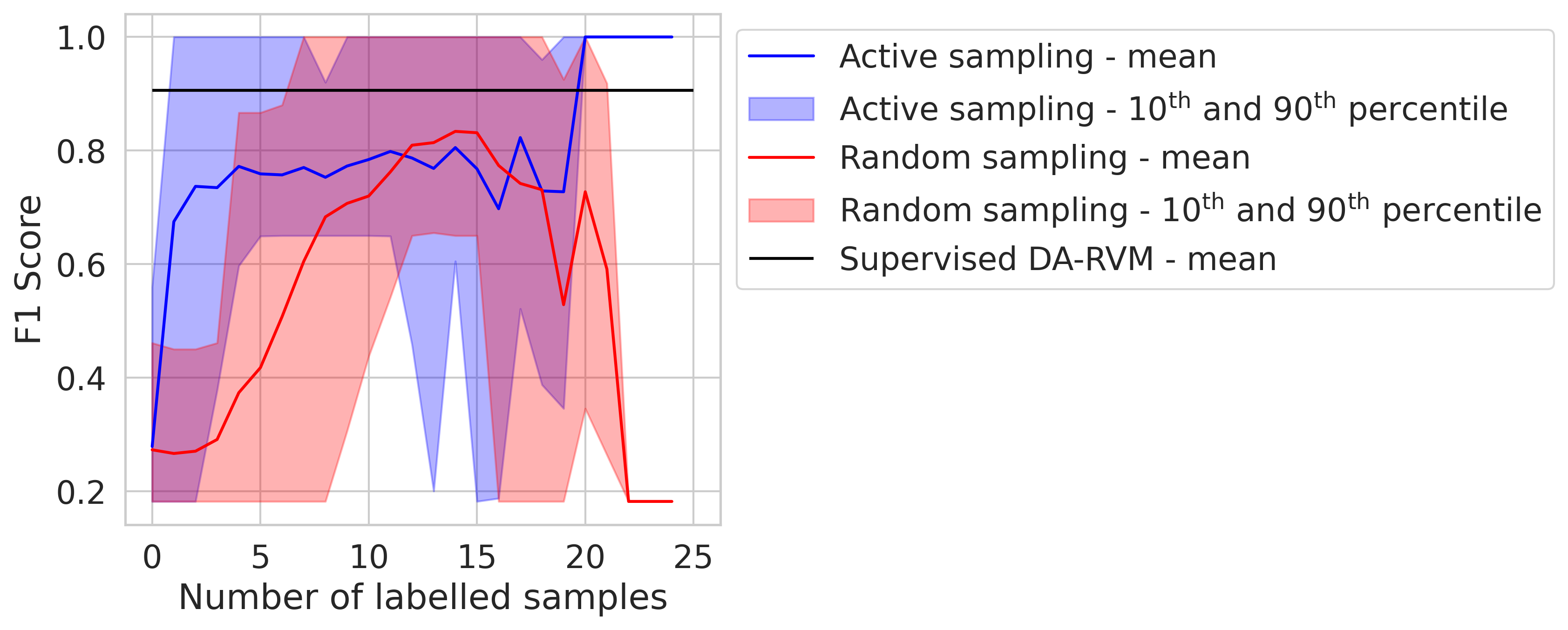}
         \caption{}
     \end{subfigure}
    \caption{The test F1 scores for the DA-RVM against the number of labelled samples selected via active sampling (shown in blue) and random sample (shown in red); B1$\rightarrow$B3 is shown in (a), and B2$\rightarrow$B3 is given in (b).}
    \label{fig:random 2}
\end{figure}

The active-sampling strategy is benchmarked against random sampling for the DA-RVM, as shown in Figure \ref{fig:random 2}. As with the previous case, active sampling results in improvements in the F1 scores with far fewer samples. In addition, the sudden changes in mean F1 scores seen as the number of labelled samples increases was caused by the small number of repeats that queried above 20 samples.\\

\newpage

\end{appendices}
\bibliographystyle{unsrt}
\bibliography{new_ref}

\begin{thebibliography}{10}

\bibitem{dervilis2014damage}
N~Dervilis, M~Choi, SG~Taylor, RJ~Barthorpe, G~Park, CR~Farrar, and K~Worden.
\newblock On damage diagnosis for a wind turbine blade using pattern recognition.
\newblock {\em Journal of Sound and Vibration}, 333:1833--1850, 2014.

\bibitem{FarrarC.R.CharlesR.2013Shm:}
C.R Farrar and K~Worden.
\newblock {\em {Structural health monitoring: A machine learning perspective}}.
\newblock Wiley, Chichester, 2013.

\bibitem{Gardner2021}
P.~Gardner, L.A. Bull, J.~Gosliga, N.~Dervilis, and K.~Worden.
\newblock {Foundations of population-based {SHM}, Part III: Heterogeneous populations – Mapping and transfer}.
\newblock {\em Mechanical Systems and Signal Processing}, 149:107142, 2021.

\bibitem{Gosliga2021}
J~Gosliga, P~Gardner, L~A Bull, N~Dervilis, and K~Worden.
\newblock {Foundations of Population-based {SHM}, Part II: Heterogeneous populations – Graphs, networks, and communities}.
\newblock {\em Mechanical Systems and Signal Processing}, 148:107144, 2021.

\bibitem{Bull2021}
L~A Bull, P~Gardner, J~Gosliga, T~J Rogers, N~Dervilis, E~J Cross, E~Papatheou, A~E Maguire, C~Campos, and K~Worden.
\newblock {Foundations of population-based {SHM}, Part I: Homogeneous populations and forms}.
\newblock {\em Mechanical Systems and Signal Processing}, 148:107141, 2021.

\bibitem{Murphy2014}
Kevin~P Murphy.
\newblock {\em Machine learning: a Probabilistic Perspective}.
\newblock MIT press, 2012.

\bibitem{yang2020transfer}
Qiang Yang, Yu~Zhang, Wenyuan Dai, and Sinno~Jialin Pan.
\newblock {\em Transfer Learning}.
\newblock Cambridge University Press, 2020.

\bibitem{Gardner2020}
P~Gardner, X~Liu, and K~Worden.
\newblock {On the application of domain adaptation in structural health monitoring}.
\newblock {\em Mechanical Systems and Signal Processing}, 138:106550, 2020.

\bibitem{gardner2022population}
P~Gardner, LA~Bull, J~Gosliga, J~Poole, N~Dervilis, and K~Worden.
\newblock A population-based {SHM} methodology for heterogeneous structures: Transferring damage localisation knowledge between different aircraft wings.
\newblock {\em Mechanical Systems and Signal Processing}, 172:108918, 2022.

\bibitem{Gardner2021a}
P~Gardner, T~J Rogers, C~Lord, and R~J Barthorpe.
\newblock {Learning model discrepancy: A Gaussian process and sampling-based approach}.
\newblock {\em Mechanical Systems and Signal Processing}, 152:107381, 2021.

\bibitem{bull2021transfer}
LA~Bull, PA~Gardner, N~Dervilis, E~Papatheou, M~Haywood-Alexander, RS~Mills, and K~Worden.
\newblock On the transfer of damage detectors between structures: an experimental case study.
\newblock {\em Journal of Sound and Vibration}, 501:116072, 2021.

\bibitem{Michau2019}
Gabriel Michau and Olga Fink.
\newblock Domain adaptation for one-class classification: monitoring the health of critical systems under limited information.
\newblock {\em arXiv preprint arXiv:1907.09204}, 2019.

\bibitem{wang2020triplet}
Xiaodong Wang and Feng Liu.
\newblock Triplet loss guided adversarial domain adaptation for bearing fault diagnosis.
\newblock {\em Sensors}, 20:320, 2020.

\bibitem{Li2020}
Yibin Li, Yan Song, Lei Jia, Shengyao Gao, Qiqiang Li, and Meikang Qiu.
\newblock Intelligent fault diagnosis by fusing domain adversarial training and maximum mean discrepancy via ensemble learning.
\newblock {\em IEEE Transactions on Industrial Informatics}, 17(4):2833--2841, 2020.

\bibitem{Cao2018a}
Zhangjie Cao, Mingsheng Long, Jianmin Wang, and Michael Jordan.
\newblock {Partial transfer learning with selective adversarial networks}.
\newblock {\em Proceedings of the IEEE Computer Society Conference on Computer Vision and Pattern Recognition}, pages 2724--2732, 2018.

\bibitem{Ben-David2010}
Shai Ben-David, John Blitzer, Koby Crammer, Alex Kulesza, Fernando Pereira, and Jennifer~Wortman Vaughan.
\newblock {A theory of learning from different domains}.
\newblock {\em Machine Learning}, 79(1-2):151--175, 2010.

\bibitem{feng2017deep}
Chen Feng, Ming-Yu Liu, Chieh-Chi Kao, and Teng-Yok Lee.
\newblock Deep active learning for civil infrastructure defect detection and classification.
\newblock In {\em Computing in Civil Engineering 2017}, pages 298--306. 2017.

\bibitem{chakraborty2015adaptive}
Debejyo Chakraborty, Narayan Kovvali, Antonia Papandreou-Suppappola, and Aditi Chattopadhyay.
\newblock An adaptive learning damage estimation method for structural health monitoring.
\newblock {\em Journal of Intelligent Material Systems and Structures}, 26:125--143, 2015.

\bibitem{bull2019probabilistic}
LA~Bull, K~Worden, TJ~Rogers, C~Wickramarachchi, EJ~Cross, T~McLeay, W~Leahy, and N~Dervilis.
\newblock A probabilistic framework for online structural health monitoring: active learning from machining data streams.
\newblock In {\em Journal of Physics: Conference Series}, volume 1264, page 012028. IOP Publishing, 2019.

\bibitem{hughes2022risk}
Aidan~J Hughes, Lawrence~A Bull, Paul Gardner, Robert~James Barthorpe, Nikolaos Dervilis, and Keith Worden.
\newblock On risk-based active learning for structural health monitoring.
\newblock {\em Mechanical Systems and Signal Processing}, 167:108569, 2022.

\bibitem{hughes2022robust}
Aidan~J Hughes, Lawrence~A Bull, Paul Gardner, Nikolaos Dervilis, and Keith Worden.
\newblock On robust risk-based active-learning algorithms for enhanced decision support.
\newblock {\em Mechanical Systems and Signal Processing}, 181:109502, 2022.

\bibitem{rai2010domain}
Piyush Rai, Avishek Saha, Hal Daum{\'e}~III, and Suresh Venkatasubramanian.
\newblock Domain adaptation meets active learning.
\newblock In {\em Proceedings of the NAACL HLT 2010 Workshop on Active Learning for Natural Language Processing}, pages 27--32, 2010.

\bibitem{saha2011active}
Avishek Saha, Piyush Rai, Hal Daum{\'e}, Suresh Venkatasubramanian, and Scott~L DuVall.
\newblock Active supervised domain adaptation.
\newblock In {\em Machine Learning and Knowledge Discovery in Databases: European Conference, ECML PKDD 2011, Athens, Greece, September 5-9, 2011, Proceedings, Part III 22}, pages 97--112. Springer, 2011.

\bibitem{xie2022active}
Binhui Xie, Longhui Yuan, Shuang Li, Chi~Harold Liu, Xinjing Cheng, and Guoren Wang.
\newblock Active learning for domain adaptation: An energy-based approach.
\newblock In {\em Proceedings of the AAAI conference on artificial intelligence}, volume~36, pages 8708--8716, 2022.

\bibitem{ma2021active}
Xinhong Ma, Junyu Gao, and Changsheng Xu.
\newblock Active universal domain adaptation.
\newblock In {\em Proceedings of the IEEE/CVF international conference on computer vision}, pages 8968--8977, 2021.

\bibitem{Zhuang2021}
Fuzhen Zhuang, Zhiyuan Qi, Keyu Duan, Dongbo Xi, Yongchun Zhu, Hengshu Zhu, Hui Xiong, and Qing He.
\newblock {A comprehensive survey on transfer learning}.
\newblock {\em Proceedings of the IEEE}, 109(1):43--76, 2021.

\bibitem{pan2020transfer}
Sinno~Jialin Pan.
\newblock Transfer learning.
\newblock {\em Learning}, 21:1--2, 2020.

\bibitem{Pan2010}
Sinno~Jialin Pan, Xiaochuan Ni, Jian~Tao Sun, Qiang Yang, and Zheng Chen.
\newblock {Cross-domain sentiment classification via spectral feature alignment}.
\newblock {\em Proceedings of the 19th International Conference on World Wide Web, WWW '10}, pages 751--760, 2010.

\bibitem{JialinPan2011}
Sinno {Jialin Pan}, Ivor~W Tsang, James~T Kwok, and Qiang Yang.
\newblock {Domain adaptation via transfer component analysis}.
\newblock {\em IEEE Transactions on Neural Networks}, 22(2), 2011.

\bibitem{Long2013}
Mingsheng Long, Jianmin Wang, Guiguang Ding, Jiaguang Sun, and Philip~S Yu.
\newblock Transfer feature learning with joint distribution adaptation.
\newblock {\em Proceedings of the IEEE International Conference on Computer Vision}, pages 2200--2207, 2013.

\bibitem{Ganin2017}
Yaroslav Ganin, Evgeniya Ustinova, Hana Ajakan, Pascal Germain, Hugo Larochelle, Fran{\c{c}}ois Laviolette, Mario Marchand, and Victor Lempitsky.
\newblock Domain-adversarial training of neural networks.
\newblock {\em The journal of machine learning research}, 17:2096--2030, 2016.

\bibitem{wang2019characterizing}
Zirui Wang, Zihang Dai, Barnab{\'a}s P{\'o}czos, and Jaime Carbonell.
\newblock Characterizing and avoiding negative transfer.
\newblock In {\em Proceedings of the IEEE/CVF conference on computer vision and pattern recognition}, pages 11293--11302, 2019.

\bibitem{poole2023towards}
J~Poole, P~Gardner, N~Dervilis, JH~Mclean, TJ~Rogers, and K~Worden.
\newblock Towards physics-based metrics for transfer learning in dynamics.
\newblock In {\em Society for Experimental Mechanics Annual Conference and Exposition}, pages 73--81. Springer, 2023.

\bibitem{hoffman2013efficient}
Judy Hoffman, Erik Rodner, Jeff Donahue, Trevor Darrell, and Kate Saenko.
\newblock Efficient learning of domain-invariant image representations.
\newblock {\em arXiv preprint arXiv:1301.3224}, 2013.

\bibitem{settles2009active}
Burr Settles.
\newblock Active learning literature survey.
\newblock Technical Report TR-1648, University of Wisconsin-Madison, Department of Computer Sciences, 2009.

\bibitem{mackay2003information}
David~JC MacKay.
\newblock {\em Information theory, inference and learning algorithms}.
\newblock Cambridge university press, 2003.

\bibitem{vapnik2013nature}
Vladimir Vapnik.
\newblock {\em The nature of statistical learning theory}.
\newblock Springer science \& business media, 2013.

\bibitem{dasgupta2008hierarchical}
Sanjoy Dasgupta and Daniel Hsu.
\newblock Hierarchical sampling for active learning.
\newblock In {\em Proceedings of the 25th international conference on Machine learning}, pages 208--215, 2008.

\bibitem{bull2018active}
Lawrence Bull, Keith Worden, Graeme Manson, and Nikolaos Dervilis.
\newblock Active learning for semi-supervised structural health monitoring.
\newblock {\em Journal of Sound and Vibration}, 437:373--388, 2018.

\bibitem{bull2022sampling}
Lawrence~A Bull, Nikolaos Dervilis, Keith Worden, Elizabeth~J Cross, and Timothy~J Rogers.
\newblock A sampling-based approach for information-theoretic inspection management.
\newblock {\em Proceedings of the Royal Society A}, 478(2262):20210790, 2022.

\bibitem{gardner2022domain}
Paul Gardner, Lawrence~A Bull, Nikolaos Dervilis, and Keith Worden.
\newblock Domain-adapted {G}aussian mixture models for population-based structural health monitoring.
\newblock {\em Journal of Civil Structural Health Monitoring}, 12:1343--1353, 2022.

\bibitem{giglioni2024domain}
Valentina Giglioni, Jack Poole, Ilaria Venanzi, Filippo Ubertini, and Keith Worden.
\newblock A domain adaptation approach to damage classification with an application to bridge monitoring.
\newblock {\em Mechanical Systems and Signal Processing}, 209:111135, 2024.

\bibitem{Gardner2020b}
P~Gardner, LA~Bull, N~Dervilis, and K~Worden.
\newblock Overcoming the problem of repair in structural health monitoring: Metric-informed transfer learning.
\newblock {\em Journal of Sound and Vibration}, page 116245, 2021.

\bibitem{Xu2020}
Susu Xu and Hae~Young Noh.
\newblock Phymdan: Physics-informed knowledge transfer between buildings for seismic damage diagnosis through adversarial learning.
\newblock {\em Mechanical Systems and Signal Processing}, 151:107374, 2021.

\bibitem{figueiredo2023transfer}
Eloi Figueiredo, Marcus Omori~Yano, Samuel Da~Silva, Ionut Moldovan, and Mihai Adrian~Bud.
\newblock Transfer learning to enhance the damage detection performance in bridges when using numerical models.
\newblock {\em Journal of Bridge Engineering}, 28:04022134, 2023.

\bibitem{Li2019}
Xiang Li, Wei Zhang, Qian Ding, and Jian~Qiao Sun.
\newblock {Multi-layer domain adaptation method for rolling bearing fault diagnosis}.
\newblock {\em Signal Processing}, 157:180--197, 2019.

\bibitem{jiao2020residual}
Jinyang Jiao, Ming Zhao, Jing Lin, and Kaixuan Liang.
\newblock Residual joint adaptation adversarial network for intelligent transfer fault diagnosis.
\newblock {\em Mechanical Systems and Signal Processing}, 145:106962, 2020.

\bibitem{cao2018preprocessing}
Pei Cao, Shengli Zhang, and Jiong Tang.
\newblock Preprocessing-free gear fault diagnosis using small datasets with deep convolutional neural network-based transfer learning.
\newblock {\em Ieee Access}, 6:26241--26253, 2018.

\bibitem{Gao2018}
Yuqing Gao and Khalid~M. Mosalam.
\newblock Deep transfer learning for image-based structural damage recognition.
\newblock {\em Computer-Aided Civil and Infrastructure Engineering}, 33(9):748--768, sep 2018.

\bibitem{dorafshan2018comparison}
Sattar Dorafshan, Robert~J Thomas, and Marc Maguire.
\newblock Comparison of deep convolutional neural networks and edge detectors for image-based crack detection in concrete.
\newblock {\em Construction and Building Materials}, 186:1031--1045, 2018.

\bibitem{zhu2020vision}
Jinsong Zhu, Chi Zhang, Haidong Qi, and Ziyue Lu.
\newblock Vision-based defects detection for bridges using transfer learning and convolutional neural networks.
\newblock {\em Structure and Infrastructure Engineering}, 16:1037--1049, 2020.

\bibitem{teng2023structural}
Shuai Teng, Xuedi Chen, Gongfa Chen, and Li~Cheng.
\newblock Structural damage detection based on transfer learning strategy using digital twins of bridges.
\newblock {\em Mechanical Systems and Signal Processing}, 191:110160, 2023.

\bibitem{xu2021attribute}
Yang Xu, Yuequan Bao, Yufeng Zhang, and Hui Li.
\newblock Attribute-based structural damage identification by few-shot meta learning with inter-class knowledge transfer.
\newblock {\em Structural Health Monitoring}, 20:1494--1517, 2021.

\bibitem{gardner2022application}
P~Gardner, LA~Bull, N~Dervilis, and K~Worden.
\newblock On the application of kernelised bayesian transfer learning to population-based structural health monitoring.
\newblock {\em Mechanical Systems and Signal Processing}, 167:108519, 2022.

\bibitem{gonen2014kernelized}
Mehmet G{\"o}nen and Adam Margolin.
\newblock Kernelized {B}ayesian transfer learning.
\newblock In {\em Proceedings of the AAAI Conference on Artificial Intelligence}, volume~28, 2014.

\bibitem{duan2012learning}
Lixin Duan, Dong Xu, and Ivor Tsang.
\newblock Learning with augmented features for heterogeneous domain adaptation.
\newblock {\em arXiv preprint arXiv:1206.4660}, 2012.

\bibitem{martinez2019towards}
Giovanna Martinez~Arellano and Svetan Ratchev.
\newblock Towards an active learning approach to tool condition monitoring with bayesian deep learning.
\newblock 2019.

\bibitem{clarkson2024active}
Daniel~R Clarkson, Lawrence~A Bull, Chandula~T Wickramarachchi, Elizabeth~J Cross, Timothy~J Rogers, Keith Worden, Nikolaos Dervilis, and Aidan~J Hughes.
\newblock Active learning for regression in engineering populations: A risk-informed approach.
\newblock {\em arXiv preprint arXiv:2409.04328}, 2024.

\bibitem{echard2011ak}
Benjamin Echard, Nicolas Gayton, and Maurice Lemaire.
\newblock Ak-mcs: an active learning reliability method combining kriging and monte carlo simulation.
\newblock {\em Structural safety}, 33(2):145--154, 2011.

\bibitem{dai2025adaptive}
Hongzhe Dai, Dashuai Li, and Michael Beer.
\newblock Adaptive kriging-assisted multi-fidelity subset simulation for reliability analysis.
\newblock {\em Computer Methods in Applied Mechanics and Engineering}, 436:117705, 2025.

\bibitem{Ben-David2007}
Shai Ben-David, John Blitzer, Koby Crammer, and Fernando Pereira.
\newblock {Analysis of representations for domain adaptation}.
\newblock {\em Advances in Neural Information Processing Systems}, pages 137--144, 2007.

\bibitem{golub2013matrix}
Gene~H Golub and Charles~F Van~Loan.
\newblock {\em Matrix computations}.
\newblock JHU press, 2013.

\bibitem{tipping1999relevance}
Michael Tipping.
\newblock The relevance vector machine.
\newblock {\em Advances in Neural Information Processing Dystems}, 12, 1999.

\bibitem{damoulas2008inferring}
Theodoros Damoulas, Yiming Ying, Mark~A Girolami, and Colin Campbell.
\newblock Inferring sparse kernel combinations and relevance vectors: an application to subcellular localization of proteins.
\newblock In {\em 2008 Seventh International Conference on Machine Learning and Applications}, pages 577--582. IEEE, 2008.

\bibitem{bingham2019pyro}
Eli Bingham, Jonathan~P. Chen, Martin Jankowiak, Fritz Obermeyer, Neeraj Pradhan, Theofanis Karaletsos, Rohit Singh, Paul~A. Szerlip, Paul Horsfall, and Noah~D. Goodman.
\newblock Pyro: deep universal probabilistic programming.
\newblock {\em J. Mach. Learn. Res.}, 20:28:1--28:6, 2019.

\bibitem{hoffman2014no}
Matthew~D Hoffman, Andrew Gelman, et~al.
\newblock The no-u-turn sampler: adaptively setting path lengths in hamiltonian monte carlo.
\newblock {\em J. Mach. Learn. Res.}, 15:1593--1623, 2014.

\bibitem{hughes2024quantifying}
Aidan~J Hughes, Jack Poole, Nikolaos Dervilis, Paul Gardner, and Keith Worden.
\newblock Quantifying the value of information transfer in population-based {SHM}.
\newblock In {\em IMAC, A Conference and Exposition on Structural Dynamics}, pages 19--31. Springer, 2024.

\bibitem{poole2023statistic}
Jack Poole, Paul Gardner, Nikolaos Dervilis, Lawrence Bull, and Keith Worden.
\newblock On statistic alignment for domain adaptation in structural health monitoring.
\newblock {\em Structural Health Monitoring}, 22(3):1581--1600, 2023.

\bibitem{Sun2015}
Shiliang Sun, Honglei Shi, and Yuanbin Wu.
\newblock {A survey of multi-source domain adaptation}.
\newblock {\em Information Fusion}, 24:84--92, 2015.

\bibitem{conti2023physics}
Zack~Xuereb Conti, Ruchi Choudhary, and Luca Magri.
\newblock A physics-based domain adaptation framework for modeling and forecasting building energy systems.
\newblock {\em Data-Centric Engineering}, 4:e10, 2023.

\bibitem{hughes2021probabilistic}
Aidan~J Hughes, Robert~J Barthorpe, N~Dervilis, Charles~R Farrar, and Keith Worden.
\newblock A probabilistic risk-based decision framework for structural health monitoring.
\newblock {\em Mechanical Systems and Signal Processing}, 150:107339, 2021.

\bibitem{HighwaysEngland2020}
{Highways England}.
\newblock {Freedom of Information Request on Bridges}.
\newblock {Response to FOI Request Ref: 101497}, November 2020.
\newblock {Highways England Official Correspondence Team, Bedford, UK}.

\bibitem{maeck2001damage}
Johan Maeck, Bart Peeters, and Guido De~Roeck.
\newblock Damage identification on the {Z}24 bridge using vibration monitoring.
\newblock {\em Smart Materials and Structures}, 10:512, 2001.

\bibitem{papatheou2010use}
Evangelos Papatheou, Graeme Manson, Robert~J Barthorpe, and Keith Worden.
\newblock The use of pseudo-faults for novelty detection in shm.
\newblock {\em Journal of Sound and Vibration}, 329(12):2349--2366, 2010.

\bibitem{garcia2020mova}
Enrique Garc{\'\i}a-Mac{\'\i}as and Filippo Ubertini.
\newblock Mova/moss: Two integrated software solutions for comprehensive structural health monitoring of structures.
\newblock {\em Mechanical Systems and Signal Processing}, 143:106830, 2020.

\bibitem{Poole2024}
Jack Poole, Valentina Giglioni, Keith Worden, and Robin Mills.
\newblock {Transfer learning for bridge monitoring: model testing of four lab-scale multi-span continuous girder bridges under changing temperatures and damage conditions}.
\newblock 11 2024.

\bibitem{giglioni2025transfer}
Valentina Giglioni, Jack Poole, Robin Mills, Ilaria Venanzi, Filippo Ubertini, and Keith Worden.
\newblock Transfer learning in bridge monitoring: Laboratory study on domain adaptation for population-based shm of multispan continuous girder bridges.
\newblock {\em Mechanical Systems and Signal Processing}, 224:112151, 2025.

\bibitem{allemang2003modal}
Randall~J Allemang.
\newblock The modal assurance criterion--twenty years of use and abuse.
\newblock {\em Sound and Vibration}, 37:14--23, 2003.

\end{thebibliography}

\newpage

\end{document}